\documentclass[twoside,11pt]{article}

\usepackage{jmlr2e}
\usepackage{amsmath}
\usepackage{color}
\usepackage{bm}
\usepackage{bbm}
\usepackage{algorithm}
\usepackage{algorithmic}
\usepackage{amsfonts}
\usepackage{mathtools}
\usepackage{comment}
\usepackage{amsmath}
\usepackage{amssymb}
\usepackage{mathrsfs}
\usepackage{eqnarray}
\usepackage{dcolumn}
\usepackage{graphicx}
\usepackage{subfig}

\definecolor{forestgreen}{rgb}{0.13, 0.55, 0.13}

\ShortHeadings{Change Surfaces}{Herlands, Neill, Nickisch, and Wilson}
\firstpageno{1}

\begin{document}

\title{Change Surfaces for Expressive Multidimensional \\Changepoints and Counterfactual Prediction}

\author{\name William Herlands \email herlands@cmu.edu \\
       \addr Event and Pattern Detection Laboratory\\
       H.J. Heinz III College and Machine Learning Department\\
       Carnegie Mellon University\\
       Pittsburgh, PA 15213, USA
       \AND
       \name Daniel B. Neill \email daniel.neill@nyu.edu \\
       \addr Center for Urban Science and Progress, NYU Wagner School of\\
       Public Service, and NYU Courant Department of Computer Science\\
       New York University\\
       Brooklyn, NY 11201, USA
       \AND
       \name Hannes Nickisch \email hannes@nickisch.org \\
       \addr Digital Imaging \\
	Philips Research Hamburg \\
	R{\"o}ntgenstra{\ss}e 24-26 \\
	22335 Hamburg, Germany
       \AND
       \name Andrew Gordon Wilson \email andrew@cornell.edu \\
       \addr Operations Research and Information Engineering\\
       Cornell University\\
       Ithaca, NY 14853, USA
       }

\editor{}

\maketitle

\begin{abstract}
Identifying changes in model parameters is fundamental in machine learning and statistics.
However, standard changepoint models are limited in expressiveness, often addressing unidimensional problems and assuming instantaneous changes. We introduce \emph{change surfaces} as a multidimensional and highly expressive generalization of changepoints. We provide a model-agnostic formalization of change surfaces, illustrating how they can provide variable, heterogeneous, and non-monotonic rates of change across multiple dimensions. Additionally, we show how change surfaces can be used for counterfactual prediction. As a concrete instantiation of the change surface framework, we develop Gaussian Process Change Surfaces (GPCS).  We demonstrate counterfactual prediction with Bayesian posterior mean and credible sets, as well as massive scalability by introducing novel methods for additive non-separable kernels. Using two large spatio-temporal datasets we employ GPCS to discover and characterize complex changes that can provide scientific and policy relevant insights. Specifically, we analyze twentieth century measles incidence across the United States and discover previously unknown heterogeneous changes after the introduction of the measles vaccine. Additionally, we apply the model to requests for lead testing kits in New York City, discovering distinct spatial and demographic patterns.
\end{abstract}


\section{Introduction} 
\label{sec:introduction}

Detecting and modeling changes in data is critical in statistical theory, scientific discovery, and public policy. For example, in epidemiology, detecting changes in disease dynamics can provide information about when and where a vaccination program becomes effective. In dangerous professions such as coal mining, changes in accident occurrence patterns can indicate which regulations impact worker safety. In city governance, policy makers may be interested in how requests for health services change across space and over time.

Changepoint models have a long history in statistics, beginning in the mid-twentieth century, when methods were first developed to identify changes in a data generating process \citep{page1954continuous, horvath2014extensions}. The primary goal of these models is to determine if a change in the distribution of the data has occurred, and then to locate one or more points in the domain where such changes occur. While identifying these changepoints is an important result in itself, changepoint methods are also frequently applied to other problems such as outlier detection or failure analysis \citep{reece2015anomaly, tartakovsky2013efficient, Kapur2011}. Different changepoint methods are distinguished by the diversity of changepoints they are able to detect and the complexity of the underlying data. The simplest models consider mean shifts between functional regimes \citep{chernoff1964estimating, killick2012optimal}, while others consider changes in the covariance structure or higher order moments \citep{keshavarz2018optimal, ross2013parametric, james2013ecp}. A \emph{regime} is a particular data generating process or underlying function that is separated from other underlying processes or functions by changepoints.  Additionally, there is a fundamental distinction between changepoint models that identify changes sequentially using online algorithms, and those that analyze data retrospectively to find one or more changes in past data \citep{brodsky2013nonparametric,chen2011parametric}.
Finally, changepoint methods may be fully parametric, semi-parametric, or nonparametric \citep{ross2013parametric, guan2004semiparametric}. For additional discussion of changepoints beyond the scope of this paper, readers may consider the literature reviews in \citet{aue2013structural}, \citet{ivanoff2010optimal}, and \citet{aminikhanghahi2017survey}.

Yet nearly all changepoint methods described in the statistics and machine learning literature consider system perturbations as discrete changepoints. This literature seeks to identify instantaneous differences in parameter distributions. The advantage of such models is that they provide definitive assessments of the location of one or more changepoints. This approach is reasonable, for instance, when considering catastrophic events in a mechanical system, such as the effect of a car crash on various embedded sensor readings. Yet the challenge with these models is that real world systems rarely exhibit a clear binary transition between regimes. Indeed, in many applications, such as in biological science, instantaneous changes may be physically impossible. While a handful of approaches consider non-discrete changepoints \citep[e.g.,][]{wilson2011gaussian, wilson2014covariance, lloyd2014automatic} they still require linear, monotonic, one-dimensional, and, in practice, relatively quick changes. Existing models do not provide the expressiveness necessary to model complex changes. 

Additionally, applying changepoints to multiple dimensions, such as spatio-temporal data, is theoretically and practically non-trivial, and has thus been seldom attempted. Notable exceptions include \citet{majumdar2005spatio} who consider discrete spatio-temporal changepoints with three additive Gaussian processes: one for $t \leq t_0$, one for $t > t_0$, and one for all $t$. Alternatively, \citet{nicholls2010building} use a Bayesian onset-field process on a lattice to model the spatio-temporal distribution of human settlement on the Fiji islands. However, the models in these papers are limited to considering discrete changepoints.

\subsection{Main contributions}

In this paper, we introduce \emph{change surfaces} as expressive, multidimensional generalizations of changepoints. We present a model-agnostic formulation of change surfaces and instantiate this framework with scalable Gaussian process models. The resulting model is capable of automatically learning expressive covariance functions and a sophisticated continuous change surface. Additionally, we derive massively scalable inference procedures, as well as counterfactual prediction techniques. Finally, we apply the proposed methods to a wide variety of numerical data and complex human systems. In particular, we:
\begin{enumerate}
\item Introduce change surfaces as multidimensional and highly flexible generalizations of changepoint modeling.
\item Introduce a procedure which allows one to specify background functions and change functions, for more powerful inductive biases and added interpretability.
\item Provide a new framework for counterfactual prediction using change surfaces.
\item Present the Gaussian Process Change Surface model (GPCS) which models change surfaces with highly flexible Random Kitchen Sink \citep{rahimi2007random} features. 
\item Develop massively scalable additive, non-stationary, non-separable kernels by using the Weyl inequality \citep{weyl1912asymptotische} and novel Kronecker methods. In addition we integrate our approach into the recent KISS-GP framework \citep{wilson2015kernel}. The resulting approach is the first scalable Gaussian process multidimensional changepoint model.
\item Describe a novel initialization method for spectral mixture kernels \citep{wilson2013gaussian} by fitting a Gaussian mixture model to the Fourier transform of the data. This method provides good starting values for hyperparameters of expressive stationary kernels, allowing for successful optimization over a multimodal parameter space.
\item Demonstrate that the GPCS approach is robust to misspecification, and automatically discourages extraneous model complexity, leading to the discovery of interpretable generative hypotheses for the data.
\item Perform counterfactual prediction in complex real world data with posterior mean and covariance estimates for each point in the input domain.
\item Use GPCS for discovering and characterizing continuous changes in large observational data. We demonstrate our approach on a recently released public health dataset providing new insight that suggests how the effect of the 1963 measles vaccine may have varied over space and time in the United States. Additionally, we apply the model to requests for lead testing kits in New York City from 2014-2016. The results illustrate distinct spatial patterns in increased concern about lead-tainted water.
\end{enumerate}

\subsection{Outline}

The paper is divided into three main units.

Section \ref{sec:Change_Surfaces} formally introduces the notion of change surfaces as a multidimensional, expressive generalization of changepoints. We discuss a variant of change surfaces in section \ref{sec:Change_Surfaces_Half} and detail how to use change surfaces for counterfactual prediction in section \ref{sec:Counterfactual_Prediction}. The discussion of change surfaces in this unit is method-agnostic, and should be relevant to experts from a wide variety of statistical and machine learning disciplines. We emphasize the novel contribution of this framework to the general field of change detection.

Section \ref{sec:GPCSmodel} presents the Gaussian Process Change Surface (GPCS) as a scalable method for change surface modeling. We review Gaussian process basics in section \ref{sec:GPs}. We specify the GPCS model in section \ref{sec:changepoint_model}. Counterfactual predictions with GPCS are derived in section \ref{sec:GPCS_CF}. Scalable inference using novel Kronecker methods are presented in section \ref{sec:inference}, and we describe a novel initialization technique for expressive Gaussian process kernels in section \ref{sec:initialization}.

Section \ref{sec:experiments} demonstrates GPCS on \emph{out-of-class} numerical data and complex spatio-temporal data. We describe our numerical setup in section \ref{sec:numerical_exp} presenting results for posterior prediction, change surface identification, and counterfactual prediction. We present a one-dimensional application of GPCS on coal mining data in section \ref{sec:coal_exp} including a comparison to state-of-the-art changepoint methods. Moving to spatio-temporal data, we apply GPCS to model requests for lead testing kits in New York City in section \ref{sec:lead_exp} and discuss the policy relevant conclusions. Additionally, we use GPCS to model measles incidence in the United States in section \ref{sec:disease_exp} and discuss scientifically relevant insights.

Finally, we conclude with summary remarks in section~\ref{sec:conclusions}.

\section{Change surfaces}
\label{sec:Change_Surfaces}

In human systems and scientific phenomena we are often confronted with changes or perturbations which may not immediately disrupt an entire system. Instead, changes such as policy interventions and natural disasters take time to affect deeply ingrained habits or trickle through a complex bureaucracy. The dynamics of these changes are non-trivial, with sophisticated distributions, rates, and intensity functions. Using expressive models to fully characterize such changes is essential for accurate predictions and scientifically meaningful results. For example, in the spatio-temporal domain, changes are often heterogeneously distributed across space and time. Capturing the complexity of these changes provides useful insights for future policy makers enabling them to better target or structure policy interventions.

In order to provide the expressive capability for such models, we introduce the notion of a \emph{change surface} as a generalization of changepoints. We assume data are $(x,y)$, where $x = \{x_1,\dots,x_n\}, x_i \in \mathbb{R}^D$, are inputs or covariates, and $y = \{y_1,\dots,y_n\}$, $y_i \in \mathbb{R}$, are outputs or response variables indexed by $x$. A change surface defines transitions between latent functions $f_1,\dots,f_r$ defining $r$ regimes in the data. Unlike with changepoints, we do not require that the transitions be discrete. Instead we define $r$ warping functions $s(x) = [s_1(x),\dots,s_r(x)]$ where $s_i(x):\mathbb{R}^D \rightarrow [0,1]$, which have support over the entire domain of $x$. Importantly, these warping functions have an inductive bias towards $\{0,1\}$ creating a soft mutual exclusivity between the functions. We define the canonical form of a change surface as
\begin{equation}
\begin{aligned}
\label{eq:y_CS}
y(x) &= s_1(x)f_1(x) + \dots + s_r(x)f_r(x) + \epsilon \\
&s.t.\\
&\sum_{i=1}^r s_i(x) = 1\\
&s_i(x) \geq 0
\end{aligned}
\end{equation}
where $\epsilon(x)$ is noise. Each $s_i(x)$ defines how the coverage of $f_i(x)$ varies over the input domain. Where $s_i(x) \approx 1$, $f_i(x)$ dominates and primarily describes the relationship between $x$ and $y$. In cases where there is no $i$ such that $s_i(x) \approx 1$, a number of functions are dominant in defining the relationship between $x$ and $y$. Since $s(x)$ has a strong inductive bias towards 1 or 0, the regions with multiple dominant functions are transitory and often the areas of interest. Therefore, we can interpret how the change surface develops and where different regimes dominate by evaluating each $s(x)$ over the input domain.

\begin{figure}[h]
\centering
 \includegraphics[width=0.5\textwidth]{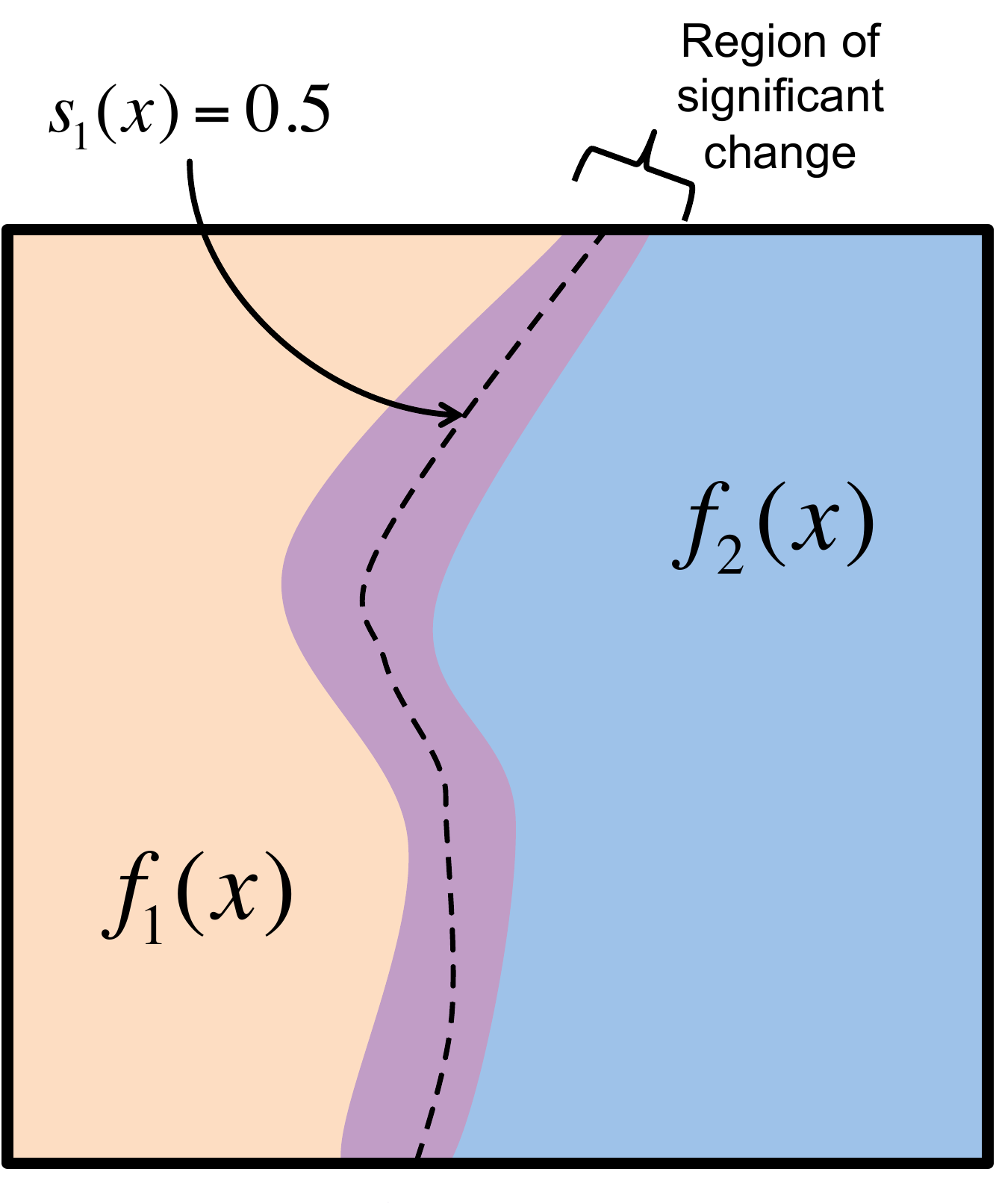}
\caption{Two-dimensional depiction of the change surface model where $f_1(x)$ is drawn in orange and $f_2(x)$ is drawn in blue. The region in purple depicts an area of transition between the two functions. The dashed line represents the domain where $s_1(x)=0.5$.}
\label{fig:CS_cartoon}
\end{figure}
Figure \ref{fig:CS_cartoon} depicts a two-dimensional change surface model where latent $f_1(x)$ is drawn in orange and latent $f_2(x)$ is drawn in blue. In those areas the first warping function, $s_1(x)$, is nearly 1 and 0 respectively. The region in purple depicts an area of transition between the two functions. We would expect that $s_1(x)\approx0.5$ in this region since both latent functions are active.
  
In many applications we can imagine that a latent background function, $f_0(x)$, exists that is common to all data regimes. One could reparametrize the model in Eq.~\eqref{eq:y_CS} by letting each latent regime be a sum of two functions: $f_0(x) + f_i(x)$. Thus each regime compartmentalizes into $f_0(x)$, a common background function, and $f_i(x)$, a regime-specific latent function. This provides a generalized change surface model,
\begin{align}
y(x) = f_0(x) + s_1(x)f_1(x) + \dots + s_r(x)f_r(x) + \epsilon(x) 
\label{eqn: canonicalnew}.
\end{align}

Change surfaces can be considered particular types of \emph{adaptive} mixture models \citep[e.g.,][]{wilson2011gaussian}, where $s(x)$ are mixture weights in a simplex that have a strong inductive bias towards discretization. There are multiple ways to induce this bias towards discretization. For example, one can choose warping functions $s(x)$ which have sharp transitions between 0 and 1, such as the logistic sigmoid function. With multiple functions, $r\geq2$, we can also explicitly penalize the warping functions from having similar values. Since each of these warping functions are constrained to be in $[0,1]$ this penalty would tend move their values towards 0 or 1. More generally, in the case of multiple functional regimes, we can penalize $s(x)$ from being far from $\{0,1\}$. For example, we could place a prior over $s(x)$ with a heavy weight on 1 and 0. 

The flexibility of $s(x)$ defines the complexity of the change surface. In the simplest case, $x_i \in \mathbb{R}^1, s(x) \in \{0,1\}$, and the change surface reduces to a univariate changepoint used in much of the changepoint literature. Alternatively, if we consider $x\in \mathbb{R}^1, s(x) = \sigma(x)$ the change surface is a smooth univariate changepoint with a fixed rate of change. Such a model only permits a monotonic rate of change and single changepoint.

\paragraph{Comparison to changepoint models:}

\begin{figure}[h]
\centering
 \includegraphics[width=1.0\textwidth]{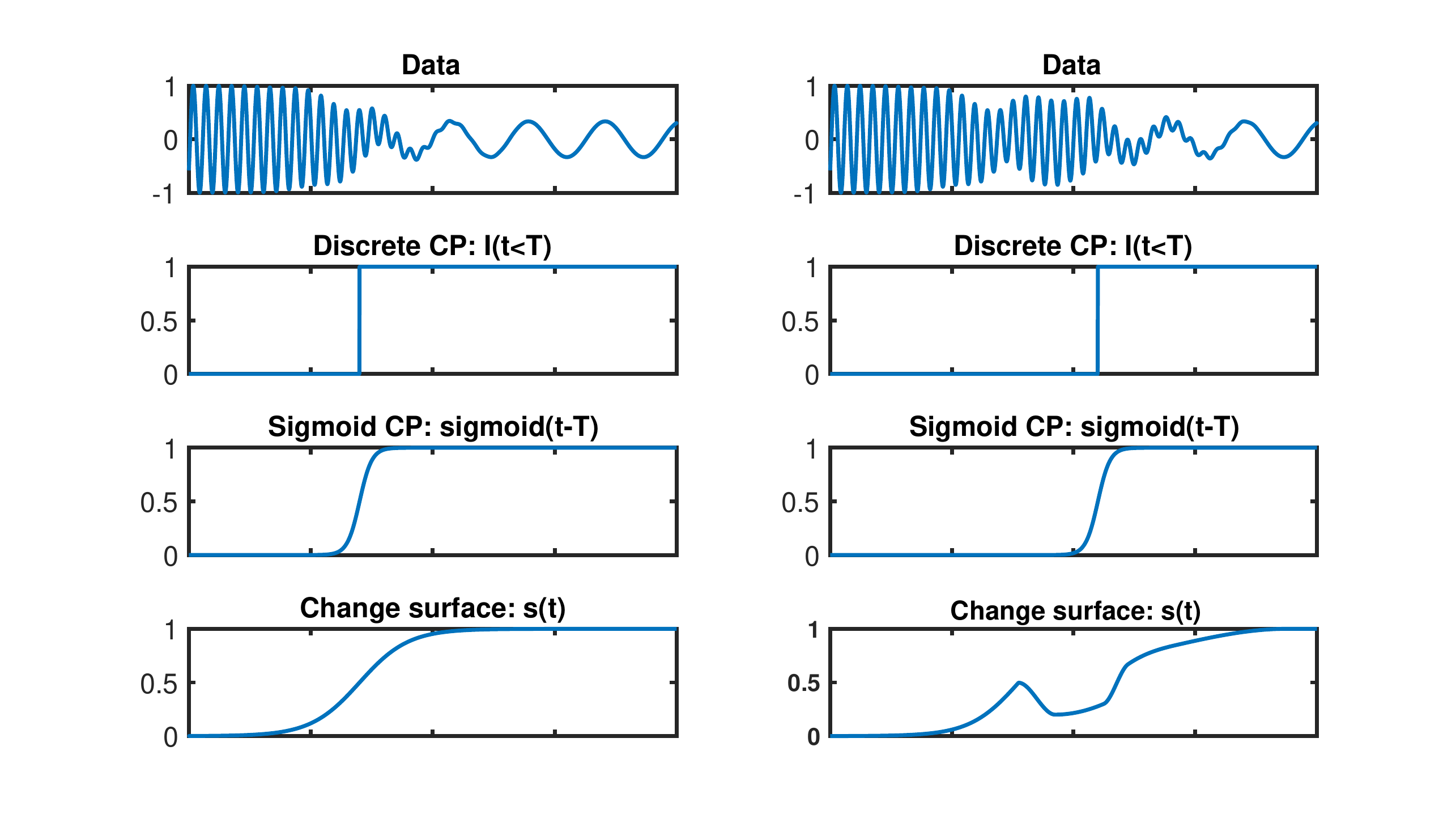}
\caption{Unidimensional comparison of changepoint and change surface methods. In each column, the top plot shows unidimensional data with a clear change between two sinusoids. The subsequent plots represent the warping functions of a discrete changepoint, sigmoid changepoint, and change surface model.}
\label{fig:CS_vs_CP}
\end{figure}

We illustrate the difference between the warping functions, $s(x)$, of a change surface model and standard changepoint methods in Figure \ref{fig:CS_vs_CP}. The top plot shows unidimensional data with a clear change between two sinusoids. The subsequent plots represent the changes modeled in a discrete changepoint, sigmoid changepoint, and change surface model respectively. The changepoint model can only identify a change at a point in time, and the sigmoid changepoint is a special case of a change surface constrained to a fixed rate of change. However, a general change surface can model gradual changes as well as non-monotonic changes, providing a much richer representation of the data's dynamics, and seamlessly extending to multidimensional data.

Expressive change surfaces consider regimes as overlapping elements in the domain. They can illustrate if certain changes occur more slowly or quickly, vary over particular subpopulations, or change rapidly in certain regions of the input domain. Such insights are not provided by standard changepoint models but are critical for understanding policy interventions or scientific processes. Table \ref{tab:Cp_vs_CS} compares some of the limitations of changepoints with the added flexibility of change surfaces.

\begin{table}[h]
\centering
\caption{Comparison of changepoint limitations to change surface flexibility.}
\label{tab:Cp_vs_CS}
\begin{tabular}{|p{6cm}|p{8cm}|}
\hline
\textbf{Changepoints limited by:}                                       & \textbf{Change surfaces allow for:}                                                                                                              \\
\hline
Considering unidimensional, often temporal-only problems                            & Multidimensional inputs with heterogeneous changes across the input dimensions. Indeed, we apply change surfaces to 3-dimensional, spatio-temporal problems in section \ref{sec:experiments}. \\
\hline
Detecting discrete or near-discrete changes in parameter distribution          & Warping functions, $s(x)$, can be defined flexibly to allow for discrete or continuous changes with variable, and even non-monotonic rates of change. \\
\hline
Not simultaneously modeling the latent functional regimes                      & Learning $s_i(x)$ and $f_i(x)$ in Equation \eqref{eq:y_CS} to simultaneously model the change surface and underlying functional regimes.              \\\hline                    
\end{tabular}
\end{table}

Yet the flexibility required by change surfaces as applied to real data sets might seem difficult to instantiate with any particular model. Indeed, machine learning methods are often desired to be expressive, interpretable, and scalable to large data. To address this challenge we introduce the Gaussian Process Change Surface (GPCS) in section \ref{sec:GPCSmodel} which uses Gaussian process priors with flexible kernels to provide rich modeling capability, and a novel scalable inference scheme to permit the method to scale to massive data.

\subsection{Change surface background model}
\label{sec:Change_Surfaces_Half}

In certain applications we are interested in modeling how a change occurs concurrent with a background function which is common to all regimes.  For example, consider urban crime. If a police department staged a prolonged intervention in one sector of the city, we expect that some of the crime dynamics in that sector might change. However, seasonal and other weather-related patterns may remain the same throughout the entire city. In this case we want a model to identify and isolate those general background patterns as well as one or more clearly interpretable functions representing regions of change from the background distribution.

\begin{figure}[h]
\centering
 \includegraphics[width=0.9\textwidth]{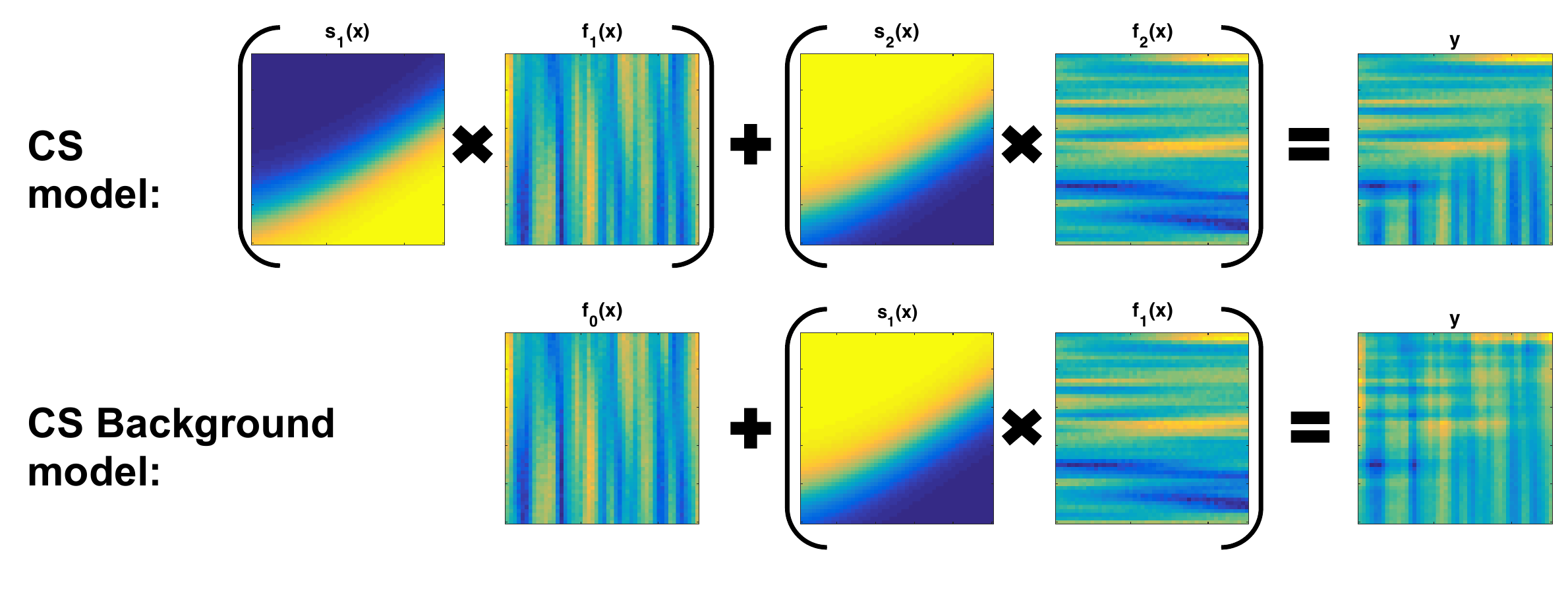}
\caption{Two-dimensional representation of the change surface model (Eq.~\ref{eq:y_CS}) and change surface background model (Eq.~\ref{eq:y_CSI}).}
\label{fig:CS_CSI}
\end{figure}

We can accommodate such a model as a special case of the generalized change surface from Eq.~\eqref{eqn: canonicalnew}. Each latent function is modeled as $f_0(x) + f_i(x)$ where $f_0(x)$ models ``background'' dynamics, and $f_i(x)$  models each \emph{change} function. Since changes do not necessarily persist over the entire domain, we fix $f_r(x)=0$, and allow $\sum_{i=1}^{r-1} s_i(x) \leq 1$. This approach results in the following \emph{change surface background model}:
\begin{equation}
\begin{aligned}
\label{eq:y_CSI}
y(x) &= f_0(x) + s_1(x)f_1(x) + \dots + s_{r-1}(x)f_{r-1}(x) + \epsilon \\
&s.t.\\
&\sum_{i=1}^{r-1} s_i(x) \leq 1\\
&s_i(x) \geq 0
\end{aligned}
\end{equation}
Figure \ref{fig:CS_CSI} presents a two-dimensional representation of the change surface and change surface background models. The data depicted comes from the numerical experiments in section \ref{sec:numerical_exp}.

The explicit decomposition into background and change functions is valuable, for instance, if we wish to model \emph{counterfactuals}: we want to know what the data in a region might look like had there been no change. The decomposition also enables us to interpret the precise effect of each change. Moreover, from a statistical perspective, the decomposition allows us to naturally encode inductive biases into the change surface model, allowing meaningful \emph{a priori} statistical dependencies between each region. In the particular case of $r=2$, the change surface background model has the form $y(x) = f_0(x) + s_1(x)f_1(x)$, where $f_1(x)$ is the only change function modulated by a change surface, $s_1(x)\in[0,1]$. This corresponds to observation studies or natural experiments where a single change is observed in the data. We explore this special case further in our discussion of counterfactual prediction, in section \ref{sec:Counterfactual_Prediction}.

Finally, for any change surface or change surface background model, it is critical that the model not overfit the data due to a proliferation of parameters, which could lead to erroneously detected changes even when no dynamic change is present. We discuss one strategy for preventing overfitting through the use of Gaussian processes in section~\ref{sec:GPCSmodel}.

\subsection{Counterfactual prediction}
\label{sec:Counterfactual_Prediction}

By simultaneously characterizing the change surface, $s(x)$, and the underlying generative functions, $f(x)$, change surface models allow us to ask questions about how the data would have looked had there been only one latent function. In other words, change surface models allow us to consider counterfactual questions. 

For example, in section \ref{sec:disease_exp} we consider measles disease incidence in the United States in the twentieth century. The measles vaccine was introduced in 1963, radically changing the dynamics of disease incidence. Counterfactual studies such as~\citet{van2013contagious} attempt to estimate how many cases of measles there would have been in the absence of the vaccine. To be clear, since change surface models do not consider explicit indicators of an intervention, they do not directly estimate the counterfactual with respect to a particular treatment variable such as vaccination. Instead, they identify and characterize changes in the data generating process that may or may not correspond to a known intervention. The change surface counterfactuals estimate the $y$ values for each functional regime in the absence of the change identified by the change surface model. In cases where the discovered change surface does correspond to a known intervention of interest, domain experts may interpret the change surface predictions as a counterfactual ``what if'' that intervention and any contemporaneous changes in the data generating process (note that we cannot disentangle these causal factors without explicit intervention labels) did not occur.

Counterfactuals are typically studied in econometrics. In observational studies econometricians try to measure the effect of a ``treatment'' over some domain. Econometric models often measure simple features of the intervention effect, such as the expected value of the treatment over the entire domain, also known as the \emph{average treatment effect}.
A nascent body of work considers machine learning approaches to provide counterfactual prediction in complex data \citep{athey2006identification, kay2015inferring, johansson2016learning, hartford2016counterfactual}, as well as richer measures of the intervention effect~\citep{athey2006identification,edTESS}.
Recent work by \cite{schulam2017reliable} uses Gaussian processes for trajectory counterfactual prediction over time. However, these methods generally follow a common framework using the potential outcomes model, which assumes  that each observation is observed with a discrete treatment \citep{rubin2005causal, holland1986statistics}. With discrete treatments a unit, $x$, is either intervened upon or not intervened upon --- there are no partial interventions. For example, in a medical study a patient may be given a vaccination, or given a placebo. Such discretization is similar to a traditional changepoint model where $s(x)\in\{0,1\}$ can only be in one of two states. Yet discrete states prove challenging in practical applications where units may be partially treated or affected through spillover. For example, there may be herd effects in vaccinations whereby a person's neighbor being vaccinated reduces the risk of infection to the person. Certain econometric models attempt to account for partial treatment such as treatment eligibility \citep{abadie2002instrumental}, where partial treatments are induced by defining proportions of the population that could potentially be treated. Yet a model that directly enables and estimates continuous levels of treatment may be more natural in such cases.

\paragraph{Counterfactuals using change surfaces.}
Change surface models enable counterfactual prediction in potentially complex data through the expressive parameterization of the latent functions, $f_1(x),\dots,f_r(x)$. Determining the individual function value $f_i(x)$ over the input domain is equivalent to determining the counterfactual of $f_i(x)$ in the absence of all other latent functions. We can compute counterfactual estimates for latent functions in either the regular change surface model or the change surface background model. In the latter case, if $r=2$ recall that the model takes the form $y = f_0(x) + s_1(x) f_1(x) + \epsilon$. Determining the counterfactual for $f_0(x)$ provides an estimate for the data without the detected change, while the counterfactual for $f_1(x)$ estimates the effect of that change across the entire regime.

\begin{figure}[h]
\centering
 \includegraphics[width=0.7\textwidth]{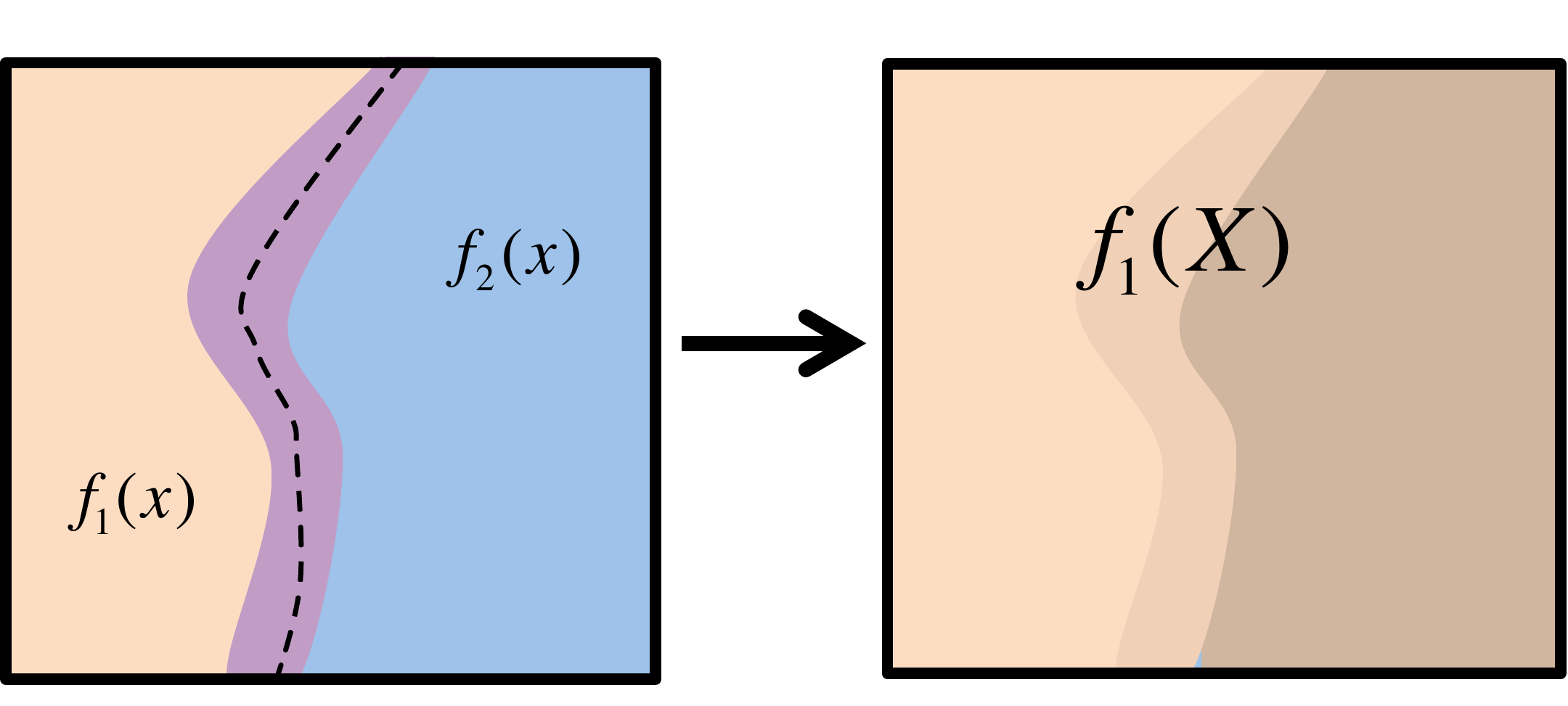}
\caption{Two-dimensional depiction of change surface counterfactual prediction. The left panel illustrates the change surface of Figure~\ref{fig:CS_cartoon}. The right image depicts the counterfactual of $f_1$ over the entire domain, $X$, representing what the observed data could look like in the \emph{absence} of an intervention. The darker shading of the picture depicts larger posterior uncertainty.}
\label{fig:CS_inter_cartoon}
\end{figure}
For example, Figure \ref{fig:CS_inter_cartoon} depicts the counterfactual of $f_1$ from Figure \ref{fig:CS_cartoon}, where $f_1$ is predicted over the entire regime, $X$. The darker shading of the picture depicts larger posterior uncertainty. As we move toward the right portion of the plot, away from data regions where $f_1$ was active, we have greater uncertainty in our counterfactual predictions.

Computing counterfactuals for each $f_i(x)$ provides insight into the effect of a change on the various regimes. When combined with domain expertise, these models may also be useful for estimating the treatment effect of specific variables. Additionally, given a Bayesian formulation of the change surface, such as that proposed in section \ref{sec:GPCSmodel}, we can compute the full posterior distribution over the counterfactual prediction rather than just a point estimate. Finally, since change surfaces model all data points as a combination of latent functions, we do not assume that observed data comes from a particular treatment or control. Rather we learn the contribution of each functional regime to each data point.

Some simple changepoint models could, in theory, provide the ability for counterfactual prediction between regimes. But since changepoint models consider each regime either completely or nearly independently of other regimes, there is no information shared between regimes. This lack of information sharing across regimes makes accurate counterfactual prediction challenging without strong assumptions about the data generating process. Indeed, to our knowledge there is no previous literature using changepoint models for counterfactual prediction. 

\paragraph{Assumptions in change surface counterfactuals:} 

Change surface models identify changing data dynamics without explicitly considering intervention labels. Instead, counterfactuals of the functional regimes are computed with respect to the change surface labels, $s(x)$. Thus these counterfactuals estimate the value of functional regimes in the absence of those changes but do not necessarily represent counterfactual estimates of any particular variable. The interpretation of these counterfactuals as estimates for each functional regime in the absence of a specific known intervention requires identification of the correct change surface, i.e.: 
\begin{itemize}
    \item The intervention induces a change in the data generation process that cannot be modeled with a single latent functional regime.
    \item The magnitude of the change is large enough to be detected.
    \item The change surface model is sufficiently flexible to accurately characterize this change.
    \item The change surface model does not overfit the data to erroneously identify a change.
\end{itemize}
Moreover, the resulting counterfactual estimates do not rule out the possibility that other changes in the data generating process occur contemporaneously with the intervention of interest.  As such, these counterfactuals are most naturally interpreted as estimating what the data would look like in the absence of the intervention and any other contemporaneous changes. Disentangling these multiple potential causal factors would require additional data about both the intervention and other potential causes.

Change surface counterfactual predictions can provide immense value in practical settings. Although in some datasets explicit intervention labels are available, many observational datasets do not have such labels. Learning a change surface effectively provides a real-valued label that can be used to predict counterfactuals. Even when the approximate boundaries of an intervention are known, change surface modeling can still provide an important advantage since the intervention labels may not capture the true complexity of the data. For example, knowing the date that the measles vaccine was introduced does not account for regional variation in vaccine distribution and uptake (see section~\ref{sec:disease_exp}). Both observational studies and randomized control trials suffer from partial treatment or spillover, where an intervention on one agent or region secondarily affects a non-intervened agent or region. For example, increasing policing in one area of the city may displace crime from the intervened region to other areas of the city \citep{verbitsky2012epidemiologic}.  This effect violates the Stable Unit Treatment Value Assumption, which is the basis for many estimation techniques in economics \citep{rubin1986comment}. By using the assumed boundaries of an intervention as a prior over $s(x)$, a change surface model can discover if, and where, spillover occurs. This spillover will be captured as a non-discrete change and can aid both in interpretability of the results and counterfactual prediction. In all these cases change surface counterfactuals may lead to more believable counterfactual predictions by using a real valued change surface to directly model spillover and interventions.

\section{Gaussian Process Change Surfaces (GPCS)}
\label{sec:GPCSmodel}
We exemplify the general concept of change surfaces using Gaussian processes \citep[e.g.,][]{rasmussen2006gaussian}. We emphasize that our change surface formulations from section~\ref{sec:Change_Surfaces} are not limited to a certain class of models. Yet Gaussian processes offer a compelling instantiation of change surfaces since they can flexibly model non-linear functions, seamlessly extend to multidimensional and irregularly sampled data, and provide naturally interpretable parameters. Perhaps most importantly, due to the Bayesian Occam's Razor principle~\citep{rasmussen2001occam, mackay2003information, rasmussen2006gaussian, wilson2014fast}, Gaussian processes do not in general overfit the data, and extraneous model components are automatically pruned. Indeed, even though we develop a rich change surface model with multiple mixture parameters, our results below demonstrate that the model does not spuriously identify change surfaces in data.

Gaussian processes have been previously used for nonparametric changepoint modeling. \citet{saatcci2010gaussian} extend the sequential Bayesian Online Changepoint Detection algorithm \citep{adams2007bayesian} by using a Gaussian process to model temporal covariance  within a particular regime. Similarly, \citet{garnett2009sequential} provide Gaussian processes for sequential changepoint detection with mutually exclusive regimes. Moreover, \citet{keshavarz2018optimal} prove asymptotic convergence bounds for a class of Gaussian process changepoint detection but are restricted to considering a single abrupt change in one-dimensional data. Focusing on anomaly detection, \citet{reece2015anomaly} develop a non-stationary kernel that could conceivably be used to model a changepoint in covariance structure. However, as with most of the changepoint models discussed in section \ref{sec:introduction}, these models all focus on discrete changepoints, where regimes defined by distinct Gaussian processes change instantaneously.

A small collection of pioneering work has briefly considered the possibility of Gaussian processes with sigmoid changepoints \citep{wilson2014covariance,lloyd2014automatic}. Yet these models rely on sigmoid transformations of linear functions which are restricted to fixed rates of change, and are demonstrated exclusively on small, one-dimensional time series data. They cannot expressively characterize non-linear changes or feasibly operate on large multidimensional data.

The limitations of these models reflect a common criticism that Gaussian processes are unable to convincingly respond to changes in covariance structure. We propose addressing this deficiency by modeling change surfaces with Gaussian processes. Thus our work both demonstrates a generalization of changepoint models and an enhancement to the expressive power of Gaussian processes.

\subsection{Gaussian processes overview}
\label{sec:GPs}

We provide a brief review of Gaussian processes.  More detail can be found in \citet{rasmussen2006gaussian}, \citet{scholkopf2002learning}, and \citet{mackay1998introduction}.

Consider data, $(x,y)$, as in section \ref{sec:Change_Surfaces}, where $x = \{x_1,\dots,x_n\}, x_i \in \mathbb{R}^D$, are inputs or covariates, and $y = \{y_1,\dots,y_n\}$, $y_i \in \mathbb{R}$ are outputs or response variables indexed by $x$. We assume that $y$ is generated from $x$ by a latent function with a Gaussian process prior (GP) and Gaussian noise.  In particular,
\begin{align}
y &= f(x) + \epsilon \\
f(x) &\sim \mathcal{GP}(\mu(x), k(x,x'))\\
\epsilon &\sim \mathcal{N}(0, \sigma_\epsilon^2)
\end{align}

A Gaussian process is a nonparametric prior over functions completely specified by mean and covariance functions. The mean function, $\mu(x)$, is the prior expectation of $f(x)$, while the covariance function, $k(x,x')$, is a positive semidefinite kernel that defines the covariance between function values $f(x)$ and $f(x')$.
\begin{align}
\mu(x) &= \mathbb{E}[f(x)]\\ 
k(x,x') &= \text{cov}(f(x), f(x'))
\end{align}
Any finite collection of function values is normally distributed $[f(x_1) ... f(x_p)] \sim \mathcal{N}(\mu(x), K)$ where $p\times p$ matrix $K_{i,j} = k(x_i,x_j)$. Thus we can draw samples from a Gaussian process at a finite set of points by sampling from a multivariate Gaussian distribution. In this paper we generally consider $\mu(x)=0$ and concentrate on the covariance function. The choice of kernel is particularly important in Gaussian process applications since the kernel defines the types of correlations encoded in the Gaussian process. For example, a common kernel choice is a Radial Basis Function (RBF), also known as a Gaussian kernel,
\begin{eqnarray}
\label{eq:RBF}
k(x,x') =  s^2 \exp[-(x-x')^T V^{-1}(x-x') / 2]
\end{eqnarray}
where $s^2$ is the signal variance and $V$ is a diagonal matrix of bandwidths. The RBF kernel implies that nearby values are more highly correlated. While this may be true in many applications, it would be inappropriate for data with significant periodicity. In such cases a periodic kernel would be more fitting. We consider more expressive kernel representations in section \ref{sec:kernel}. This formulation of Gaussian processes naturally accommodates inputs $x$ of arbitrary dimensionality. 

\paragraph{Prediction with Gaussian processes}
Given a set of kernel hyperparameters, $\theta$, and data, $(x,y)$, we can derive a closed form expression for the predictive distribution of $f(x^*)$ evaluated at points $x^*$,
\begin{equation}
\begin{split}
f(x^*) | \theta,x,y,x^* \sim \mathcal{N}\Big(& k(x^*, x) [k(x,x) + \sigma_\epsilon^2I]^{-1}(y-\mu(x)) +\mu(x^*),\\
&k(x^*,x^*) - k(x^*,x) [k(x,x) + \sigma_\epsilon^2I]^{-1} k(x,x^*) \Big)
\end{split}
\end{equation}
The predictive distribution provides posterior mean and variance estimates that can be used to define Bayesian credible sets. Thus Gaussian process prediction is useful both for estimating the value of a function at new points, $x^*$, and for deriving a function's distribution in the domain, $x$, for which we have data.

\paragraph{Learning Gaussian process hyperparameters}
In order to learn kernel hyperparameters we often desire to optimize the marginal likelihood of the data conditioned on the kernel hyperparameters, $\theta$, and inputs, $x$.
\begin{eqnarray}
p(y|\theta, x) = \int p(y|f,x)p(f|\theta) df
\end{eqnarray}
Thus we choose the kernel which maximizes the likelihood that the observed data is generated by the Gaussian process prior with hyperparameters $\theta$. In the case of a Gaussian observation model we can express the log marginal likelihood as,
\begin{equation}
\begin{aligned}
\label{eq:GPloglik}
\log p(y|\theta,x) = - \frac{1}{2}\log|K + \sigma_\epsilon^2 I| - \frac{1}{2} (y-\mu(x))^T(K + \sigma_\epsilon^2 I)^{-1} (y-\mu(x)) + \text{constant}
\end{aligned}
\end{equation}
However, solving linear systems and log determinants involving the $n \times n$ covariance matrix $K$ which incurs $\mathcal{O}(n^3)$ computations and $\mathcal{O}(n^2)$ memory, for $n$ training points, using standard approaches based on the Cholesky decomposition \citep{rasmussen2006gaussian}. These computational requires are prohibitive for many applications, particularly in public policy --- the focus of this paper --- where it is normal to have more than  few thousand training points. Accordingly, we develop alternative scalable inference procedures, presented in section \ref{sec:inference}, which enable tractability on much larger datasets.

\subsection{Model specification}
\label{sec:changepoint_model}

Change surface data consists of latent functions $f_1,\dots,f_r$ defining $r$ regimes in the data.  The change surface defines the transitions between these functions.  We could initially consider an input-dependent mixture model such as in \citet{wilson2011gaussian},
\begin{eqnarray}
\label{eq:GPRN}
y(x) = w_1(x)f_1(x) + \dots + w_r(x)f_r(x) + \epsilon 
\end{eqnarray}
where the weighting functions, $w_i(x): \mathbb{R}^D \rightarrow \mathbb{R}^1$, describe the mixing proportions over the input domain. However, for data with changing regimes we are particularly interested in latent functions that exhibit some amount of mutual exclusivity. 

We induce this partial discretization with $\sigma(z): \mathbb{R}^r \rightarrow [0,1]^r$. These functions have support over the entire real line, but a range in $[0,1]$ and concentrated towards $0$ and $1$.  Thus, each $w_i(x)$ in Eq.~\eqref{eq:GPRN} becomes $\sigma_i(w(x))$, where $w(x)=[w_1(x),..,w_r(x)]$.  Additionally, we choose $\sigma(z)$ such that it produces a convex combination over the weighting functions, $\sum_{i=1}^r \sigma_i(w(x)) = 1$. In this way, each $w_i(x)$ defines the strength of latent $f_i$ over the domain, while $\sigma(z)$ normalizes these weights to induce weak mutual exclusivity. Thus considering the general model of change surfaces in Eq.~\eqref{eq:y_CS} we define each warping function as $s_i(x) = \sigma_i(w(x))$.

A natural choice for flexible change surfaces is to let $\sigma(z)$ be the softmax function. In this way the change surface can approximate a Heaviside step function, corresponding to the sharp transitions of standard changepoints, or more gradual changes. For $r$ latent functions, the resulting warping function is:
\begin{eqnarray}
\label{eq:softmax}
s_i(x) = \sigma_i(w(x)) = \text{softmax}(w(x))_i = \frac{\exp(w_i(x))}{\sum_{j=1}^r \exp(w_j(x))}
\end{eqnarray}
The Gaussian process change surface (GPCS) model is thus
\begin{eqnarray}
\label{eq:y_GP_changepoint}
y(x) = \sigma_1(w(x))f_1(x) + \dots + \sigma_r(w(x))f_r(x) + \epsilon
\end{eqnarray}
where each $f_i$ is drawn from a Gaussian process.  Importantly, we expect that each Gaussian process, $f_i(x)$, will have different hyperparameter values corresponding to different dynamics in the various regimes.

Since a sum of Gaussian processes is a Gaussian process, we can re-write Eq.~\eqref{eq:y_GP_changepoint} as $y(x) = f(x) + \epsilon$, where $f(x)$ has a single Gaussian process prior with covariance function,
\begin{equation}
\begin{aligned}
\label{eq:k_GP_changepoint}
k(x,x') = \sigma_1(w(x))k_1(x,x')\sigma_1(w(x')) +   \dots + \sigma_r(w(x))k_r(x,x')\sigma_r(w(x'))
\end{aligned}
\end{equation}
In this form we can see that $\sigma_1(w(x))\dots\sigma_r(w(x))$ induce non-stationarity since they are dependent on the input $x$. Thus, even if we use stationary kernels for all $k_i$, GPCS observations follow a Gaussian process with a flexible, non-stationary kernel.

\subsubsection{Design choices for $w(x)$}
\label{sec:wx}

The functional form of $w(x)$ determines how changes can occur in the data, and how many can occur. For example, a linear parametric weighting function,
\begin{eqnarray}
\label{eq:w_linear}
w(x) = \beta_0 + \beta_1^T x
\end{eqnarray}
only permits a single linear change surface in the data. Yet even this simple model is more expressive than discrete changepoints since it permits flexibility in the rate of change and extends to change regions in $\mathbb{R}^D$.

In order to develop a general framework, we introduce a flexible $w(x)$ that is formed as a finite sum of Random Kitchen Sink (RKS) features which map the $D$ dimensional input $x$ to an $m$ dimensional feature space. We use RKS features from a Fourier basis expansion with Gaussian parameters and employ marginal likelihood optimization to learn the parameters of this expansion. Similar expansions have been used to efficiently approximate flexible non-parametric Gaussian processes \citep{lazaro2010sparse, rahimi2007random}.

Using $m$ RKS features, $w(x)$ is defined as,
\begin{equation}
w(x) = \sum_{i=1}^m a_i \cos(\omega_i^T x_i + b_i)
\end{equation}
where we initially sample,
\begin{align}
a_i &\sim \mathcal{N}(0, \frac{\sigma_0}{m}I)\\
\omega_i &\sim \mathcal{N}(0, \frac{1}{4\pi^2}\Lambda^{-1}) \\
b_i &\sim \mbox{Uniform}(0, 2\pi)
\end{align}
Initialization of hyperparameters $\sigma_0$ and diagonal matrix of length-scales, $\Lambda = \mbox{diag}(l_1^2,\dots,l_D^2)$, is discussed in section \ref{sec:initialization}.

Experts with domain knowledge can specify a parametric form for $w(x)$ other than RKS features. Such specification can be advantageous, requiring relatively few, highly interpretable parameters to optimize. For example, in an industrial setting where we are modeling failure of parts in a factory we could define $w(x)$ such that it was monotonically increasing since machine parts do not self-repair. This bias could take the form of a linear function as in Equation \eqref{eq:w_linear}. Note that since parameters are learned from data, the functional form of $w(x)$ does not require prior knowledge about if or where changes occur. 

\subsubsection{Kernel specification}
\label{sec:kernel}

Each latent function is specified by a kernel with its own set of hyperparameters. By design, each $k_i$ may be of a different form. For example, one function may have a Mat\'ern kernel, another a periodic kernel, and a third an exponential kernel. Such specification is useful when domain knowledge provides insight into the covariance structure of the various regimes. 

In order to maintain maximal generality and expressivity, we develop GPCS using multidimensional spectral mixture kernels \citep{wilson2013gaussian} where $x\in \mathbb{R}^D$.
\begin{equation}
\label{SM_kernel}
k_{\text{SM}}(x,x') = \sum_{q=1}^Q \omega_q \cos(2\pi (x-x')^T\mu_q) \prod_{d=1}^D\exp(-2\pi^2 (x^{(d)} - x'^{(d)})^2 v_q^{(d)})
\end{equation}
This kernel is derived via spectral densities that are scale-location mixtures of $Q$ Gaussians. Each component in this mixture has mean $\mu_q\in \mathbb{R}^D$, covariance matrix $\text{diag}(v_q^{(1)},...,v_q^{(D)})$, and signal variance parameter $\omega_q \in \mathbb{R}^1$.
With a sufficiently large $Q$, spectral mixture kernels can approximate any stationary kernel, providing the flexibility to capture complex patterns over multiple dimensions. These kernels have been used in pattern prediction, outperforming complex combinations of standard stationary kernels \citep{wilson2014fast}.  

Previous work on Gaussian processes changepoint modeling has typically been restricted to RBF \citep{saatcci2010gaussian,garnett2009sequential} or exponential kernels \citep{majumdar2005spatio}. However, expressive covariance functions are particularly critical for modelling multidimensional and spatio-temporal data -- a key application for change surfaces -- where structure is often complex and unknown a priori. 

Initializing and training expressive kernels is often challenging.  We propose a practical initialization procedure in section \ref{sec:initialization}, which can be used quite generally to help learn flexible kernels.

\subsubsection{GPCS background model}

Following section \ref{sec:Change_Surfaces_Half} we extend GPCS to the ``GPCS background model.'' For this model we add a latent background function, $f_0(x)$, with an independent Gaussian process prior. Using the same choices for expressive $w(x)$ and covariance functions, we define the GPCS background model as,
\begin{equation}
\begin{aligned}
\label{eq:GPCS-Half}
y(x) &= f_0(x) + \sigma_1(w(x))f_1(x) + \dots + \sigma_{r-1}(w(x))f_{r-1}(x) + \epsilon
\end{aligned}
\end{equation}
Recall that in this model we set $f_r(x)=0$. Additionally, since we continue to enforce $\sum_{i=1}^{r} \sigma_i(w(x)) = 1$, thus $\sum_{i=1}^{r-1} \sigma_i(w(x)) \leq 1$.

This model effectively places different priors on the background and change regions, as opposed to the the standard GPCS model which places the same GP prior on each regime. The different priors in the GPCS background model reflect an intentional inductive bias which could be advantageous in certain domain settings, such as policy interventions, as discussed in section \ref{sec:Change_Surfaces_Half} above.

\subsection{GPCS Counterfactual Prediction}
\label{sec:GPCS_CF}

We consider counterfactuals when using two latent functions in a GPCS, $f_1(x)$ and $f_2(x)$. This two-function setup addresses a typical setting for counterfactual prediction when considering two alternatives. The derivations below can be extended to multiple functional regimes. As discussed above, we note that change surface counterfactuals are only valid with respect to the regimes of the data as identified by GPCS. Subsequent analysis and domain expertise are necessary to make any further claims about the relationship between an identified change surface and some latent intervention.

In counterfactual prediction we wish to infer the value of $f_1(x)$ and $f_2(x)$ in the absence of the other function. Therefore we condition on the observations, $(x,y)$, and GPCS model parameters in order to compute the conditional distribution $p\big([f_1(x), f_2(x)] | y\big)$ from the multivariate Gaussian joint distribution $p\big([f_1(x), f_2(x)], y\big)$. For notational convenience we omit explicit reference to the model parameters in the subsequent derivations but note that all distributions are conditional on these parameters.

To recall, for two latent functions, $f_1(x)$ and $f_2(x)$, GPCS specifies
\begin{align}
y(x) &=\sigma_1 f_1(x) + \sigma_2 f_2(x) + \epsilon \\
\epsilon &\sim \mathcal{N}(0, \sigma_\epsilon^2) \\
f_1(x) &\sim \mathcal{GP}(0,K_1) \\
f_2(x) &\sim \mathcal{GP}(0,K_2)
\end{align}
where for notational simplicity we let $K_1=k_1(x,x')$, $K_2=k_2(x,x')$, $\sigma_1 = \sigma_1(w(x))$, and $\sigma_2 = \sigma_2(w(x))$.

We consider the most general case when we want to predict counterfactuals for both $f_1(x)$ and $f_2(x)$ over the domain $X$. No restrictions are placed over $X$. It can include the entire original domain, parts of the original domain, or different inputs entirely. We concatenate $f(X)$ and $g(X)$ together,
\begin{eqnarray}
u=[f(X), g(X)] \,.
\end{eqnarray}
Since in section \ref{sec:changepoint_model} we assumed that $f_1(x)$ and $f_2(x)$ have independent Gaussian process priors, we know that,
\begin{align}
\label{eq:u_distro}
u \sim \mathcal{N}\Big(0, \begin{bmatrix}
    K_1 & 0  \\
    0 & K_2  \end{bmatrix}   \Big)
\end{align}
Considering the observed data, $y$, we know that $u$ and $y$ are jointly Gaussian,
\begin{align}
\begin{bmatrix}
    u  \\
    y  \end{bmatrix} \sim \mathcal{N}\Big(0, \begin{bmatrix}
    \Sigma_{u,u} & \Sigma_{u,y}  \\
    \Sigma_{u,y}^T & \Sigma_{y,y}  \end{bmatrix}   \Big)
\end{align}
and using multivariate Gaussian identities, we find that $u$ has the conditional Gaussian distribution
\begin{align}
u|y \sim \mathcal{N}\Big( \Sigma_{u,y} \Sigma_{y,y}^{-1} y, \Sigma_{u,u}-\Sigma_{u,y}\Sigma_{y,y}^{-1}\Sigma_{u,y}^T \Big)
\end{align}
Thus in order to derive counterfactuals for both $f(X)$ and $g(X)$ we only need to compute $\Sigma_{u,y}, \Sigma_{y,y}$, and $\Sigma_{u,u}$. Note that with respect to $\Sigma_{u,u}$ we have already derived the covariance structure for $u$ in Equation \eqref{eq:u_distro}.

\paragraph{Computation for $\Sigma_{u,y}$}
In order to compute $\Sigma_{u,y}$, we expand the multiplication noting that $y$ is defined to be a two-function GPCS,
\begin{align}
\Sigma_{u,y} &= E\Huge[uy^T]\\
 &= \mathbb{E}[\begin{bmatrix}
    f_1(x_1)  \\
    ...\\
    f_1(x_n)  \\ 
    f_2(x_1)  \\ 
    ...\\
    f_2(x_n)  \end{bmatrix}\begin{bmatrix}
    \sigma_1(x_1)f_1(x_1) + \sigma_2(x_1)f_2(x_1) + \epsilon  \\
    ...\\
    \sigma_1(x_n)f_1(x_n) + \sigma_2(x_n)f_2(x_n) + \epsilon  \end{bmatrix}^T\Huge]
    \end{align}
Multiplying these elements is assisted by the following identities. Recall that kernels $K_1$ and $K_2$ define the covariance among function values in $f$ and $g$ respectively,
\begin{align}
\mathbb{E}[f_1(x_i)f_1(x_j)] &=k_1(i,j)\\
\mathbb{E}[f_2(x_i)f_2(x_j)] &=k_2(i,j)
\end{align}
Additionally, since $f_1(x)$ and $f_2(x)$ have independent Gaussian process priors, $\mathbb{E}[f_1(x_i)f_2(x_j)]=0$. Furthermore, because $\epsilon$ is distributed with mean zero, $\mathbb{E}[\epsilon_i]=0$. Finally, since $\sigma_1(x)$ and $\sigma_2(x)$ are constant (conditional on hyperparameters) $\mathbb{E}[\sigma_1(x_i)] = \sigma_1(x_i)$ and $\mathbb{E}[\sigma_2(x_i)] = \sigma_2(x_i)$. Thus we can conclude that
\begin{align}
\Sigma_{u,y} &= \begin{bmatrix}
    \sigma_1(x_1)k_1(1,1) & \sigma_1(x_2)k_1(1,2) & ... & \sigma_1(x_n)k_1(1,n) \\
    \sigma_1(x_1)k_1(2,1) & \sigma_1(x_2)k_1(2,2) & ... & \sigma_1(x_n)k_1(2,n) \\
    ...\\
    \sigma_1(x_1)k_1(n,1) & \sigma_1(x_2)k_1(n,2) & ... & \sigma_1(x_n)k_1(n,n) \\
    \sigma_2(x_1)k_2(1,1) & \sigma_2(x_2)k_2(1,2) & ... & \sigma_2(x_n)k_2(1,n) \\
    \sigma_2(x_1)k_2(2,1) & \sigma_2(x_2)k_2(2,2) & ... & \sigma_2(x_n)k_2(2,n) \\
    ...\\
    \sigma_2(x_1)k_2(n,1) & \sigma_2(x_2)k_2(n,2) & ... & \sigma_2(x_n)k_2(n,n)  \end{bmatrix} \\
&= \begin{bmatrix}
K_1 \odot \mathbbm{1}\sigma_1^T \\
K_2 \odot \mathbbm{1}\sigma_2^T
\end{bmatrix}
\end{align}
where $\odot$ is elementwise multiplication.

\paragraph{Computation for $\Sigma_{y,y}$}
The computation for $\Sigma_{y,y}$ is very similar to that of $\Sigma_{u,y}$ so we omit its expansion for the sake of brevity. The slight difference is that we must consider $\mathbb{E}[\epsilon_i\epsilon_i]$ which equals $\sigma_\epsilon^2$.

Thus,
\begin{align}
\Sigma_{y,y} &= E\Huge[yy^T]\\
&= K_1\odot[\sigma_1\sigma_1^T] + K_2\odot[\sigma_2\sigma_2^T] + I_n\sigma_\epsilon^2
\end{align}

\subsubsection{GPCS background model counterfactuals}

The counterfactual derivations above directly apply to the GPCS background model with $r=2$, where $y(x) = f_0(x) + \sigma_1(w(x))f_1(x)$. Recall that as we discussed in section \ref{sec:Change_Surfaces_Half}, this is a special case of the GPCS background model where $f_1(x)$ is an additive change function. In this case, the counterfactual for $f_0(x)$ estimates what would have occurred in the absence of the identified change. The counterfactual for $f_1(x)$ models how the change would have affected the entire domain.

If we let $u=[f_0(X), f_1(X)]$ we can derive counterfactuals for the GPCS background model by setting $\sigma_0=1$ in the equations for $\Sigma_{u,u}$, $\Sigma_{u,y}$, and $\Sigma_{y,y}$ above. Explicitly,
\begin{align}
\Sigma_{u,u} &= \begin{bmatrix}
    K_0 & 0  \\
    0 & K_1  \end{bmatrix}\\
\Sigma_{u,y} &= \begin{bmatrix}
K_0  \\
K_1 \odot \mathbbm{1}\sigma_1^T
\end{bmatrix}\\
\Sigma_{y,y} &= K_0 + K_1\odot[\sigma_1\sigma_1^T] + I_n\sigma_\epsilon^2
\end{align}

\subsection{Scalable inference}
\label{sec:inference}

Analytic optimization and inference for Gaussian processes requires computation of the log marginal likelihood from Eq.~\eqref{eq:GPloglik}. Yet solving linear systems and computing log determinants over $n \times n$ covariance matrices, using standard approaches such as the Cholesky decomposition, requires $O(n^3)$ computations and $O(n^2)$ memory, which is impractical for large datasets. Recent advances in scalable Gaussian processes \citep{wilson2014covariance} have reduced this computational burden by exploiting Kronecker structure under two assumptions: (1) the inputs lie on a grid formed by a Cartesian product, $x \in X = X^{(1)} \times...\times X^{(D)}$; and, (2) the kernel is multiplicative across each dimension. Multiplicative kernels are commonly employed in spatio-temporal Gaussian process modeling \citep{martin1990use,majumdar2005spatio,flaxman2015fast}, corresponding to a soft a priori assumption of independence across input dimensions, without ruling out posterior correlations.  The popular RBF and ARD kernels, for instance, already have this multiplicative structure.  Under these assumptions, the $n\times n$ covariance matrix $K=K_1\otimes \dots \otimes K_D$, where each $K_d$ is $n_d \times n_d$ such that $\prod_1^D n_d = n$.

Using efficient Kronecker algebra, \citet{saatcci2012scalable} shows how one can solve linear systems and compute log determinants in $O(Dn^{\frac{D+1}{D}})$ operations using $O(Dn^{\frac{2}{D}})$ memory. Furthermore, \citet{wilson2014fast} extends the Kronecker methods for incomplete grids.
Yet for additive compositions of kernels, such as those needed for change surface modeling in Eq.~\eqref{eq:k_GP_changepoint}, the resulting sum of matrix Kronecker products does not decompose as a Kronecker product. Thus, the standard Kronecker approaches for scalable inference and learning are inapplicable.  Instead, solving linear systems for the kernel inverse can be efficiently carried out through linear conjugate gradients as in \citet{flaxman2015fast} that only rely on matrix vector multiplications, which can be performed efficiently with sums of Kronecker matrices.

However, there is no exact method for efficient computation of the log determinant of the sum of Kronecker products. Instead, \citet{flaxman2015fast} upper bound the log determinant using the Fiedler bound \citep{fiedler1971bounds} which says that for $n \times n$ Hermitian matrices $A$ and $B$ with sorted eigenvalues $\alpha_1,\dots,\alpha_n$ and $\beta_1,\dots,\beta_n$ respectively,
\begin{eqnarray}
\label{eq:fiedler}
\log(| A + B |) \leq \sum_{i=1}^n \log(\alpha_i + \beta_{n-i+1})
\end{eqnarray}
While efficient, the Fiedler bound does not generalize to more than two matrices. 

\subsubsection{Weyl bound}

In order to achieve scalable computations for an arbitrary additive composition of Kronecker matrices, we propose to bound the log determinant of the sum of multiple covariance matrices using Weyl's inequality \citep{weyl1912asymptotische} which states that for $n\times n$ Hermitian matrices, $M = A + B$, with sorted eigenvalues $\mu_1,\dots,\mu_n$, $\alpha_1,\dots,\alpha_n$, and $\beta_1,\dots,\beta_n$ respectively,
\begin{eqnarray}
\mu_{i+j-1} \leq \alpha_i + \beta_j \quad \forall i,j \geq 1
\end{eqnarray}
Since $\log(| A + B |) = \log(|M|) = \sum_{i=1}^n \log(\mu_i)$ we can bound the log determinant by $\sum_{i+j-1=1}^n \log(\alpha_i + \beta_j)$. Furthermore, we can use the Weyl bound iteratively over pairs of matrices to bound the sum of $r$ covariance matrices $K_1,\dots,K_r$.

As the bound indicates, there is flexibility in the choice of which eigenvalue pair $\{\alpha_i, \beta_j\}$ to use for bounding $\mu_{i+j-1}$. Thus for each eigenvalue, $\mu_k$, we wish to choose $i, j$ that minimizes $\alpha_i + \beta_j$ subject to $k=i+j-1$. One might be tempted to minimize over all possible pairs for each eigenvalue, $\mu_1,\dots,\mu_n$, in order to obtain the tightest bound on the log determinant. Unfortunately, such a procedure requires $O(n^2)$ computations. Instead we explore two possible alternatives:
\begin{enumerate}
\item For each $\mu_{i+j-1}$ we choose the ``middle'' pair, $\{\alpha_i, \beta_j\}$, such that $i = j$ when possible, and $i = j+1$ otherwise. This ``middle'' heuristic requires $O(n)$ computations.
\item We employ a greedy search to choose the minimum of $v$ possible pairs of eigenvalues. Using the previous $i'$ and $j'$, we consider $\{\alpha_i, \beta_j\}$ for all $i=i'-\frac{v}{2},...,i'+\frac{v}{2}$ and the corresponding $j$ values. Setting $v=1$ corresponds to the middle heuristic. Setting $v=n$ corresponds to the exact Weyl bound. The greedy search requires $O(vn)$ computations.
\end{enumerate}
In addition to bounding the sum of kernels, we must also deal with the scaling functions, $\sigma_i(w(x))$. We can rewrite Eq. \eqref{eq:k_GP_changepoint} in matrix notation,
\begin{eqnarray}
\label{eq:k_GP_changepoint_matrix}
K = S_1K_1S_1' +  \dots + S_rK_rS_r'
\end{eqnarray}
where $S_i = \mbox{diag}(\sigma_i(w(x)))$ and $S_i' = \mbox{diag}(\sigma_i(w(x')))$. Employing the bound on eigenvalues of matrix products \citep{bhatia2013matrix},
\begin{eqnarray}
\mbox{sort}(\mbox{eig}(A B)) \leq \mbox{sort}(\mbox{eig}(A)) \mbox{sort}(\mbox{eig}(B))
\end{eqnarray}
we can bound the log determinant of $K$ in Eq.~\eqref{eq:k_GP_changepoint_matrix} with an iterative Weyl approximation over $[\{s_{i,l}  k_{i,l}  s_{i,l}'\}_{l=1}^n]_{i=1}^r$ where $s_{i,l}$, $k_{i,l}$, and $s_{i,l}'$ are the $l^{th}$ largest eigenvalue of $S_i$, $K_i$, and $S_i'$ respectively.

We empirically evaluate the exact Weyl bound, middle heuristic, and greedy search with $v=80$ pairs of eigenvalue indexes to search above and below the previous index. All experiments are evaluated using GPCS with synthetic data generated according to the procedure in section \ref{sec:numerical_exp}. We also compare these results against the Fiedler bound in the case of two kernels.

\begin{figure}[ht]
  \centering
  \begin{minipage}[b]{0.49\textwidth}
    \includegraphics[width=\textwidth]{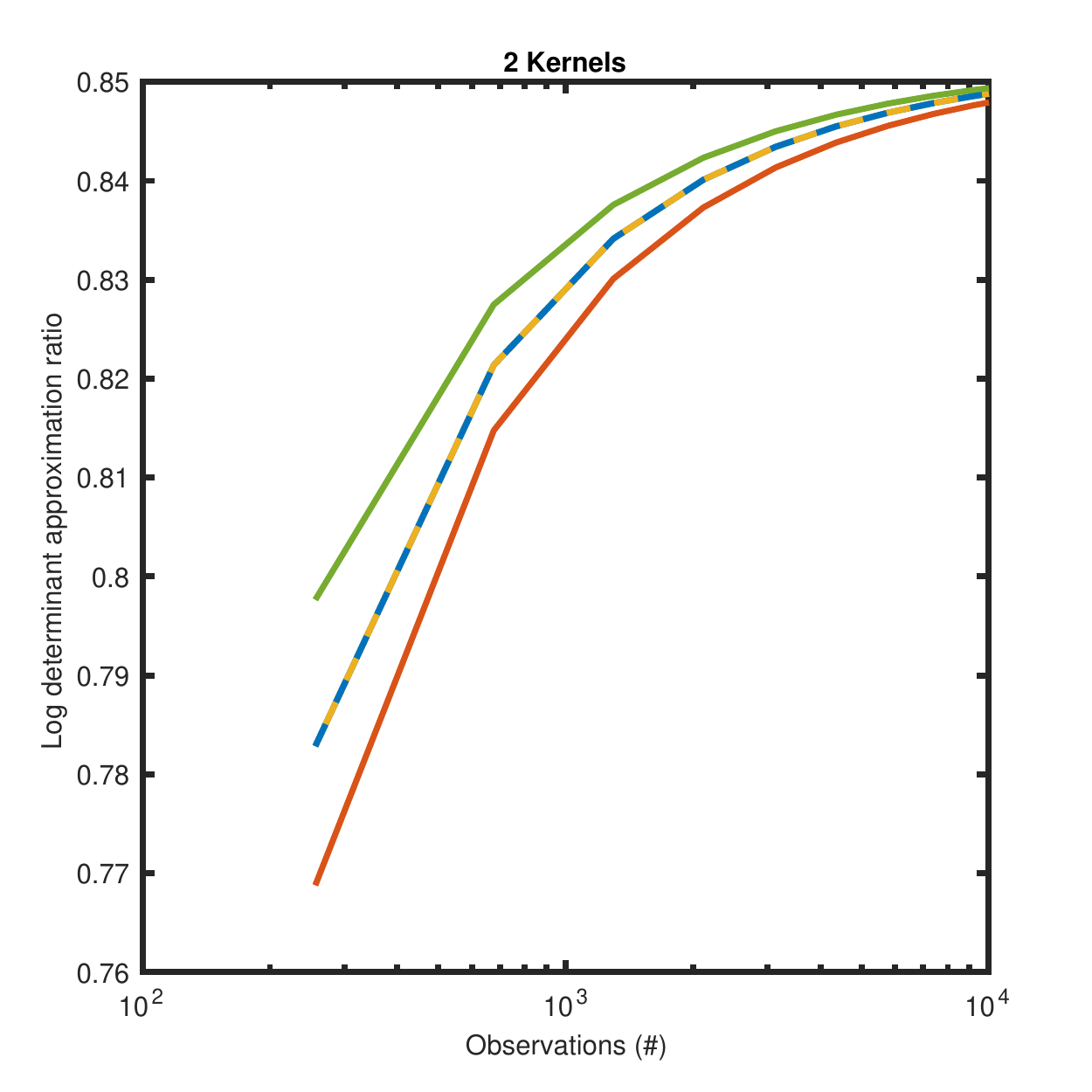}
  \end{minipage}
  \hfill
  \begin{minipage}[b]{0.49\textwidth}
    \includegraphics[width=\textwidth]{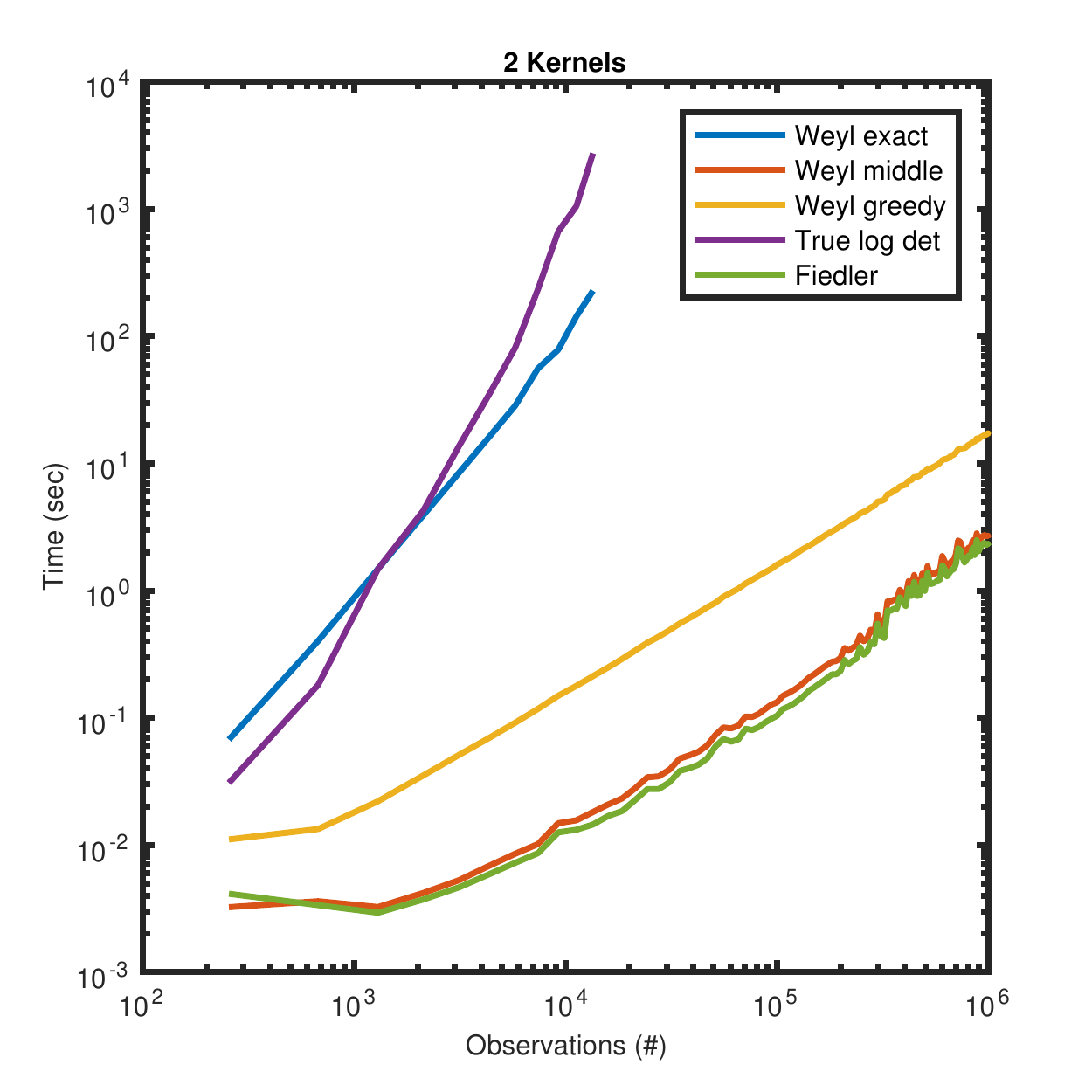}
  \end{minipage}
      \caption{Left plot shows the ratio of log determinant approximations to the true log determinant of two additive kernels. Note that the y-axis is scaled to a relatively narrow band. The dashed line indicates that both the Weyl exact and Weyl greedy method performed similarly. Right plot shows the time to compute each approximation and the true log determinant.}
	\label{fig:weyl_2kernels}
\end{figure}

Figure \ref{fig:weyl_2kernels} depicts the ratio of each approximation to the true log determinant, and the time to compute each approximation over increasing number of observations for two kernels. While the Fiedler approximation is more accurate than any Weyl approach, all approximations perform quite similarly (note the fine grained axis scale) and converge to $\approx 0.85$ of the true log determinant. In terms of computation time, the exact Weyl bound scales poorly with data size as expected. Yet both approximate Weyl bounds scale well. In practice, we use the middle heuristic described above, since it provides the fastest results, nearly equivalent to the Fiedler bound.

\begin{figure}[ht]
  \centering
  \begin{minipage}[b]{0.49\textwidth}
    \includegraphics[width=\textwidth]{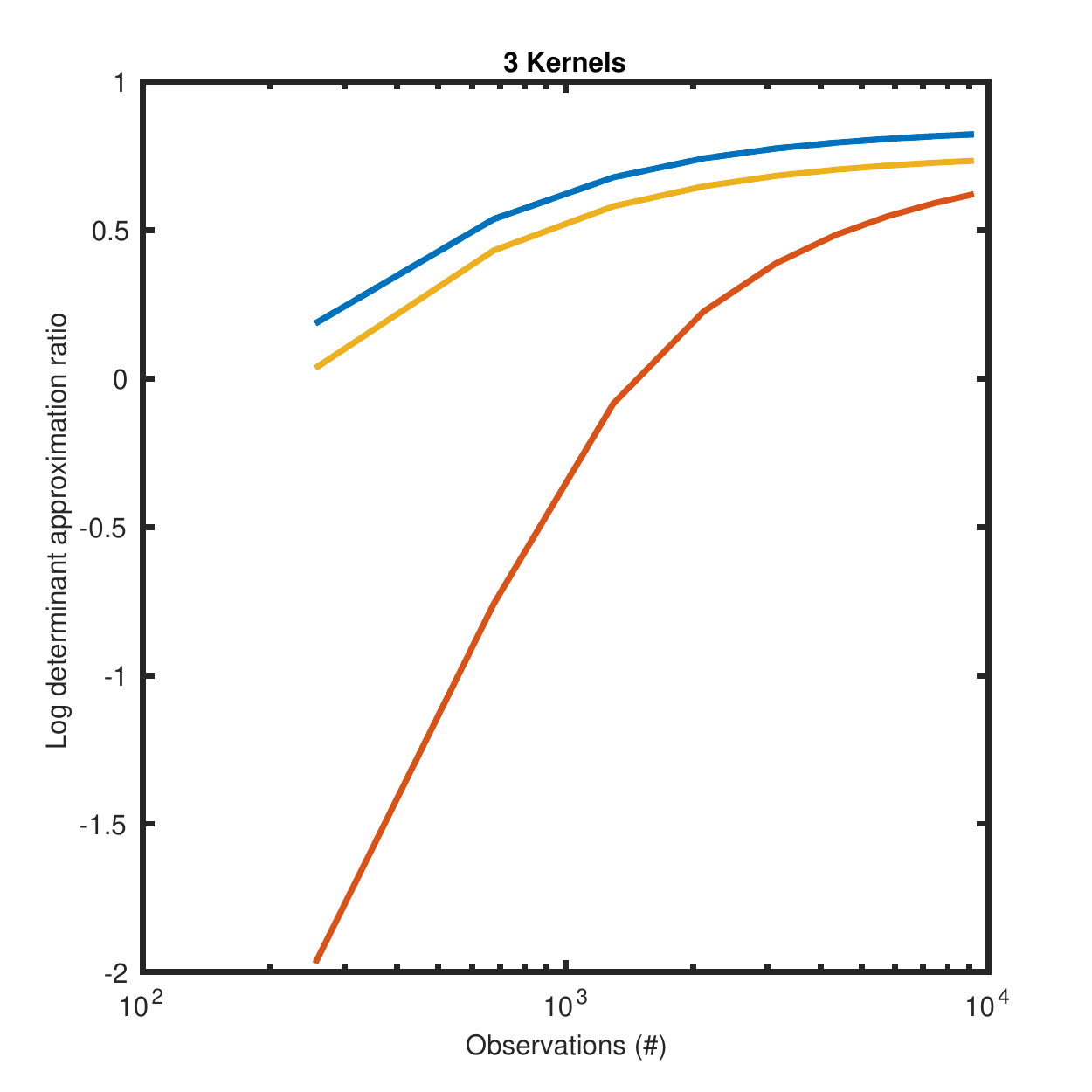}
  \end{minipage}
  \hfill
  \begin{minipage}[b]{0.49\textwidth}
    \includegraphics[width=\textwidth]{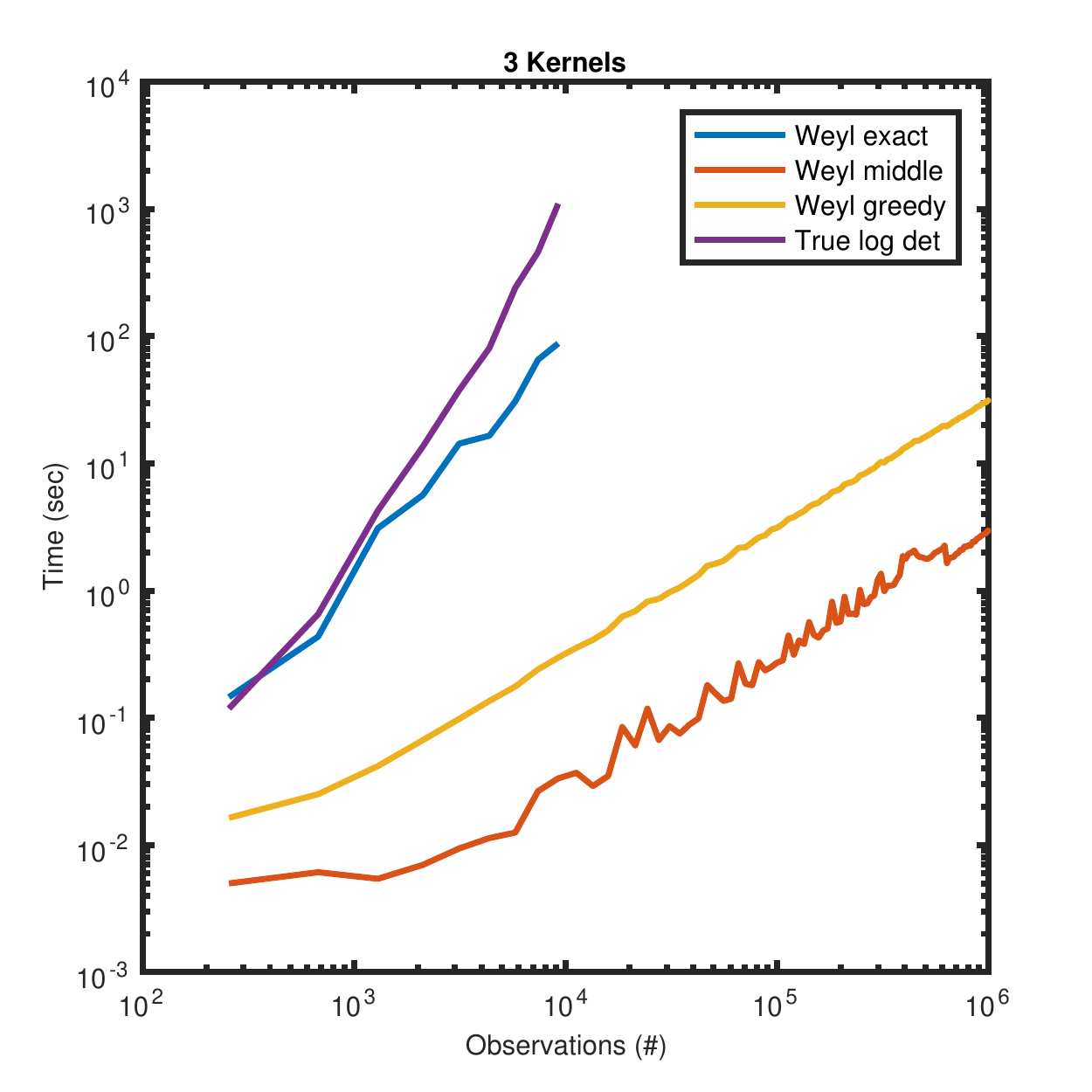}
  \end{minipage}
      \caption{Left plot shows the ratio of approximations to the true log determinant of 3 additive kernels. Note that the y-axis has a much larger scale than in Figure \ref{fig:weyl_2kernels}. Right plot shows the time to compute each approximation and the true log determinant of 3 additive kernels.}
	\label{fig:weyl_3kernels}
\end{figure}

Figure \ref{fig:weyl_3kernels} depicts the same quantities as Figure \ref{fig:weyl_2kernels} but using three additive kernels. Since the Fiedler approximation is only valid for two kernels it is excluded from these plots. While the log determinant approximation ratios are less accurate for small datasets, as the data size increases all Weyl approximations converge to $\approx 0.8$. 

In addition to enabling scalable change surface kernels, the Weyl bound method permits scalable additive kernels in general. When applied to the spatio-temporal domain this yields the first scalable Gaussian process model which is non-separable in space and time.

\subsubsection{Massively Scalable Inference}

We further extend the scalability and flexibility of the Weyl bound method by leveraging a structured kernel interpolation methodology from the KISS-GP framework \citep{wilson2015kernel}. Although many spatiotemporal policy relevant applications naturally have near-grid structure, such as readings over a nearly dense set of latitudes, longitudes, and times, this integration with KISS-GP further relaxes the dependencies on grid assumptions. The resulting approach scales to much larger problems by interpolating data to a smaller, user-defined grid. In particular, with local cubic interpolation, the error in the kernel approximation is upper bounded $O(1/m^3)$ for $m$ latent grid points, and $m$ can be very large because the kernel matrices in this space are structured. These scalable approaches are thus very generally applicable as demonstrated in an extensive range of previously published experiments in \cite{wilson2016deep,wilson2016stochastic} based on these techniques. Additionally, KISS-GP enables the Weyl bound approximation methods to apply to arbitrary, non-grid data.

\begin{figure}[ht]
  \centering
  \begin{minipage}[b]{0.49\textwidth}
    \includegraphics[width=\textwidth]{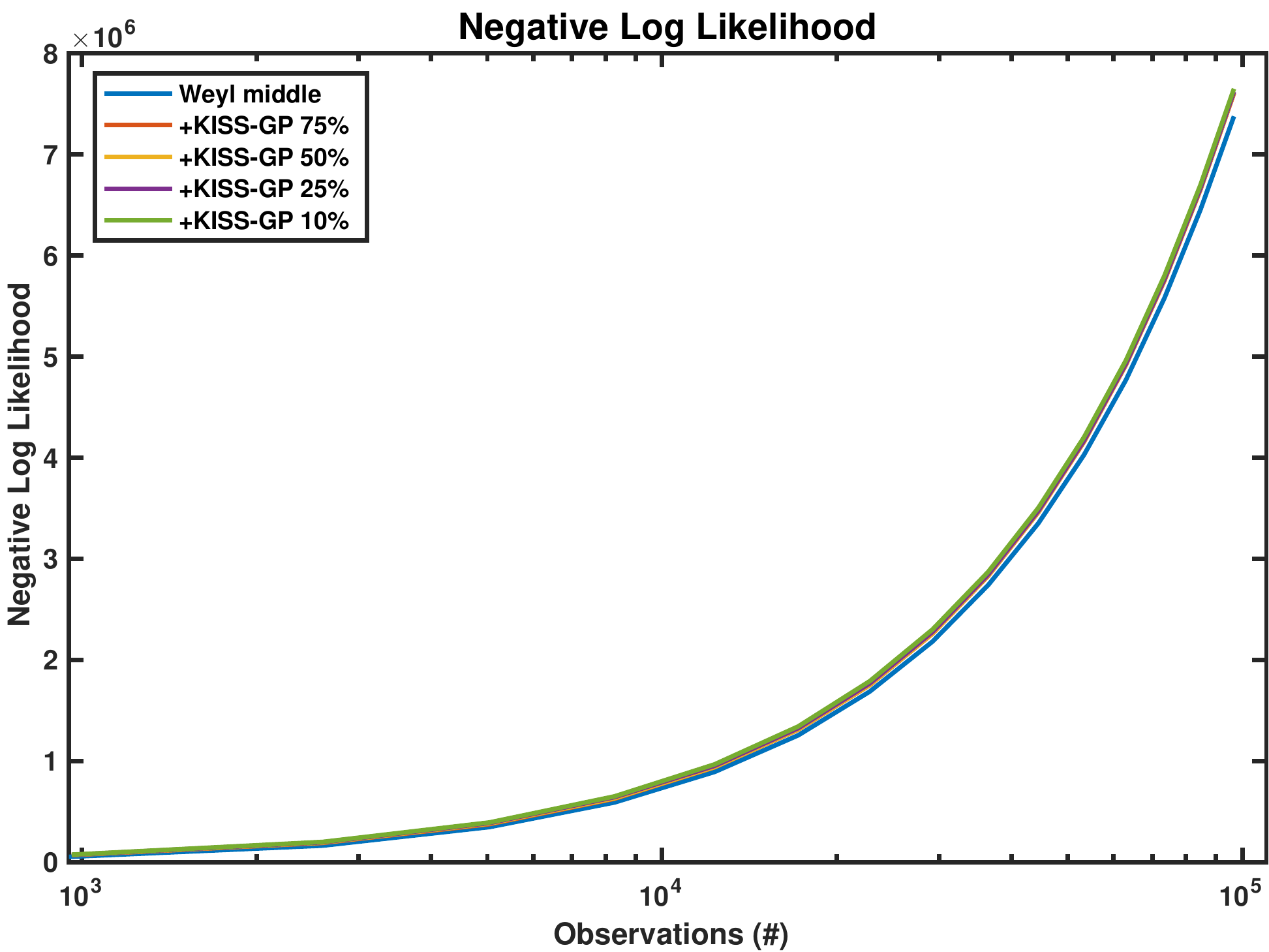}
  \end{minipage}
  \hfill
  \begin{minipage}[b]{0.49\textwidth}
    \includegraphics[width=\textwidth]{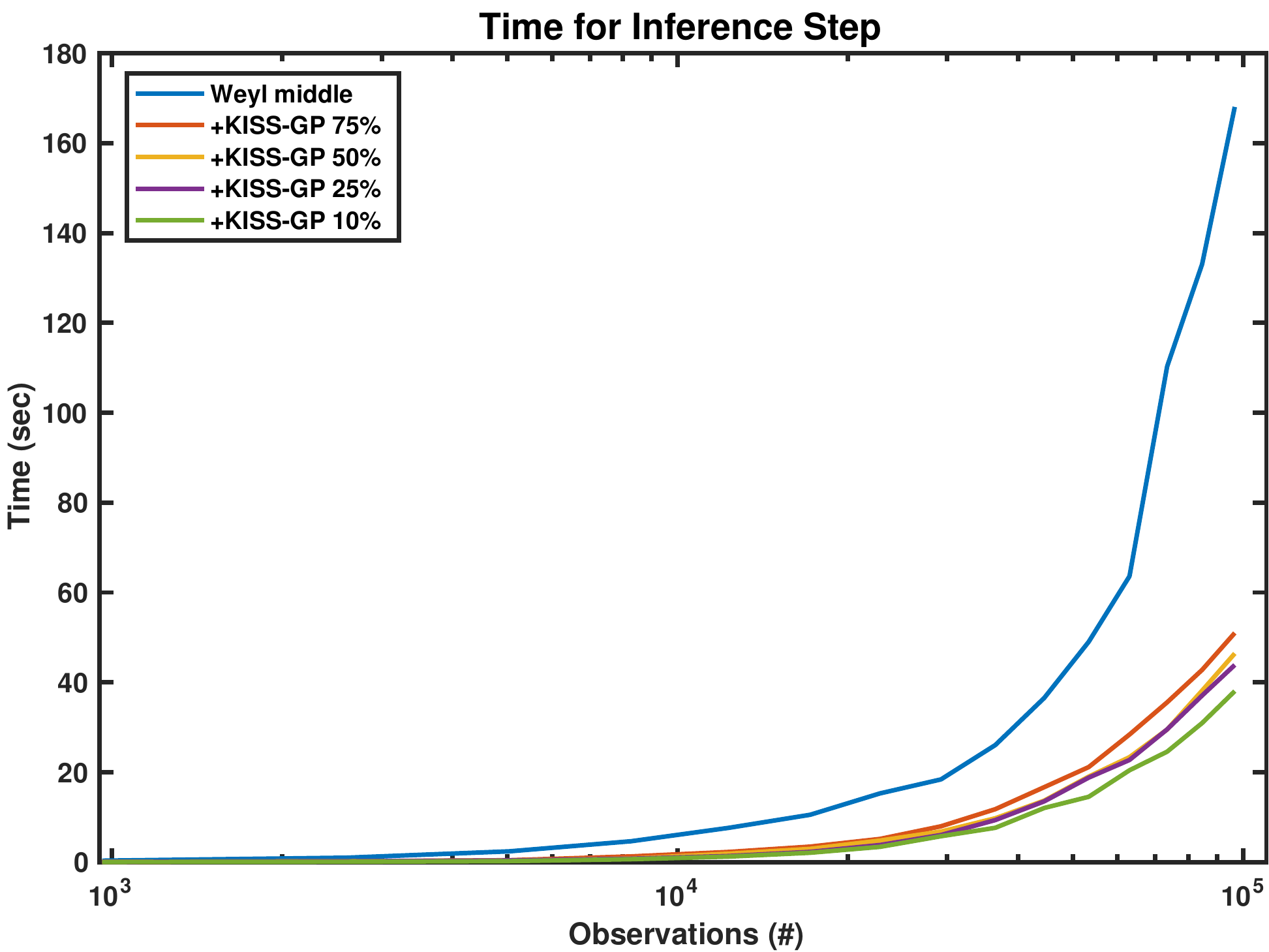}
  \end{minipage}
      \caption{Plots showing negative log likelihood and time for inference on two additive kernels using the Weyl bound on grids of decreasing size. For example, `KISS-GP 75\%' computes the Weyl middle bound on a grid which is 75\% the size of the original grid used to compute the first line.}
	\label{fig:KISSGP}
\end{figure}
We empirically demonstrate the advantages of integration with KISS-GP by evaluating an additive GPCS on the two-dimensional data described above. Although the original data lies on a grid, we use KISS-GP interpolation to compute the negative log likelihood on four grids of increasingly smaller size. Figure \ref{fig:KISSGP} depicts the negative log likelihood and the computation time for these experiments using the Weyl middle heuristic. The plot legend indicates the size of the induced grid size. For example, `KISS-GP 75\%' is 75\% the size of the original grid.  Note that the time and log likelihood scales in Figure \ref{fig:KISSGP} are different from those in Figures \ref{fig:weyl_2kernels} and \ref{fig:weyl_3kernels} since we are now computing full inference steps as opposed to just computing the log determinant. The results indicate that with minimal error in negative log likelihood accuracy we can substantially reduce the time for inference.

\subsection{Initialization}
\label{sec:initialization}

Since GPCS uses flexible spectral mixture kernels, as well as RKS features for the change surface, the parameter space is highly multimodal. Therefore, it is essential to initialize the model hyperparameters appropriately.  Below we present an approach where we first initialize the $w(x)$ RKS features and then use those values in a novel initialization method for the spectral mixture kernels. Like most GP optimization problems, GPCS hyperparameter optimization is non-convex and there are no provable guarantees that the proposed initialization will result in optimal solutions. However, it is  our experience that this initialization procedure works well in practice for the GPCS as well as spectral mixture kernels in general.

To initialize $w(x)$ defined by RKS features we first simplify the change surface model by assuming that each latent function, $f_1,...,f_r$, from Eq. \eqref{eq:y_GP_changepoint} is drawn from a Gaussian process with an RBF kernel. Since RBF kernels have far fewer hyperparameters than spectral mixture kernels, starting with RBF kernels helps our approach find good starting vaglues for $w(x)$. Algorithm \ref{alg:init_wx} provides the procedure for initializing this simplified change surface model. Note that depending on the application domain, a model with latent functions defined by RBF kernels may be sufficient as a terminal model.

\begin{algorithm}
\caption{Initialize RKS $w(x)$ by optimizing a simplified model with RBF kernels}
\label{alg:init_wx}
\begin{algorithmic}[1]
\FOR{$i=1:m_1$}
\STATE Draw $a$, $\omega$, $b$ for RKS features in $w(x)$
\STATE Draw $m_2$ sets of hyperparameter values for RBF kernels, $\{\theta_1,...,\theta_{m_2}\}$
\STATE Choose the best hyperparameter set, $\theta^{(i)} =\texttt{max-likelihood} (\theta_1,...,\theta_{m_2})$
\STATE Partial optimization of $\{a, \omega, b, \theta\} \rightarrow \Theta^{(i)}$
\ENDFOR
\STATE Choose the best set of hyperparameters, $\Theta = \texttt{max-likelihood} (\Theta^{(1)},...,\Theta^{(m_1)})$
\STATE Optimize $\Theta$ until convergence
\end{algorithmic}
\end{algorithm}

In the algorithm, we test multiple possible sets of values for $w(x)$ by drawing the hyperparameters $a$, $\omega$, and $b$ from their respective prior distributions (see section \ref{sec:wx}) $m_1$ number of times. We set reasonable values for hyperparameters in those prior distributions. Specifically, we let $\Lambda = (\frac{\text{range}(x)}{2})^2$, $\sigma_0=\mbox{std}(y)$, and $\sigma_n= \frac{\mbox{mean}(|y|)}{10}$. These choices are similar to those employed in \cite{lazaro2010sparse}.

For each sampled set of $w(x)$ hyperparameters, we sample $m_2$ sets of hyperparameters for the RBF kernels and select the set with the highest marginal likelihood. Then we run an abbreviated optimization procedure over the combined $w(x)$ and RBF hyperparameters and  select the joint set that achieves the highest marginal likelihood. Finally, we optimize  the resulting hyperparameters until convergence.

In order to initialize the spectral mixture kernels, we use the initialized $w(x)$ from above to define the subset $\{x : \sigma_i(w(x)) > 0.5\}$ where each latent function, $f_i$ from Eq. \eqref{eq:y_GP_changepoint}, is dominant. We then take a Fourier transform of $y(x)$ over each dimension, $x^{(d)}$, of $\{x : \sigma_i(w(x)) > 0.5\}$ to obtain the empirical spectrum in that dimension. Note that we consider each dimension of $x$ individually since we have a multiplicative Q-component spectral mixture kernel over each dimension~\citep{wilson2014covariance}. Since spectral mixture kernels model the spectral density with $Q$ Gaussians on $\mathbb{R}^1$, we fit a 1-dimensional Gaussian mixture model,
\begin{eqnarray}
\label{eq:GMM}
p(x) = \sum_{q=1}^Q \phi_q \mathcal{N}(\mu_q, v_q)
\end{eqnarray}
to the empirical spectrum for each dimension. Using the learned mixture model we initialize the parameters of the spectral mixture kernels for $f_i(x)$.

\begin{algorithm}
\caption{Initialize spectral mixture kernels}
\label{alg:init_SM}
\begin{algorithmic}[1]
\FOR{$k_i:i=1:r$} 
\FOR{$d=1:D$}
\STATE Compute $x^{(d)} \in \{x : \sigma_i(w(x)) > 0.5\}$
\STATE Sample $s \sim | \text{FFT}(\text{sort}(y(x^{(d)}))) |^2$
\STATE Fit Q component GMM as  $p(s) = \sum_{q=1}^Q \phi_q^{(d)} \mathcal{N}(\mu_q^{(d)}, v_q^{(d)})$
\STATE Initialize $\omega_q = \text{std}(y(x^{(d)})) * \phi_q$
\ENDFOR
\ENDFOR
\end{algorithmic}
\end{algorithm}
After initializing $w(x)$ and spectral mixture hyperparameters, we jointly optimize the entire model using marginal likelihood and non-linear conjugate gradients \citep{rasmussen2010gaussian}.


\section{Experiments}
\label{sec:experiments}

We demonstrate the power and flexibility of GPCS by applying the model to a variety of numerical simulations and complex human settings. We begin with 2-dimensional numerical data in section \ref{sec:numerical_exp}, and show that GPCS is able to correctly model out-of-class polynomial change surfaces, and that it provides higher accuracy regressions than other comparable methods. Additionally we compute highly accurate counterfactual predictions for both GPCS and GPCS background models and discuss how the posterior distribution varies over the prediction domain as a function of the change surface.

We next consider coal mining, epidemiological, and urban policy data to provide additional analytical evidence for the effectiveness of GPCS and to demonstrate how GPCS results can be used to provide novel policy-relevant and scientifically-relevant insights. The ground truth against which GPCS is evaluated are the domain specific interventions in these case studies. 

In order to compare GPCS to standard changepoint models, we use a 1-dimensional dataset on the frequency of coal mining accidents. After fitting GPCS, we show that the change surface is able to identify a region of change similar to other changepoint methods. However, unlike changepoint methods that only identify a single moment of change, GPCS models how the data changes over time.

We then employ GPCS to analyze two complex spatio-temporal settings involving policy and scientific questions. First we examine requests for residential lead testing kits in New York City between 2014-2016, during a time of heightened concern about lead-tainted water. GPCS identifies a spatially and temporally varying change surface around the period when issues of water contamination were being raised in the news. We conduct a regression analysis on the resulting change surface features to better understand demographic factors that may have affected residents' concerns about lead-tainted water. 

Second, we apply GPCS to model state-level measles incidence in the United States during the twentieth century. GPCS identifies a substantial change around the introduction of the measles vaccine in 1963. However, the shape of the change surface varies over time for each state, indicating possible spatio-temporal heterogeneity in the adoption and effectiveness of the vaccination program during its initial years. We use regression analysis on the change surface features to explore possible institutional and demographic factors that may have influenced the impacts of the measles vaccination program. Finally, we estimate the counterfactual of measles incidence without vaccination by filtering out the detected change function and examining the inferred latent background function.

\subsection{Numerical Experiments}
\label{sec:numerical_exp}

We generate a $50\times 50$ grid of synthetic data by drawing independently from two latent functions, $f_1(x)$ and $f_2(x)$. Each function is characterized by an independent Gaussian process with a two-dimensional RBF kernel of different length-scales and signal variances. The synthetic change surface between the functions is defined by $\sigma(w_{\text{poly}}(x))$ where $w_{\text{poly}}(x)=\sum_{i=0}^3 \beta_i^T x^i$, $\beta_i \sim \mathcal{N}(0,3 I_D)$. We chose a polynomial change surface because it generates complex change patterns but is out-of-class when we use RKS features for $w(x)$, thus testing the robustness of GPCS to change surface misspecification.

\subsubsection{GPCS model}
\label{sec:num_exp_GPCS}

\begin{figure}[h]
  \centering
  \subfloat[]{\includegraphics[width=0.5\textwidth]{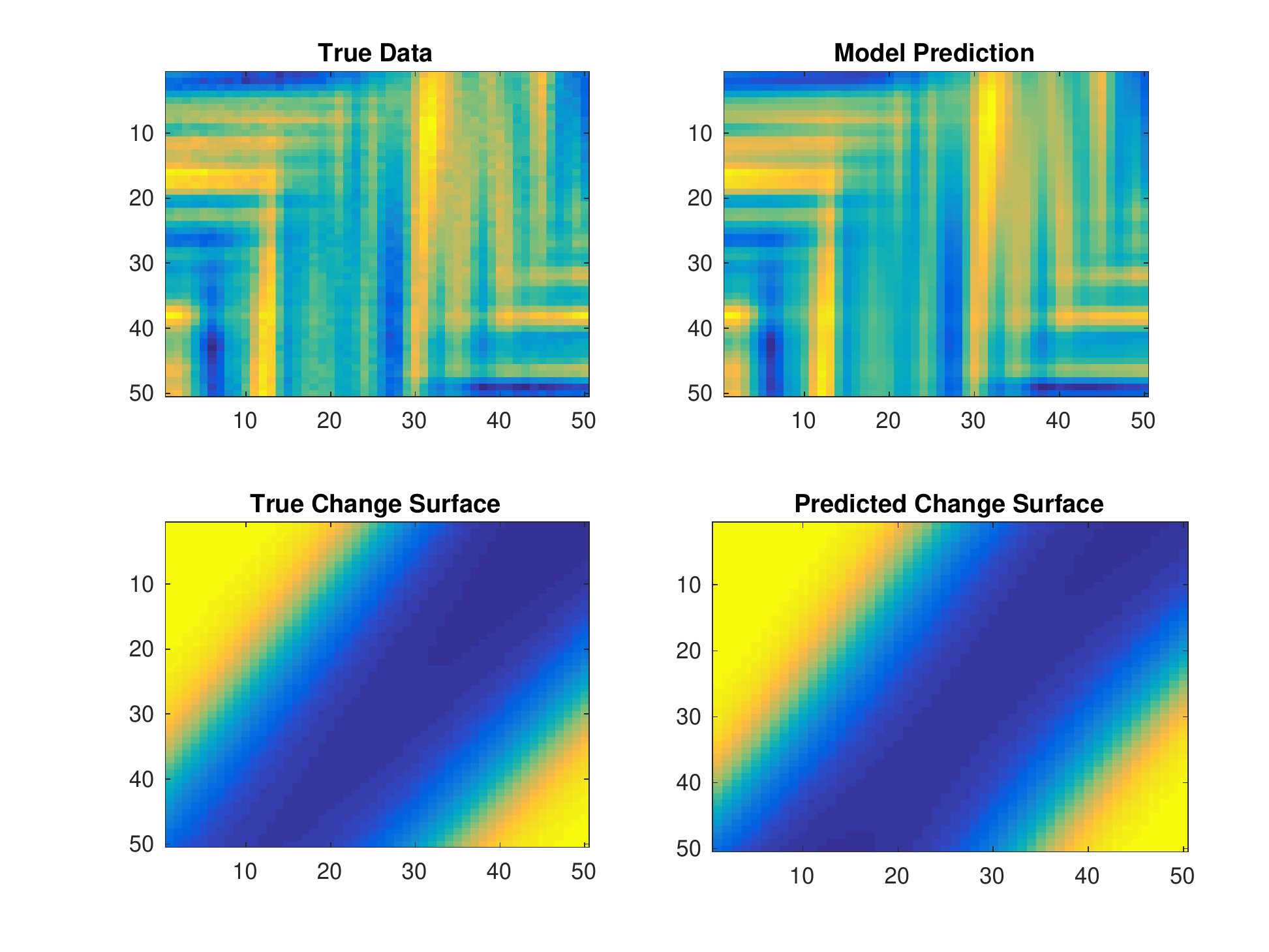}\label{fig:num_exp_GPCSf1}}
  \hfill
  \subfloat[]{\includegraphics[width=0.5\textwidth]{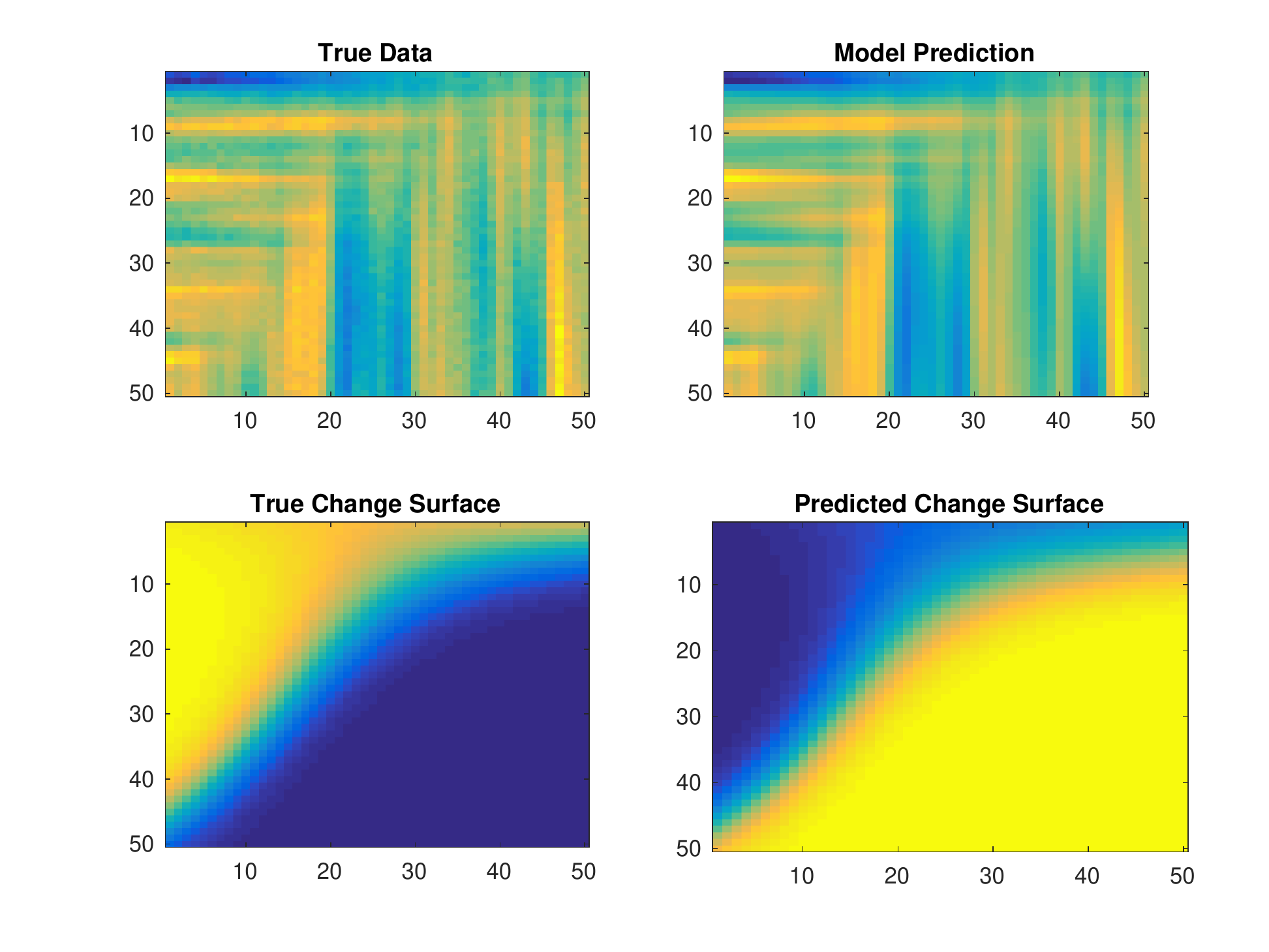}\label{fig:num_exp_GPCSf2}}
  \caption{Two numerical data experiments. In each of (a) and (b) the top-left plot depicts the data (e.g., observations indexed by two dimensional spatial inputs); the bottom-left shows the true change surface with the range from blue to yellow depicting $\sigma_1(w(x))$. The top-right depicts the predicted output; the bottom-right shows the predicted change surface. Note that the predicted change surface in plot (b) is flipped since the order of functions is not important.}
  \label{fig:num_exp_GPCS}
\end{figure}

Using the synthetic data generation technique described above we simulate data as $y = \sigma(w_{poly}(x)) f_1(x) + (1-\sigma(w_{\text{poly}}(x))) f_2(x) + \epsilon$, where $\epsilon \sim \mathcal{N}(0,\sigma_\epsilon^2)$. We apply GPCS  with two latent functions, spectral mixture kernels, and $w(x)$ defined by RKS features. We do not provide the model with prior information about the change surface or latent functions. As emphasized in section \ref{sec:initialization}, successful convergence is dependent on reasonable initialization. Therefore, we use $m_1=100$ and $m_2=20$ for Algorithm \ref{alg:init_wx}. Figure \ref{fig:num_exp_GPCS} depicts two typical results using the initialization procedure followed by analytic optimization. The model captures the change surface and produces an appropriate regression over the data. Note that in Figure \ref{fig:num_exp_GPCSf2} the predicted change surface is flipped since the order of functions is not important in GPCS.

\begin{figure}[h]
\centering
 \includegraphics[width=0.75\textwidth]{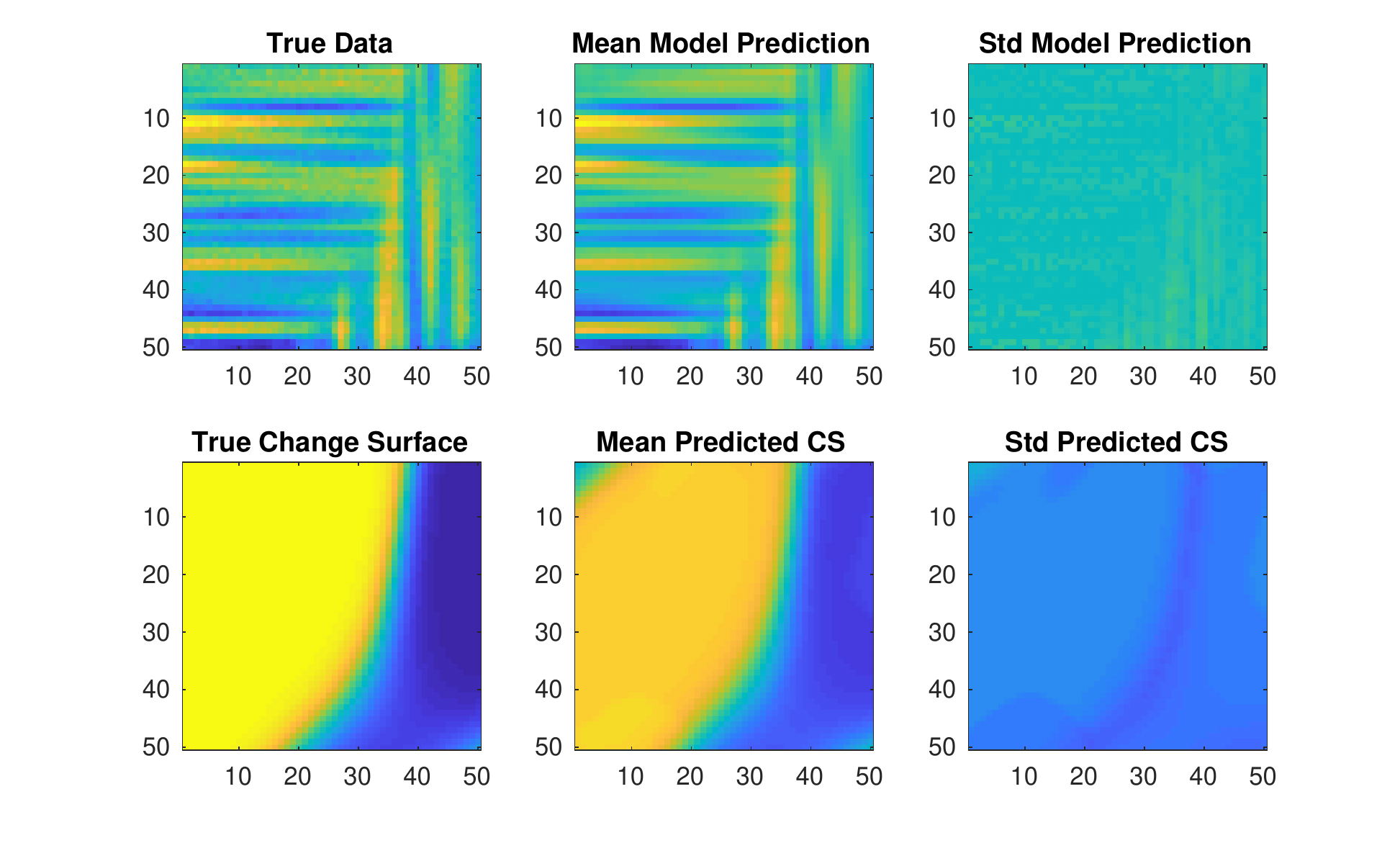}
\caption{Consistency results across 30 runs with different random seeds. True data and change surface are on the left, while the mean and standard deviation of the predicted results are in center and right panels.}
\label{fig:consistency}
\end{figure}

To demonstrate that the initialization method from section~\ref{sec:initialization} provides consistent results, we consider a numeric example and run GPCS 30 times with different random seeds. Figure~\ref{fig:consistency} provides the true data and change surface as well as the mean and standard deviation over the 30 experimental results using the section~\ref{sec:initialization} initialization procedure. For the predicted change surface we manually normalized the orientation of the change surface before computing summary statistics. The results illustrate that the initialization procedure provides accurate and consistent results for both $y$ and $\sigma(w(x))$ across multiple runs. Indeed, when we repeat these experiments with random initialization, instead of the procedure from section~\ref{sec:initialization}, the MSE between the predicted and true change surface is $58\%$ greater than when using our initialization procedure. Additionally, the results have a $17\%$ larger standard deviation of $\sigma(w(x_i))$ over the 30 runs, demonstrating that the produre we propose provides more consistent and accurate results.

Using synthetic data, we create a predictive test by splitting the data into training and testing sets. We compare GPCS to three other expressive, scalable methods: sparse spectrum Gaussian process with 500 basis functions \citep{lazaro2010sparse}, sparse spectrum Gaussian process with fixed spectral points with 500 basis functions \citep{lazaro2010sparse}, and a Gaussian process with multiplicative spectral mixture kernels in each dimension. For each method we average the results for 10 random restarts. For each method Table \ref{tab:comparison} shows the normalized mean squared error (NMSE),
\begin{eqnarray}
\label{eq:NMSE}
\text{NMSE} = \frac{\| y_{test} - y_{pred} \|_2^2}{\| y_{test} - \bar{y}_{train} \|_2^2}
\end{eqnarray}
where $\bar{y}_{train}$ is the mean of the training data. 
\begin{table}[]
\centering
\caption{Comparison of prediction accuracy (normalized mean squared error) using flexible and scalable Gaussian process methods on synthetic multidimensional change-surface data.}
\label{tab:comparison}
\begin{tabular}{|l|l|}
\hline
\textbf{Method}       & \textbf{NMSE} \\ \hline \hline
GPCS    & 0.00078       \\ \hline
SSGP       & 0.01530       \\ \hline
SSGP fixed  & 0.02820       \\ \hline
Spectral mixture      & 0.00200       \\ \hline
\end{tabular}
\end{table}

GPCS performed best due to the expressive non-stationary covariance function that fits to the different functional regimes in the data. Although the other methods can flexibly adapt to the data, they must account for the change in covariance structure by setting a shorter global length-scale over the data, thus underestimating the correlation of points in each regime. Thus their predictive accuracy is lower than GPCS, which can accommodate changes in covariance structure across the boundary of a change surface while retaining reasonable covariance length-scales within each regime.

\begin{figure}[h]
\centering
 \includegraphics[width=1.0\textwidth]{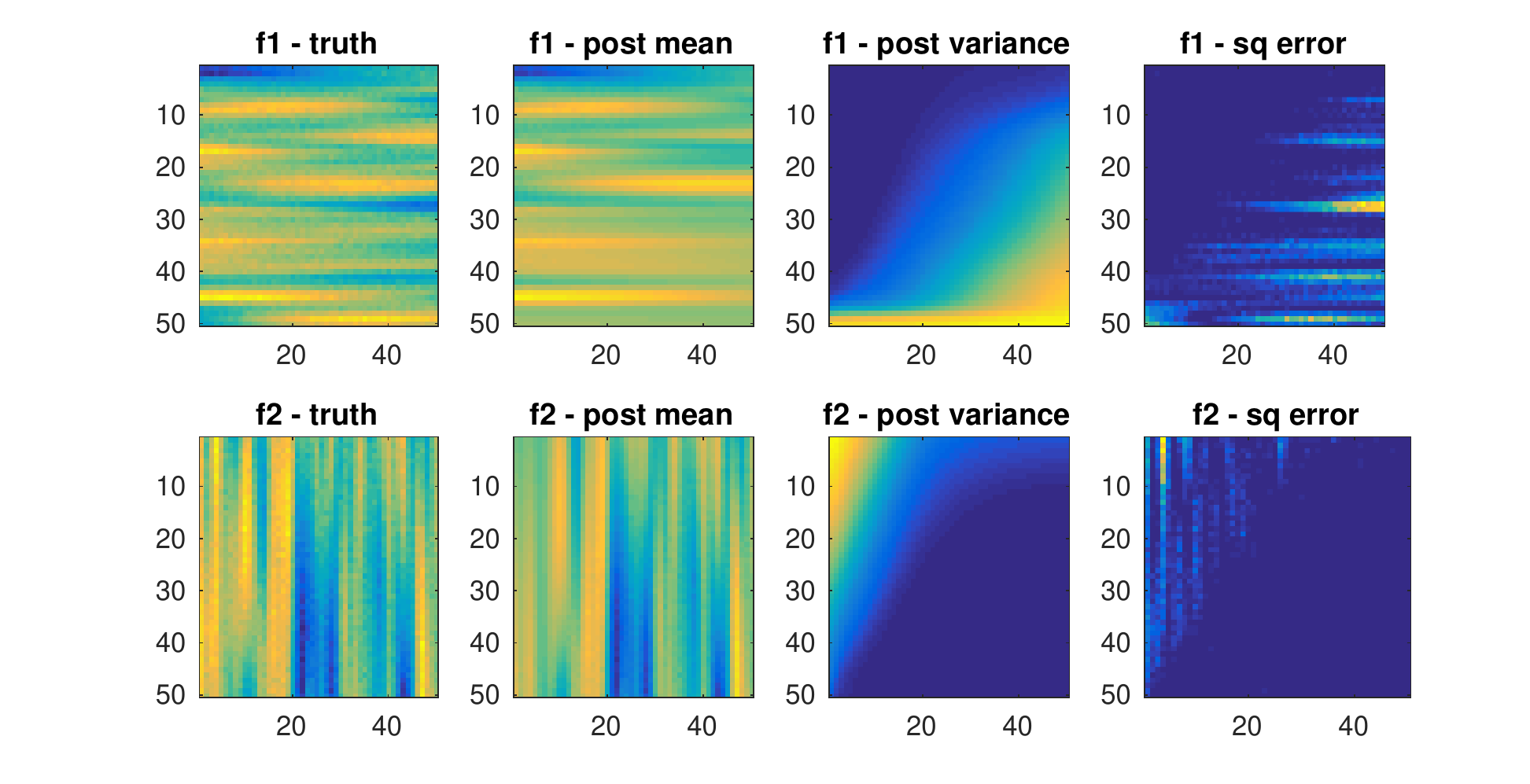}
\caption{Posterior counterfactual predictions using hyperparameters derived from GPCS model. We plot the true latent function as well as the posterior mean and variance estimates for each function. Additionally, we plot the squared error between the true and posterior mean values.}
\label{fig:GPCS_CF_truth_cf}
\end{figure}

We use GPCS to compute counterfactual predictions on the numerical data. In the previous experiments we used the data, $(x,y)$, to fit the parameters of GPCS, $\theta$. Now we condition on $(x,y,\theta)$ to infer the individual latent functions $f_1(x)$ and $f_2(x)$ over the entire domain, $x$. By employing the marginalization procedure described in section \ref{sec:GPCS_CF} we derive posterior distributions for both $f_1(x)$ and $f_2(x)$. Since we have synthetic data we can then compare the counterfactual predictions to the true latent function values. Specifically, we use $(x,y,\theta)$ from Figure \ref{fig:num_exp_GPCSf2} to infer the posterior counterfactual mean and variance for both $f_1(x)$ and $f_2(x)$ and show the results in Figure \ref{fig:GPCS_CF_truth_cf}.
Note how the posterior mean predictions of $f_1(x)$ and $f_2(x)$ are quite similar to the true values.  Moreover, the posterior uncertainty estimates are very reasonable. For both $f_1(x)$ and $f_2(x)$ the posterior variance varies over the two-dimensional domain, $x$, as a function of the change surface. Where $s_1(x)\approx1$ the posterior variance of $f_1(x)\approx0$ while the posterior variance of $f_2(x)$ is large. In areas where $s_2(x)\approx1$ the posterior variance of $f_1(x)$ is large, while the posterior variance of $f_2(x)\approx0$. The uncertainty is also evident in the squared error, $\frac{1}{n}\sum (f_i(x) - \hat{f_i(x)})$, where, as expected, each function has larger error in areas of high posterior variance.

As discussed in section~\ref{sec:GPCSmodel}, the 
underlying probabilistic Gaussian process model behind GPCS automatically \emph{discourages} extraneous complexity, favoring the simplest explanations consistent with the data \citep{mackay2003information, rasmussen2001occam, rasmussen2006gaussian, wilson2014fast, wilson2014covariance}.
This property enables GPCS to discover interpretable generative hypothesis for the data, which is crucial for public policy applications. This Bayesian Occam's razor principle is a cornerstone of many probabilistic approaches, such as automatically determining the intrinsic dimensionality for probabilistic PCA \citep{minka2001automatic}, or hypothesis testing for Bayesian neural network architectures \citep{mackay2003information}. In the absence of such automatic complexity control, these methods would always favour the highest intrinsic dimensionality or the largest neural network respectively.

\begin{figure}[h]
  \centering
  \subfloat[]{\includegraphics[width=0.5\textwidth]{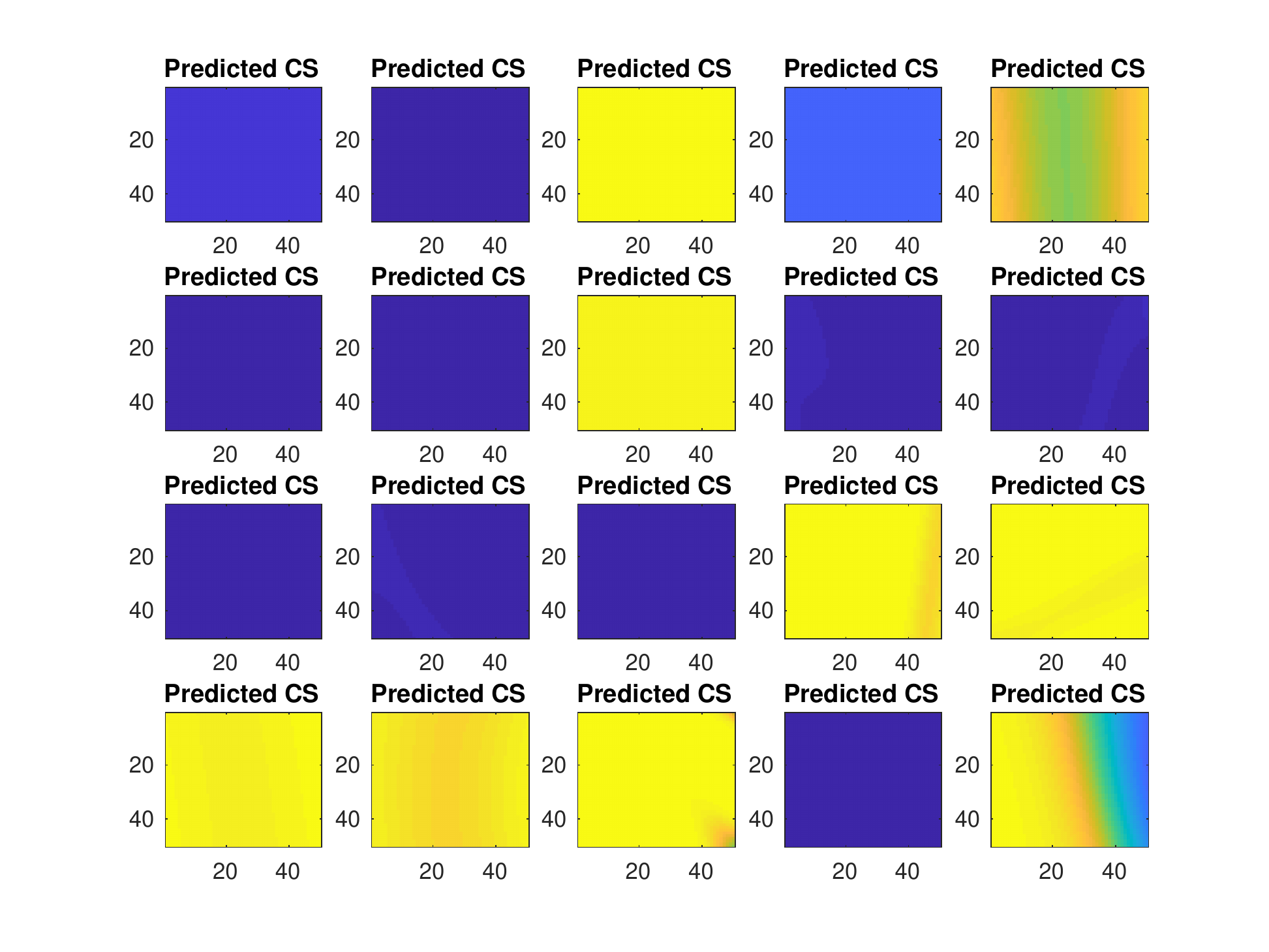}\label{fig:nointervention_colorplots}}
  \hfill
  \subfloat[]{\includegraphics[width=0.5\textwidth]{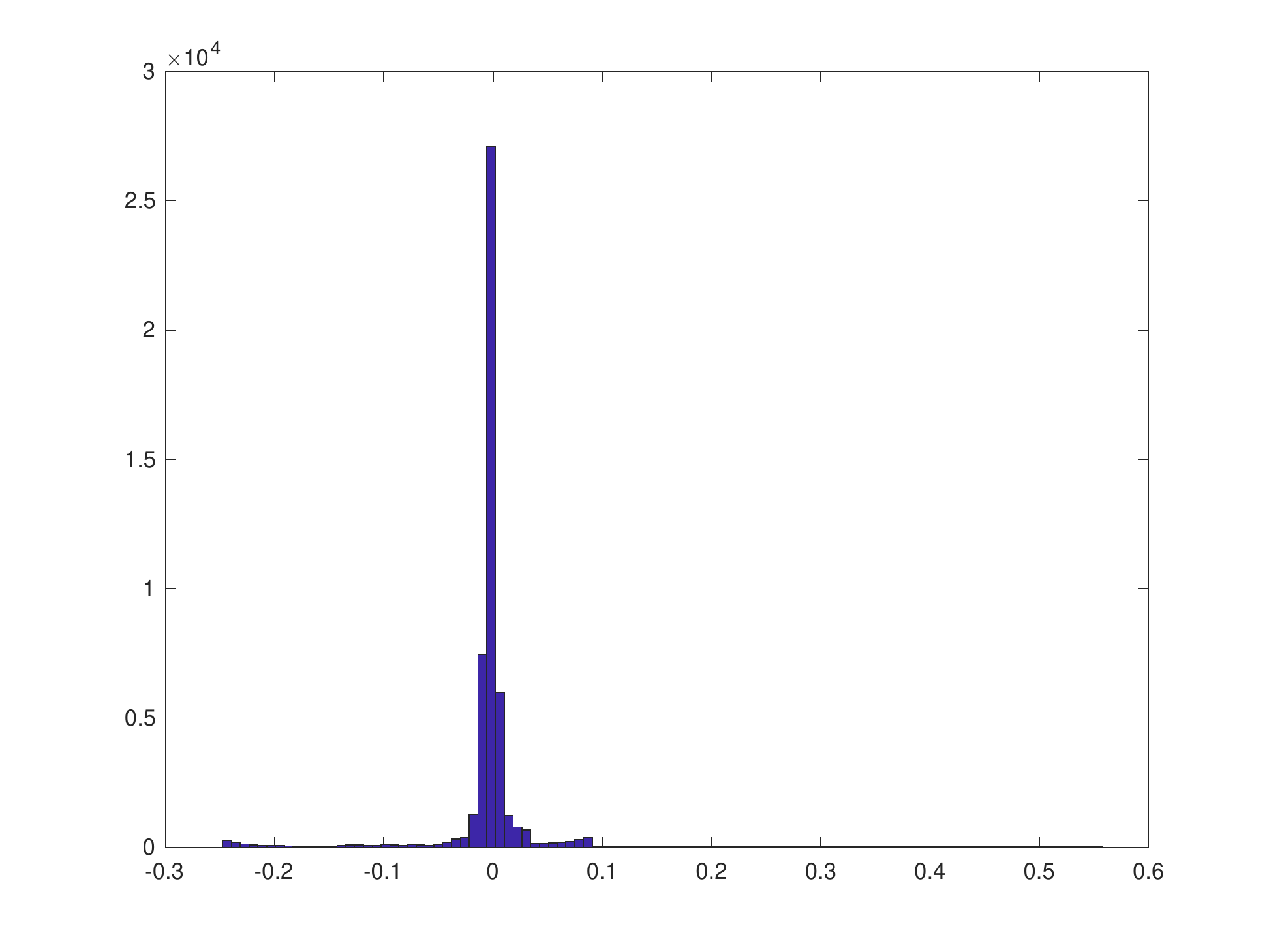}\label{fig:nointervention_hist100}}
  \caption{Data without any change surface, $\sigma(w_{\text{poly}}(x))=0$. The left panel depicts $\sigma_1(w(x))$ for each experiment. The right panel provides a histogram of the mean centered change surfaces values, $\sigma_1(w(x)) - \sum_{i\in n} \sigma_1(w(x_i))$.}
  \label{fig:nointervention}
\end{figure}

To demonstrate this Occam's razor principle in our context, we generate numeric data from a single GP without any change surface by setting $\sigma(w_{\text{poly}}(x))=0$, and fit a \emph{misspecified} GPCS model assuming two latent regimes. Figure~\ref{fig:nointervention} depicts the predicted change surfaces for 20 experiments of such data.
The left panel illustrates pictorially that the change surfaces are nearly all flat at either $\sigma_1(w(x))=0$ or $\sigma_1(w(x))=1$ for these experiments. Specifically, $\text{std}[\sigma_1(w(x))] < 0.03$ for all but two runs. This finding indicates that GPCS discovers that no dynamic transition exists and does not overfit the data, despite the added flexibility afforded by multiple mixture components. Only one of the 20 results (bottom-right) indicates a change, and even in that case the magnitude of the transition is markedly subdued as compared to the results in Figures~\ref{fig:num_exp_GPCS} and \ref{fig:num_exp_GPCS-Half}. While the upper-right result appears to have a large transition, in fact it has a flat change surface with $\text{std}[\sigma_1(w(x))] = 0.07$. The right panel provides a histogram of the mean centered change surface values for all experiments, $\sigma_1(w(x)) - \sum_{i\in n} \sigma_1(\omega(x_i))$, again demonstrating that GPCS learns very flat change surfaces and does not erroneously identify a change.

\subsubsection{GPCS background model}
\label{sec:num_exp_GPCS-Half}

We test the GPCS background model with a similar setup. Using the synthetic data generation technique described above, we simulate data as $y =f_0(x) + \sigma(w_{\text{poly}}(x)) f_1(x) + \epsilon$, where $\epsilon \sim \mathcal{N}(0,\sigma_\epsilon^2)$. We again note that the polynomial change surface is out-of-class.

\begin{figure}[h]
  \centering
  \subfloat[]{\includegraphics[width=0.5\textwidth]{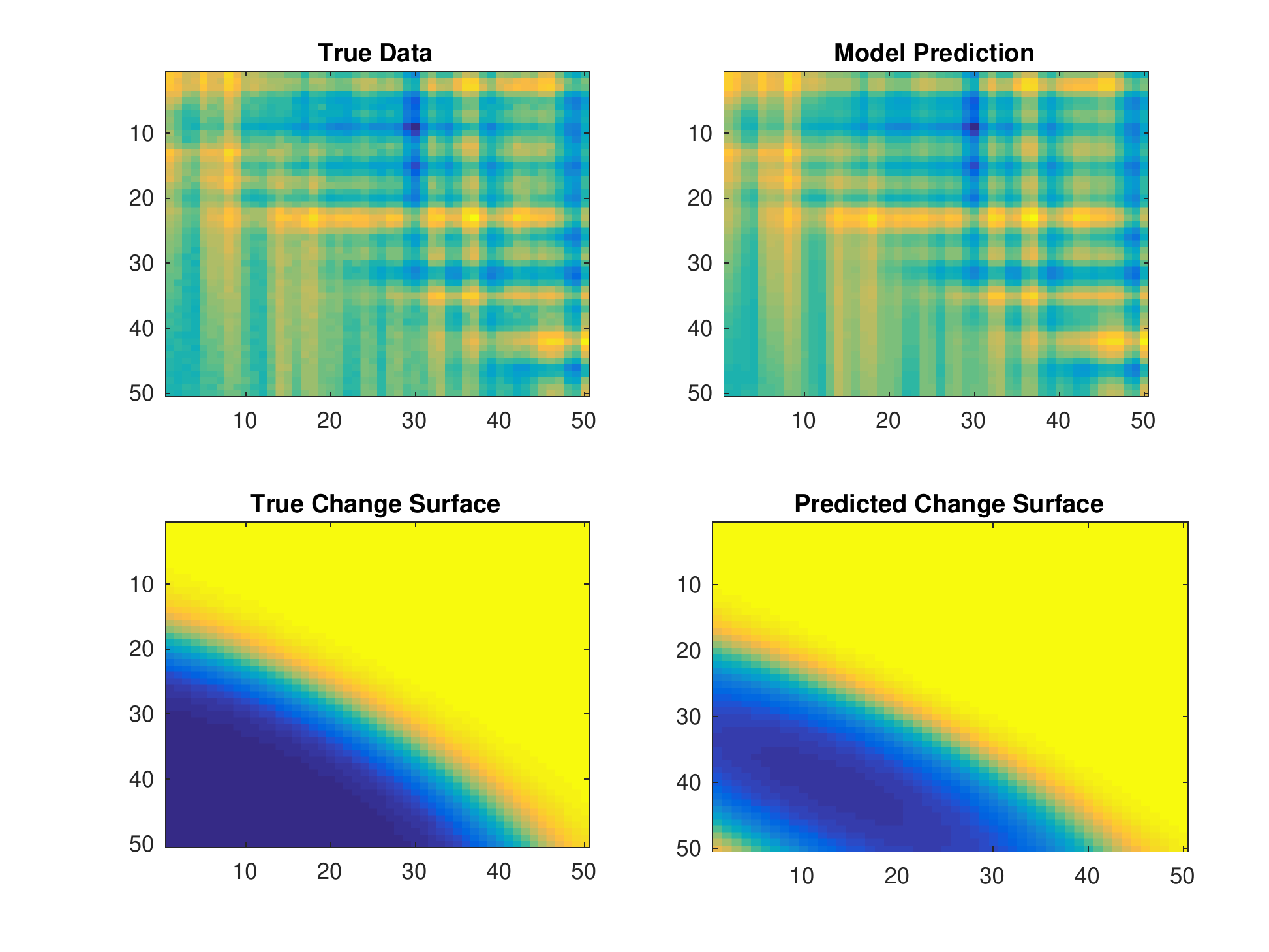}\label{fig:num_exp_GPCS-Half_f1}}
  \hfill
  \subfloat[]{\includegraphics[width=0.5\textwidth]{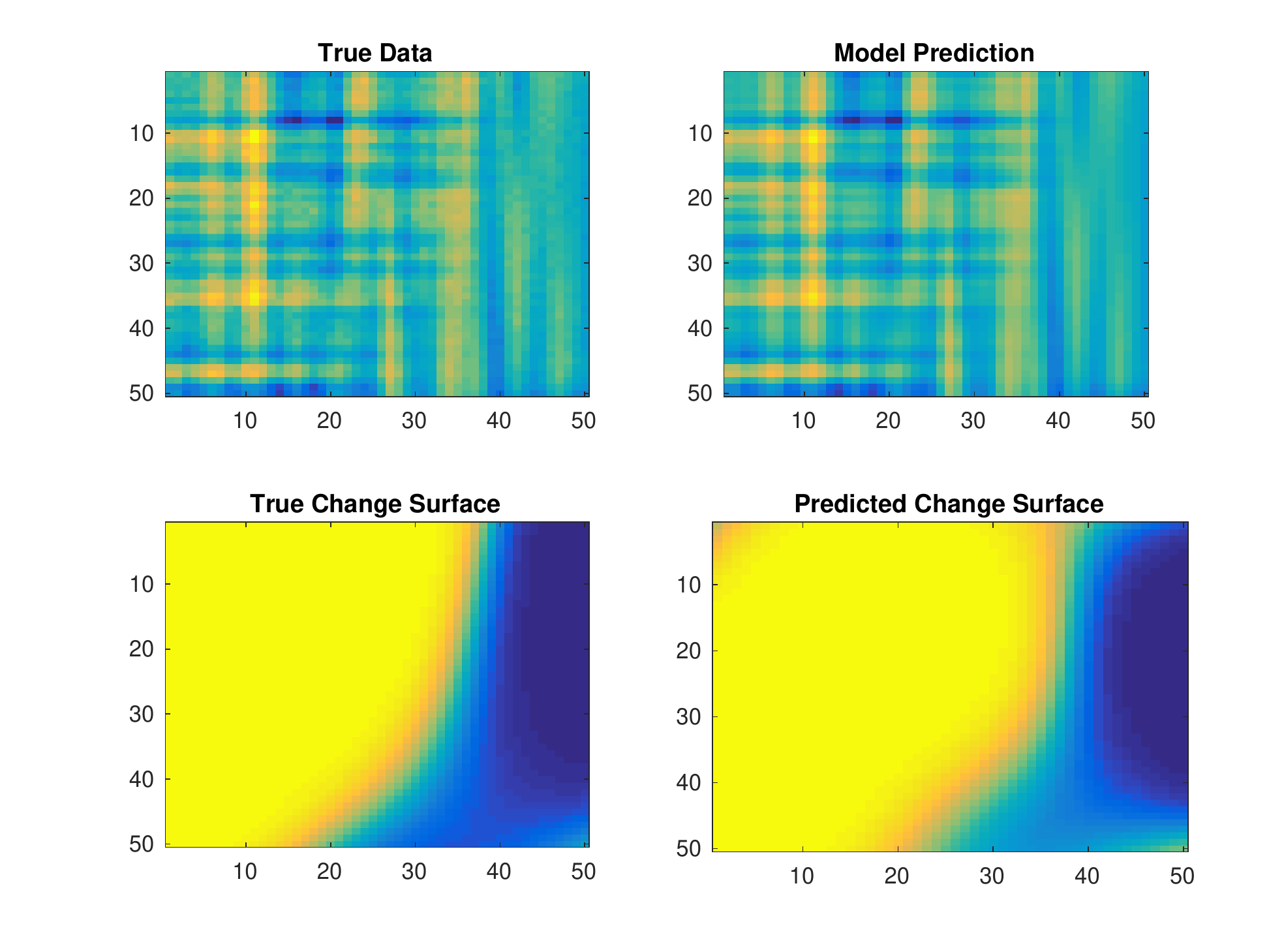}\label{fig:num_exp_GPCS-Half_f2}}
  \caption{Two numerical data experiments. In each of (a) and (b) the top-left plot depicts the data; the bottom-left shows the true change surface with the range from blue to yellow depicting $\sigma_1(w(x))$. The top-right depicts the predicted output; the bottom-right shows the predicted change surface.}
  \label{fig:num_exp_GPCS-Half}
\end{figure}

We apply the GPCS background model with one background function and one latent function scaled by a change surface. Both Gaussian process priors use spectral mixture kernels, and $w(x)$ is defined by RKS features. We do not provide the model with prior information about the change surface or latent functions. Figure \ref{fig:num_exp_GPCS-Half} depicts two typical results using the initialization procedure followed by analytic optimization. The model captures the change surface and produces an appropriate regression over the data.

\begin{figure}[h]
\centering
 \includegraphics[width=1.0\textwidth]{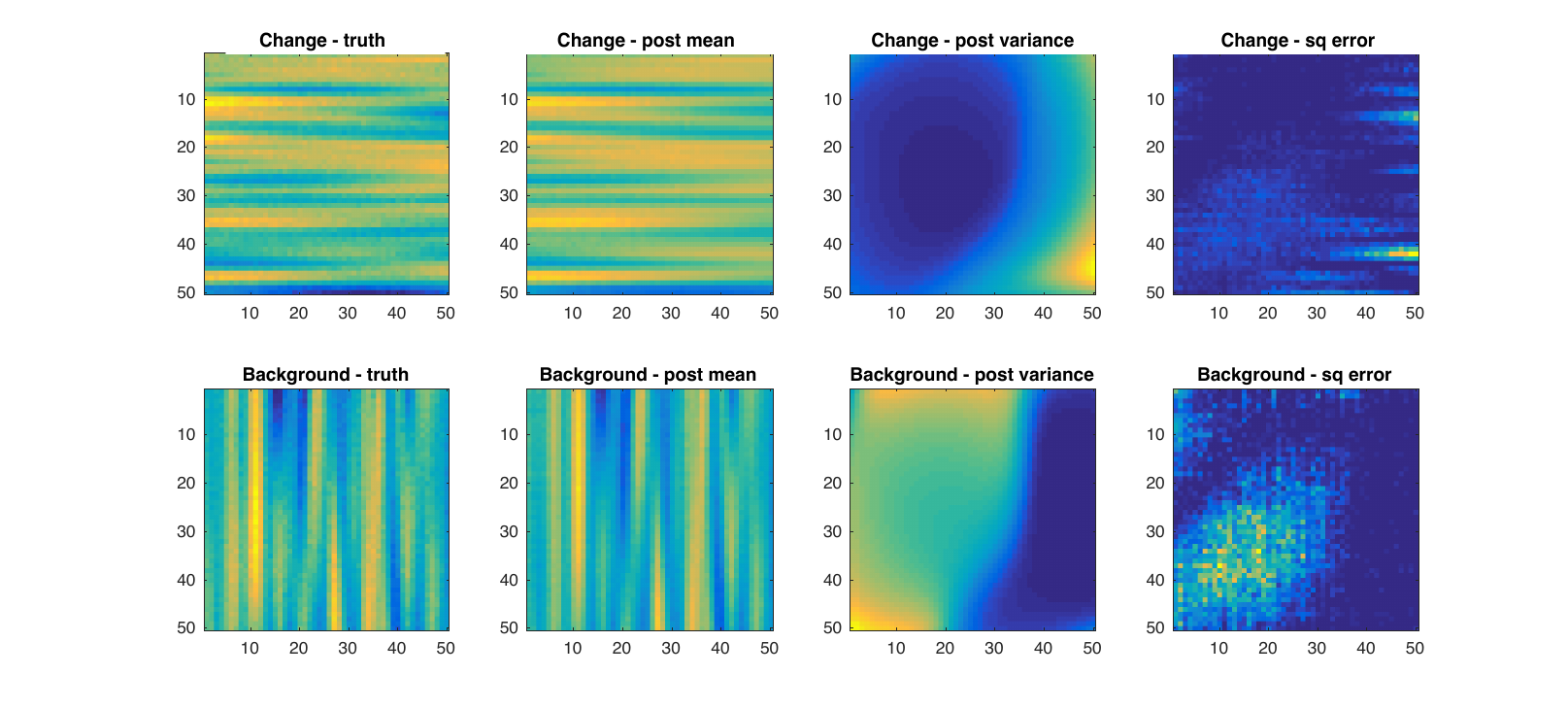}
\caption{Posterior counterfactual predictions using hyperparameters derived from GPCS background model. We plot the true latent function as well as the posterior mean and variance estimates for each function. Additionally, we plot the squared error between the true and posterior mean values.}
\label{fig:GPCS-Half_CF_truth_cf}
\end{figure}

We use the GPCS background model to compute counterfactual predictions on the data from Figure \ref{fig:num_exp_GPCS-Half_f2}. Conditioning on $(x,y,\theta)$ we employ the marginalization procedure described in section \ref{sec:GPCS_CF} to infer posterior distributions for the background function, $f_0(x)$, and the change function, $f_1(x)$, over the entire domain, $x$. The results are shown in Figure \ref{fig:GPCS-Half_CF_truth_cf}.
Note how the posterior mean predictions of both the background and change functions are quite similar to the true values. As in the case of GPCS, the posterior variance for each function varies over the two-dimensional domain, $x$, as a function of the change surface, $\sigma(w_{\text{poly}}(x))$.

\subsubsection{Log Gaussian Cox Process}
\label{sec:LGCP}
The numerical experiments above demonstrate the consistency of GPCS for identifying out-of-sample change surfaces and modeling complex data for high accuracy prediction. To further demonstrate the flexibility of the model, we apply GPCS to data generated by a log-Gaussian Cox process~\citep{moller1998log, flaxman2015fast}. This inhomogeneous Poisson process is modulated by a stochastic intensity defined as a GP,
\begin{align}
\lambda &= f\\
f &\sim \mathcal{GP}(\mu, K)
\end{align}
Conditional on $\lambda$, and letting $s$ denote a region in space-time, the resulting small-area count data are non-negative integers distributed as
\begin{align}
y(s)\:|\:\lambda \sim \text{Poisson}\big( \exp \int_s \lambda(x) dx \big).
\end{align}
We let this GP model be a convex combination of two GPs with an out-of-sample change surface, as described in section~\ref{sec:numerical_exp}. Thus we generated data from this model as
\begin{align}
y\:|\:f_1(x_i),f_2(x_i) \sim \text{Poisson}\Big( \exp \big[ \sigma(w_{poly}(x)) f_1(x) + (1-\sigma(w_{\text{poly}}(x))) f_2(x) + \epsilon  \big] \Big).
\end{align}
Such data substantially departs from the type of data that GPCS is designed to model. Indeed, while custom approaches are often created to handle inhomogeneous Poisson data~\citep{flaxman2015fast, shirota2016inference}, we use GPCS to demonstrate its flexibility and applicability to complex non-Gaussian data.
\begin{figure}[h]
  \centering
  \subfloat[]{\includegraphics[width=0.5\textwidth]{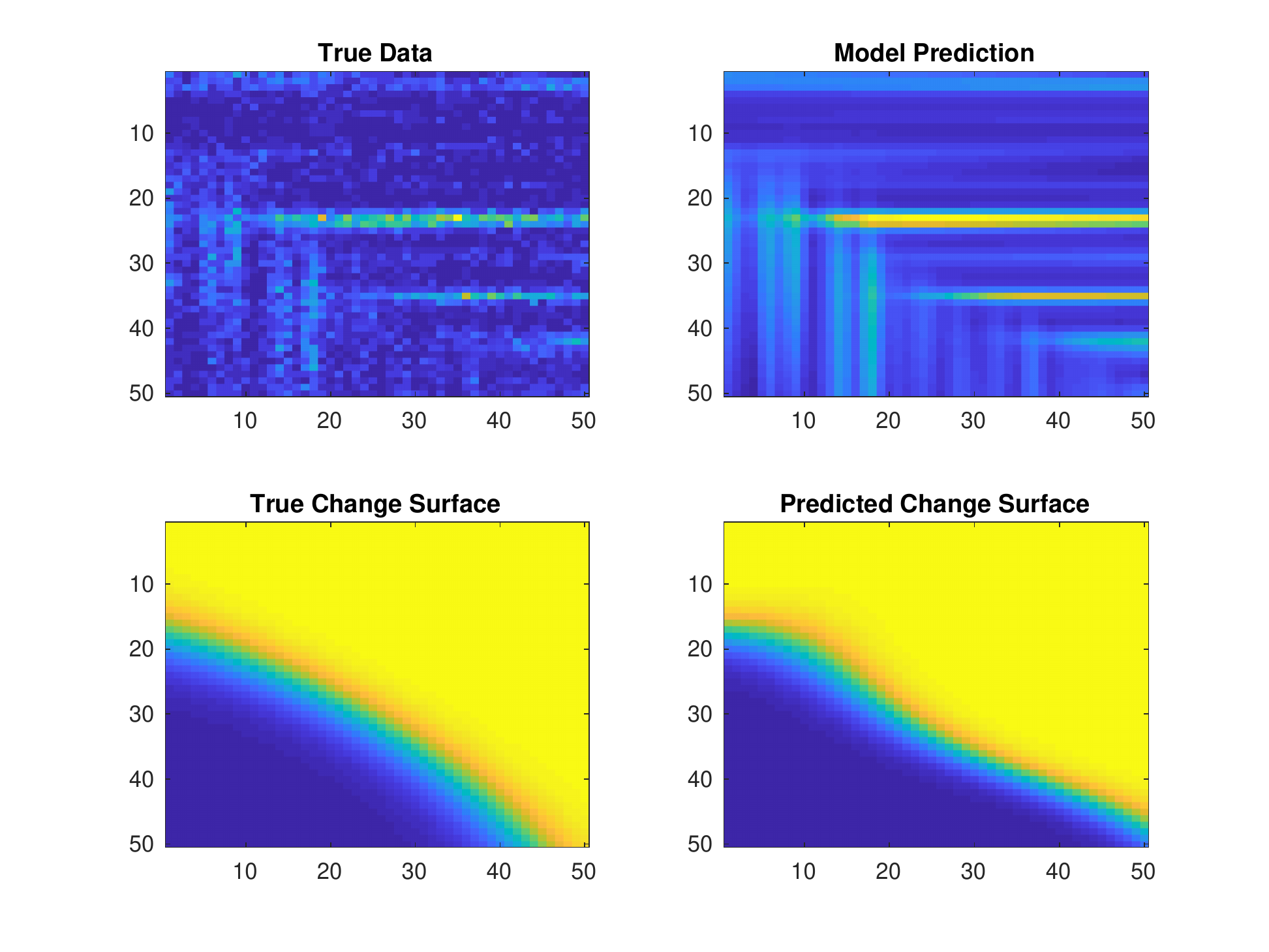}\label{fig:LGCP_a}}
  \hfill
  \subfloat[]{\includegraphics[width=0.5\textwidth]{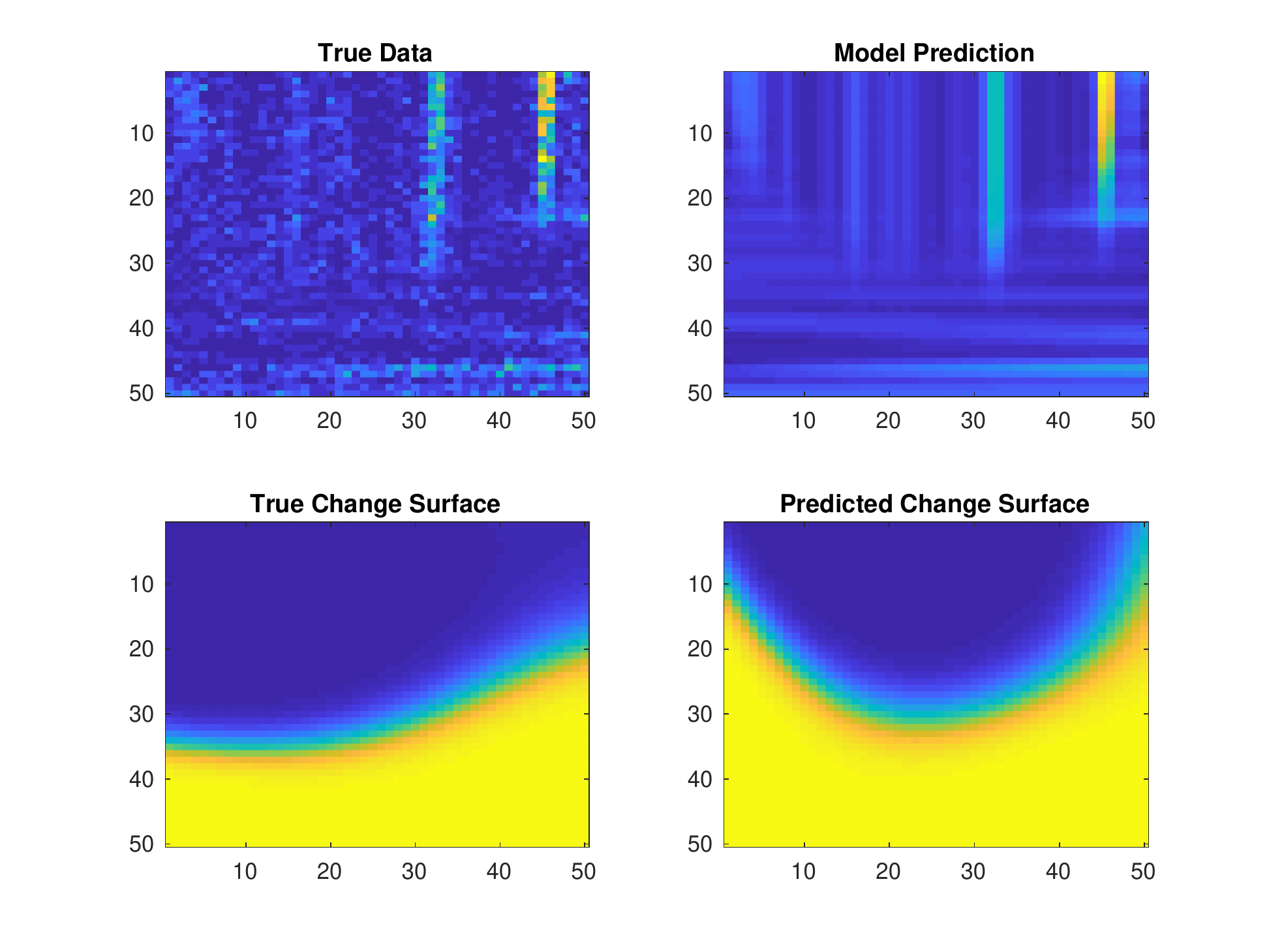}\label{fig:LGCP_b}}
  \caption{Two numerical data experiments with data from a log-Gaussian Cox process. In each of (a) and (b) the top-left plot depicts the data (e.g., observations indexed by two dimensional spatial inputs); the bottom-left shows the true change surface with the range from blue to yellow depicting $\sigma_1(w(x))$. The top-right depicts the predicted output; the bottom-right shows the predicted change surface.}
  \label{fig:LGCP}
\end{figure}
The results are shown in Figure~\ref{fig:LGCP}. The model provides accurate change surfaces and predictions even though the data is substantially out-of-class -- even beyond the out-of-class change surface data from sections \ref{sec:num_exp_GPCS} and \ref{sec:num_exp_GPCS-Half}. The precise location of change surfaces deviates slightly in GPCS, particularly on the left edge of Figure~\ref{fig:LGCP_b} where the raw data is highly stochastic. Additionally, the model predictions are smoothed versions of the true latent data, which reflects the fundamental difference between Gaussian and Poisson models.

\subsection{British Coal Mining Data}
\label{sec:coal_exp}

British coal mining accidents from 1861 to 1962 have been well studied as a benchmark in the point process and changepoint literature \citep{raftery1986bayesian, carlin1992hierarchical, adams2007bayesian}. We use yearly counts of accidents from \citet{jarrett1979note}. \citet{adams2007bayesian} indicate that the Coal Mines Regulation Act of 1887 affected the underlying process of coal mine accidents. This act limited child labor in mines, detailed inspection procedures, and regulated construction standards \citep{law1887coal}. We apply GPCS to show that it can detect changes corresponding to policy interventions in data while providing additional information beyond previous changepoint approaches.

\begin{figure}[h]
\centering
 \includegraphics[width=0.75\textwidth]{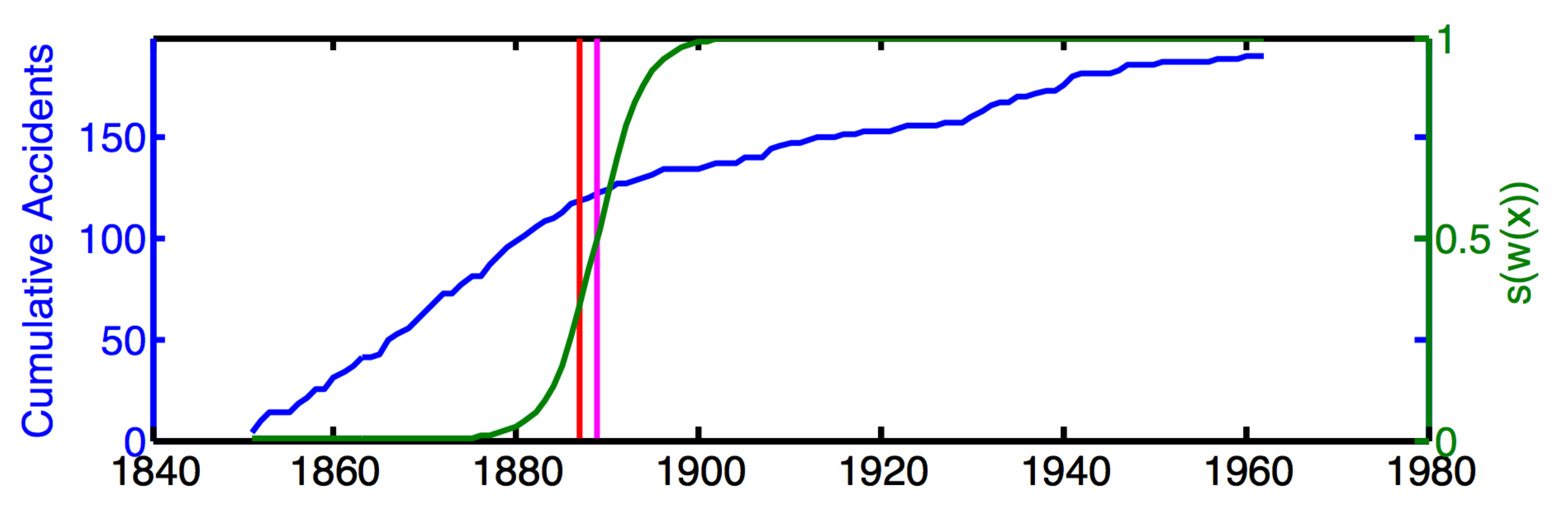}
\caption{British coal mining accidents from 1851 to 1962. The blue line depicts cumulative annual accidents, the green line plots $\sigma(w(x))$, the vertical red line marks the Coal Mines Regulation Act of 1887, and the vertical magenta line indicates $\sigma(w(x))=0.5$.}
\label{fig:coal_results}
\end{figure}

We consider GPCS with two latent functions, spectral mixture kernels, and $w(x)$ defined by RKS features. We do not provide the model with prior information about the 1887 legislation date. Figure \ref{fig:coal_results} depicts the cumulative data and predicted change surface. The red line marks the year 1887 and the magenta line marks $x:\sigma(w(x))=0.5$. GPCS correctly identified the change region and suggests a gradual change that took 5.6 years to transition from $\sigma(w(x))=0.25$ to $\sigma(w(x))=0.75$.

Using the coal mining data we apply a number of well known univariate changepoint methods using their standard settings. We compared Pruned Exact Linear Time (PELT) \citep{killick2012optimal} for changes in mean and variance and a nonparametric method named ``ecp'' \citep{james2013ecp}. Additionally, we tested the batch changepoint method described in \cite{ross2013parametric} with Student-$t$ and Bartlett tests for Gaussian data as well as Mann-Whitney and Kolmogorov-Smirnov tests for nonparametric changepoint estimation \citep{sharkey2014nonparametric}. Figure \ref{tab:compare_changepoints} compares the dates of change identified by these methods to the midpoint date where $\sigma(w(x))=0.5$ in GPCS.

\begin{table}[]
\centering
\caption{Comparing methods for estimating the date of change in coal mining data.}
\label{tab:compare_changepoints}
\begin{tabular}{|l|l|}
\hline
\textbf{Method} &  \textbf{Estimated date} \\ \hline
GPCS $\sigma(w(x))=0.5$ &   1888.8  \\ \hline
PELT mean change &	1886.5\\ \hline
PELT variance change &	1882.5\\ \hline
ecp	&1887.0  \\ \hline
Student-t test	&1886.5  \\ \hline
Bartlett test	& 1947.5 \\ \hline
Mann-Whitney test	& 1891.5 \\ \hline
Kolmogorov-Smirnov	 test  &    1896.5 \\ \hline
\end{tabular}
\end{table}

Most of the methods identified a midpoint date between 1886 and 1895. While each method provides a point estimate of the change, only GPCS provides a clear, quantitative description of the development of this change. Indeed the 5.6 years during which the change surface transitions between $\sigma(w(x))=0.25$ to $\sigma(w(x))=0.75$ nicely encapsulate most of the point estimate method results.

\subsection{New York City Lead Data}
\label{sec:lead_exp}

In recent years there has been heightened concern about lead-tainted water in major United States metropolitan areas. For example, concerns about lead poisoning in Flint, Michigan's water supply garnered national attention in 2015 and 2016, leading to Congressional hearings. Similar lead contamination issues have been reported in a spate of United States cities such as Cleveland, OH, New York, NY, and Newark, NJ \citep{newark}. Lead concerns in New York City have focused on lead-tainted water in schools and public housing projects, prompting reporting in some local and national media \citep{nycleadhousing}.

\begin{figure}[h]
\centering
  \includegraphics[width=0.65\textwidth]{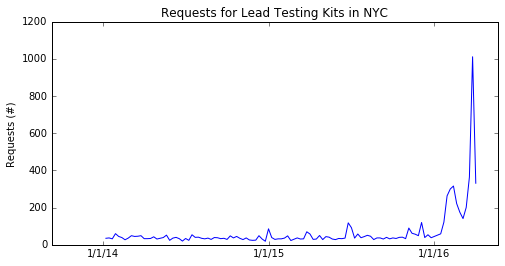}
\caption{Requests for residential lead testing kits in New York City aggregated at a weekly level across the entire city.}
\label{fig:lead_time_agg}
\end{figure}

In order to understand the evolving dynamics of New York City residents' concerns about lead-tainted water, we analyzed requests for residential lead testing kits in New York City. These kits can be freely ordered by any resident of New York City and allow individuals to test their household's water for elevated levels of lead \citep{leadtestingkit}. We considered weekly requests for each zip code in New York City from January 2014 through April 2016. This provides a proxy for measuring the concern about lead tainted water. Figure \ref{fig:lead_time_agg} shows the aggregated requests over the entire city for lead testing kits during the observation period. It could be argued that this is an imperfect reflection of citizen concern since is unlikely that a household will request more than one testing kit within a relatively short period of time. Thus a reduction in requests may be due to saturation in demand for kits rather than a decrease in concern. However, we contend that since there were only 28,057 requests for lead testing kits over the entire observation period, and New York City contains approximately 3,148,067 households, there is a substantial pool of households in New York City that are able to signal their concern through requesting a lead testing kit \citep{acs2014year1}.

While there is a distinct uptick in requests for kits towards the middle and end of the observation period, there is no ground truth change point, unlike the coal mining example in section \ref{sec:coal_exp} and the measles incidence example in section \ref{sec:disease_exp}. We apply GPCS with two latent functions, spectral mixture kernels, and $w(x)$ defined by RKS features. Note that the inputs are three dimensional, $x\in \mathbb{R}^{3}$, with two spatial dimensions representing centroids of each zipcode and one temporal dimension.

\begin{figure}[h]
\centering
 \includegraphics[width=0.65\textwidth]{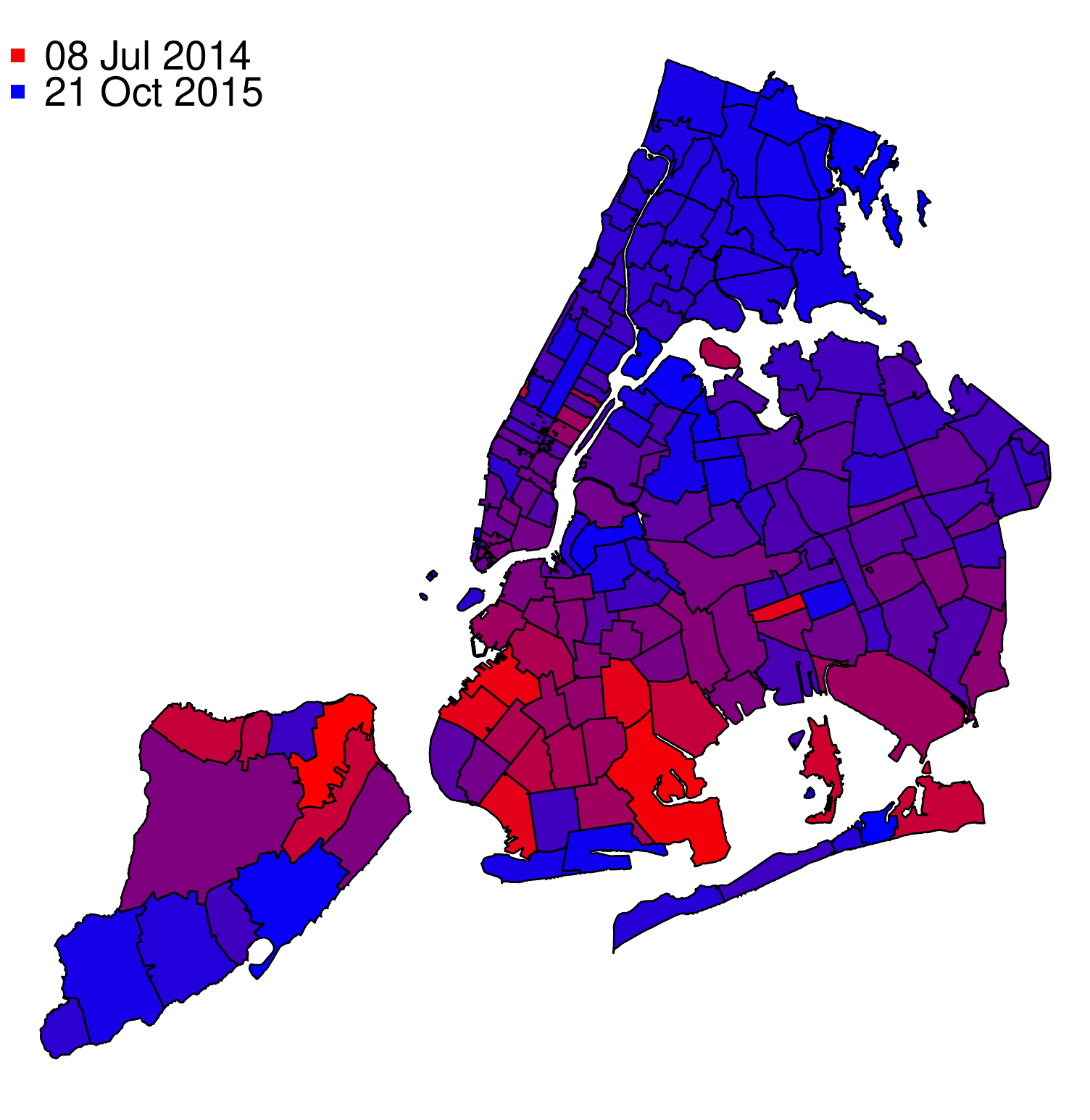}
\caption{NYC zip codes colored by the date where $\sigma(w(x_{\text{zip}}))=0.5$. Red indicates earlier dates, with Bulls Head in Staten Island being the earliest. Blue indicates later dates, with New Hyde Park at the eastern edge of Queens being the latest.}
\label{fig:NYC_dates}
\end{figure}
The model suggests that residents' concerns about lead tainted water had distinct spatial and temporal variation. In Figure \ref{fig:NYC_dates} we depict the midpoint, $\sigma(w(x_{\text{zip}}))=0.5$, for each zip code. We illustrate the spatial variation in the midpoint date by shading zip codes with an early midpoint in red and zip codes with later midpoint in blue. Regions in Staten Island and Brooklyn experienced the earliest midpoints, with Bulls Head in Staten Island (zip code 10314) being the first area to reach $\sigma(w(x_{\text{zip}}))=0.5$ and New Hyde Park at the eastern edge of Queens (zip code 11040) being the last. The model detects certain zip codes changing in mid to late 2014, which somewhat predates the national publicity surrounding the Flint water crisis. However, most zip codes have midpoint dates sometime in 2015.

\begin{figure}[h]
\centering
  \includegraphics[width=0.65\textwidth]{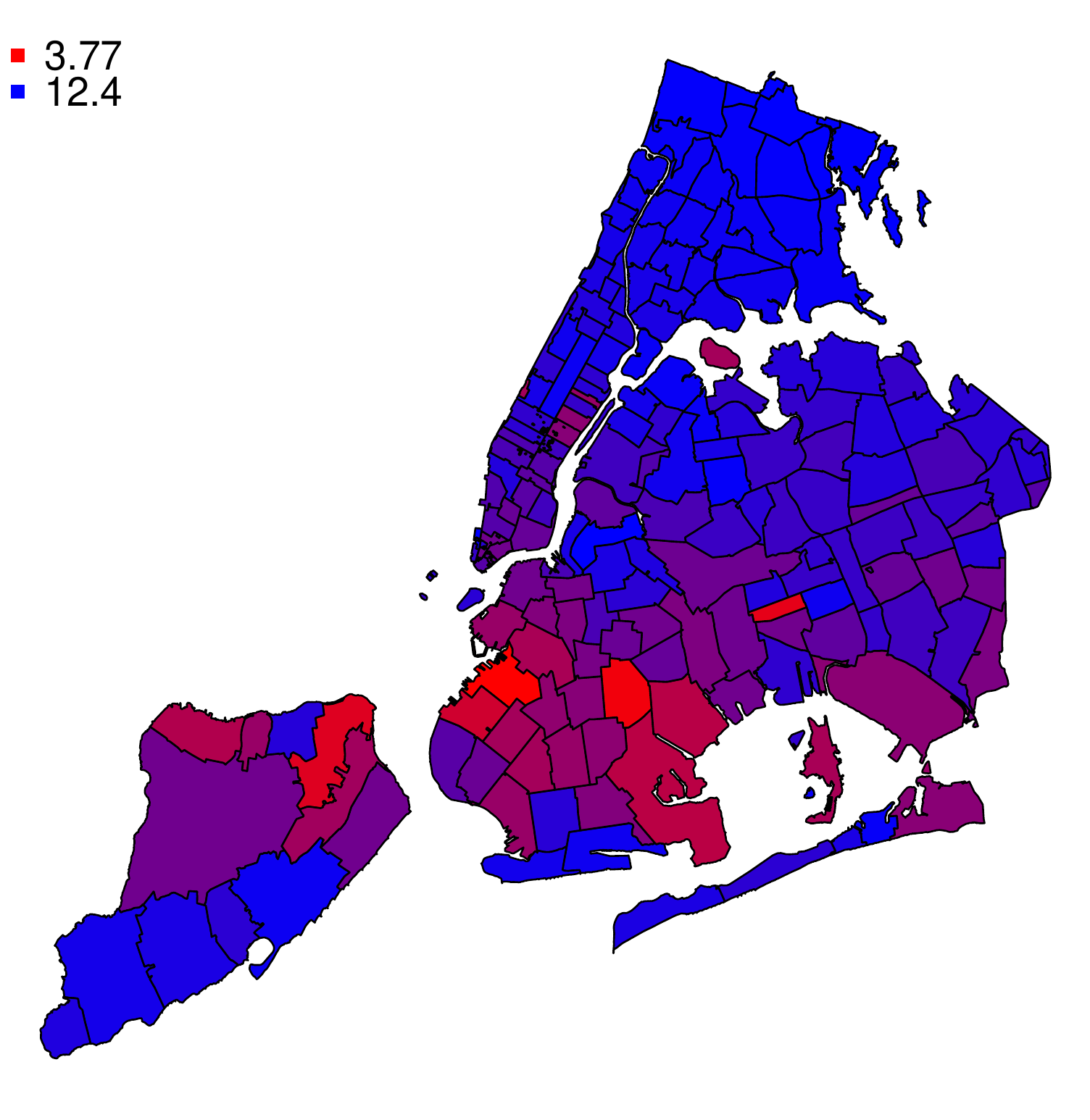}
\caption{NYC zip codes colored by the slope of $\sigma(w(x_{\text{zip}}))$ from $0.25$ to $0.75$. Red indicates flatter slopes, with Mariner's Harbor in Staten Island being the flattest. Blue indicates steeper slopes, with Woodlawn Heights in the Bronx being the steepest.}
\label{fig:NYC_slope}
\end{figure}
In Figure \ref{fig:NYC_slope} we depict the change surface slope from $\sigma(w(x_{\text{zip}}))=0.25$ to $\sigma(w(x_{\text{zip}}))=0.75$ for each zip code to estimate the rate of change. We illustrate the variation in slope by shading zip codes with flatter change slopes in red and the steeper change slopes in blue. The flattest change surface occurred in Mariner's Harbor in Staten Island (zip code 10303) while the steepest change surface occurred in Woodlawn Heights in the Bronx (zip code 10470).  We find that some zip codes had approximately four times the rate of change as others.

\paragraph{Regression analysis:} The variations in the change surface indicate that the concerns about lead-tainted water may have varied heterogeneously over space and time. In order to better understand these patterns we considered demographic and housing characteristics that may have contributed to differential concern among residents in New York City. Specifically we examined potential factors influencing the midpoint date between the two regimes. All data were taken from the 2014 American Community Survey 5 year average at the zip code level \citep{acs2014year5}. Factors considered included information about residents such as education of householder, whether the householder was the home owner, previous year's annual income of household, number of people per household, and whether a minor or senior lived in the household. Additionally, we considered information about when the homes were built.

\begin{table}[!htbp] \centering 
  \caption{Results from a linear regression to the NYC lead midpoint date, $\sigma(w(x_{\text{zip}}))=0.5$. Variables are listed on the left while their coefficients, with standard errors in parentheses, are listed on the right. Asterisks indicate statistically significant variables.}
    \label{tab:lead_demographic}
\begin{tabular}{@{\extracolsep{5pt}}lD{.}{.}{-3} } 
\\[-1.8ex]\hline 
\hline \\[-1.8ex] 
 & \multicolumn{1}{c}{\textit{Dependent variable:}} \\ 
\cline{2-2} 
\\[-1.8ex] & \multicolumn{1}{c}{Midpoint date} \\ 
\hline \\[-1.8ex] 
Log median household income & 21.916^{**} \\ 
  & (7.912) \\ 
\% homes built after 2010 & 0.549 \\ 
  & (0.724) \\ 
\% homes built 2000-2009 & 0.061 \\ 
  & (0.164) \\ 
\% homes built 1980-1999 & -0.070 \\ 
  & (0.153) \\ 
\% homes built 1960-1979 & 0.027 \\ 
  & (0.094) \\ 
\% homes built 1940-1959 & 0.667^{**} \\ 
  & (0.092) \\ 
\% education high school equivalent & -1.609^{**} \\ 
  & (0.331) \\ 
\% education some college & 0.143 \\ 
  & (0.312) \\ 
\% education college and above & -0.864^{**} \\ 
  & (0.303) \\ 
\% households owner occupied & -0.310^{*} \\ 
  & (0.126) \\ 
 Average family size & 9.507 \\ 
  & (6.453) \\ 
\% households with member 18 or younger & -0.020 \\ 
  & (0.282) \\ 
\% households with member 60 or older & 0.202 \\ 
  & (0.215) \\ 
\% households with only one member & 0.283 \\ 
  & (0.227) \\ 
 Constant & -149.602 \\ 
  & (77.036) \\ 
\hline \\[-1.8ex] 
Observations & \multicolumn{1}{c}{176} \\ 
R$^{2}$ & \multicolumn{1}{c}{0.420} \\ 
Adjusted R$^{2}$ & \multicolumn{1}{c}{0.370} \\
\hline 
\hline \\[-1.8ex] 
\textit{Note:}  & \multicolumn{1}{r}{$^{*}$p$<$0.05; $^{**}$p$<$0.01} \\ 
\end{tabular} 
\end{table} 

Results of a linear regression over all factors can be seen in Table \ref{tab:lead_demographic}.  Five variables were statistically significant at a p-value $<0.05$: median annual household income, percentage of houses built 1940-1959, percentage of householders with high school equivalent education, percentage of householders with at least a college education, and percentage of owner occupied households. Median annual household income had a positive correlation with the change date, suggesting that higher household income is associated with later midpoint dates. People with lower incomes may tend to live in housing that is less well maintained, or is perceived to be less well maintained. Thus they may require less ``activation energy'' to request lead testing kits when faced with possible environmental hazards. Education levels were compared to a base value of householders with less than a high school education. Thus zip codes with more educated householders tended to have earlier midpoint dates, and more concern about lead-tainted water. Similarly, owner occupied households had a negative correlation with the midpoint date. Since owner occupiers may tend to have more knowledge about their home infrastructure and may expect to remain in a location for longer than renters -- perhaps even over generations -- they could have a greater interest in ensuring low levels of water-based lead. The positive correlation of homes built between 1940-1959 may be due to a geographic anomaly since zip codes with the highest proportion of these homes are all in Eastern Queens. This region has very high median incomes which may ultimately explain the later midpoint dates.

This analysis indicates that more education and outreach to lower-income families by the New York City Department of Environmental Protection could be an effective means of addressing residents' concerns about future health risks. Additionally, it suggests an information disparity between renters and owner-occupiers that may be of interest to policy makers. Beyond the statistical analysis of demographic data, we also qualitatively examined media coverage related to the Flint water crisis as detailed by the Flint Water Study \citep{flintwaterstudy}. While some articles and news reports were reported in 2014, the vast majority began in 2015. The increased rate and national scope of this coverage in 2015 and 2016 may explain why zip codes with later midpoint dates shifted more rapidly. Additionally, it may be that residents with lower incomes identified earlier with those in Flint and thus were more concerned about potentially contaminated water than their more affluent neighbors.

In addition to the regression factors, there is a significant positive correlation between change slope and midpoint date with a p-value of $4\times 10^{-4}$. The positive correlation between midpoint date and change slope is evident from a visual inspection of Figures \ref{fig:NYC_dates} and \ref{fig:NYC_slope}. This relation indicates that in zip codes that changed later, their changes were relatively quicker perhaps due to the prevalence of news coverage at that later time.

\subsection{United States Measles Data}
\label{sec:disease_exp}

Measles was nearly eradicated in the United States following the introduction of the measles vaccine in 1963. However, due to the vast geographic, ethnic, bureaucratic, and socio-economic heterogeneity in the United States we may expect differential effectiveness of the vaccination program, particularly in its initial years. We analyze monthly incidence data for measles from 1935 to 2003 in each of the continental United States and the District of Columbia. Incidence rates per 100,000 population based on historical population estimates are made publicly available by Project Tycho \citep{van2013contagious}. We fit the model to $\approx 33,000$ data points where $x\in \mathbb{R}^{3}$ with two spatial dimensions representing centroids of each state and one temporal dimension.

\begin{figure}[h]
\centering
 \includegraphics[width=0.75\textwidth]{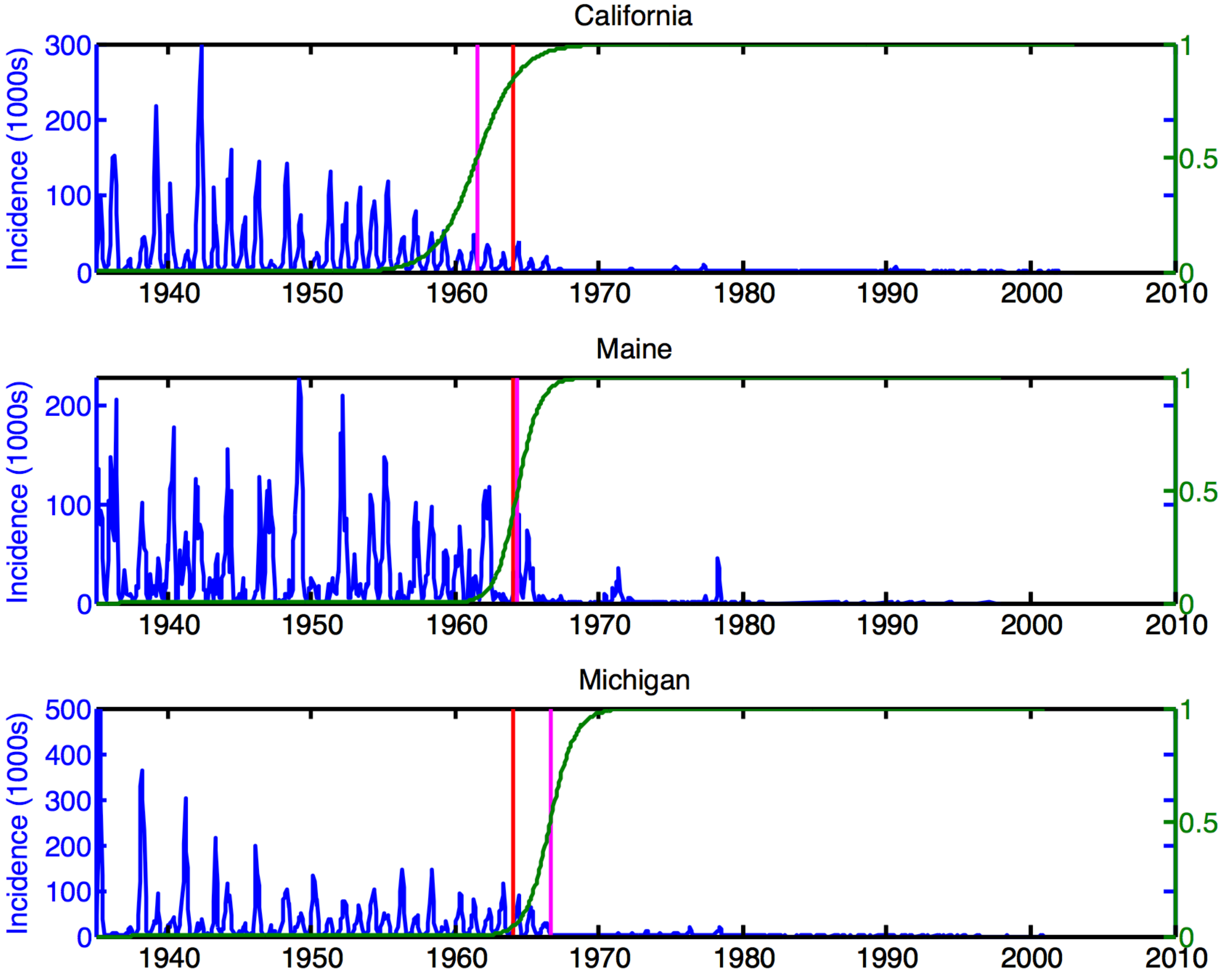}
\caption{Measles incidence levels from three states, 1935 to 2003. The green line plots $\sigma(w(x_{\text{state}}))$, the vertical red line indicates the vaccine in 1963, and the magenta line indicates $\sigma(w(x_{\text{state}}))=0.5$.}
\label{fig:tycho_states3}
\end{figure}

We apply GPCS with two latent functions, spectral mixture kernels, and $w(x)$ defined by RKS features. We do not provide prior information about the 1963 vaccination date. Results for three states are shown in Figure \ref{fig:tycho_states3} along with the predicted change surface for each state. The red line marks the vaccine year of 1963, while the magenta line marks where $\sigma(w(x_{\text{state}}))=0.5$. 

\begin{figure}[h]
\centering
 \includegraphics[width=1\textwidth]{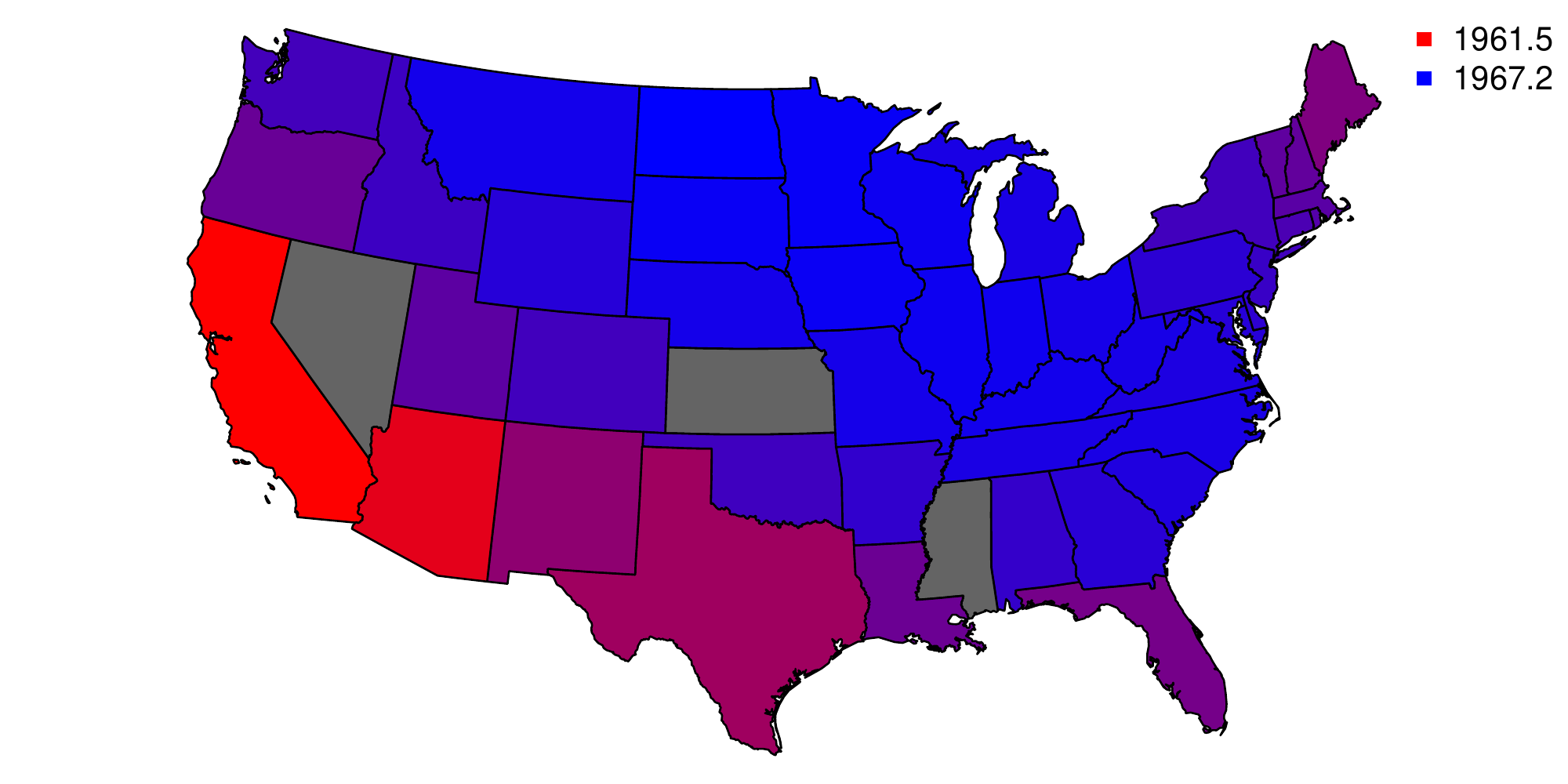}
\caption{U.S. states colored by the date where $\sigma(w(x_{\text{state}}))=0.5$. Red indicates earlier dates, with California being the earliest. Blue indicates later dates, with North Dakota being the latest. Grayed out states were missing in the dataset.} 
\label{fig:US_dates}
\end{figure}

GPCS correctly identified the time frame when the measles vaccine was released in the United States. Additionally, the model suggests that the effect of the measles vaccine varied both temporally and spatially. This finding again demonstrates the effectiveness of GPCS to detect changes in real world data while providing important insight into the change's dynamics that are not ascertainable through existing models. In Figure \ref{fig:US_dates} we depict the midpoint, $\sigma(w(x_{\text{state}}))=0.5$, for each state. We illustrate the spatial variation in the change surface midpoint by shading states with an early midpoint in red and states with a later midpoint in blue. We discover that there is an approximately 6 year range of midpoints between states, with California being the earliest and North Dakota being the latest.

\begin{figure}[h]
\centering
  \includegraphics[width=1\textwidth]{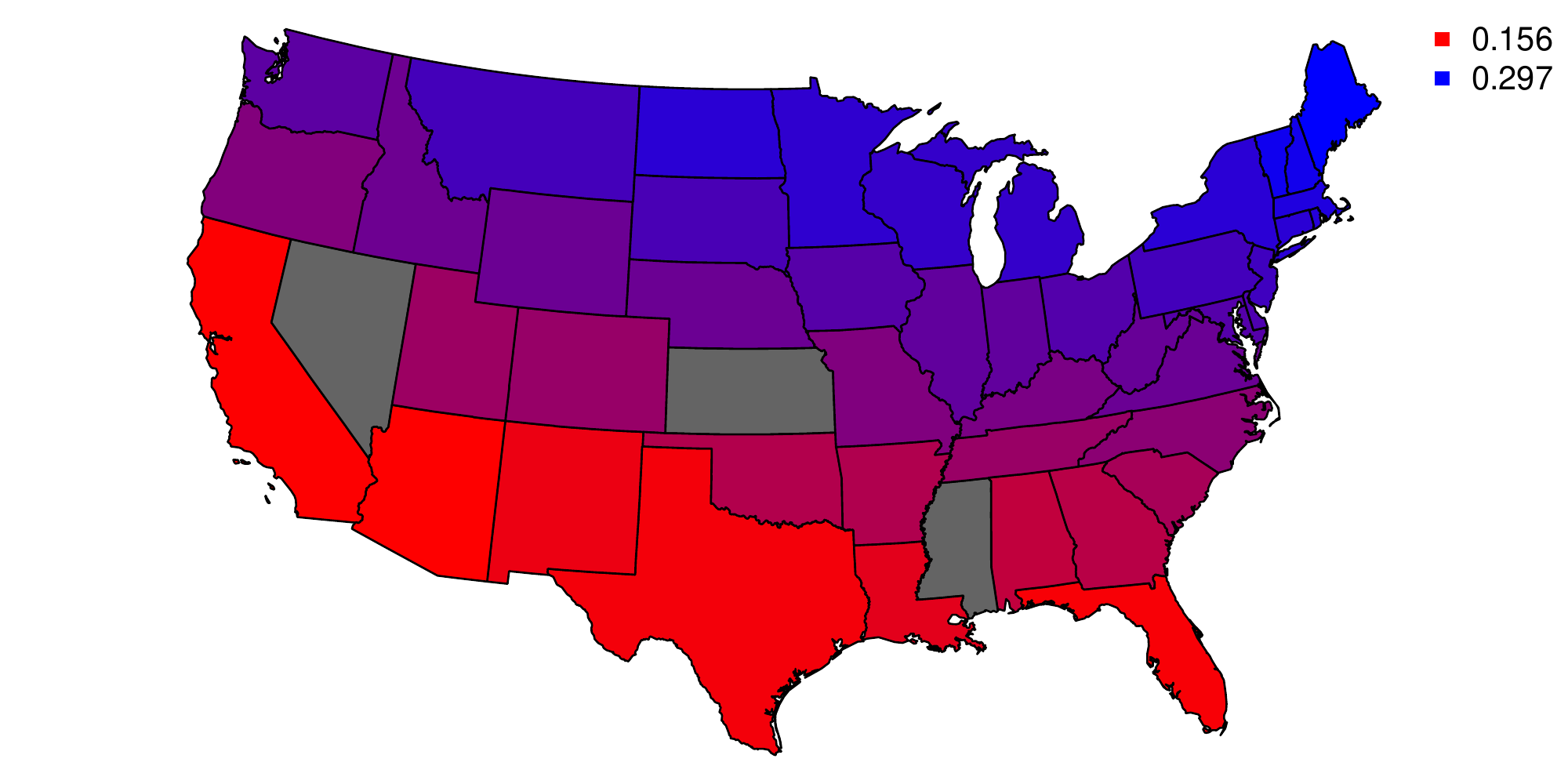}
\caption{U.S. states colored by the slope of $\sigma(w(x_{\text{state}}))$ from $0.25$ to $0.75$. Red indicates flatter slopes, with Arizona being the lowest. Blue indicates steeper slopes, with Maine being the highest. Grayed out states were missing in the dataset.}
\label{fig:US_slope}
\end{figure}

In Figure \ref{fig:US_slope} we depict the change surface slope from $\sigma(w(x_{\text{state}}))=0.25$ to $\sigma(w(x_{\text{state}}))=0.75$ for each state to estimate the rate of change. We illustrate the variation in slope by shading states with the flatter change regions in red and the steeper change regions in blue. Here we find that some states had approximately twice the rate of change as others, with Arizona having the flattest slope and Maine the steepest.

\paragraph{Regression analysis:} These variations in the change surface indicate that the measles vaccine may have affected states heterogeneously over space and time. In order to better understand these dynamics we considered demographic information that may have contributed to differences in measles vaccine program implementation and effectiveness. Specifically we examined potential factors influencing the midpoint shift date between the two regimes, $\sigma(w(x_{\text{state}}))=0.5$. Since the change surface shifts primarily during the 1960s and the measles vaccine is introduced in 1963, we consider historical census data only from 1960-1962 \citep{censuswebsite}. Factors included annual birth rate, death rates of different age segments, and population in each state. Since measles is often contracted by children and people are rarely diagnosed for the disease twice in their life (it is a permanently immunizing disease), previous literature has shown that birth rates and the size of a young non-immune population is important for understanding the pre-vaccination dynamics of measles \citep{Earn667}. Indeed, before the measles vaccine 5-9 year olds comprised 50\% of disease incidence \citep{CDCmeasles}. We also consider median household income and household income inequality for each state. Finally, we also consider the average annual temperature in each state.

\begin{table}[!htbp] \centering 
  \caption{Results from a linear regression to the measles incidence midpoint date, $\sigma(w(x_{\text{state}}))=0.5$. Variables are listed on the left while their coefficients, with standard errors in parentheses, are listed on the right. Asterisks indicate statistically significant variables.} 
  \label{tab:tycho_demographic} 
\begin{tabular}{@{\extracolsep{5pt}}lD{.}{.}{-3} } 
\\[-1.8ex]\hline 
\hline \\[-1.8ex] 
 & \multicolumn{1}{c}{\textit{Dependent variable:}} \\ 
\cline{2-2} 
\\[-1.8ex] & \multicolumn{1}{c}{Midpoint date} \\ 
\hline \\[-1.8ex] 
Log death rate aged 0-4 & -1.614 \\ 
  & (2.186) \\ 
  & \\ 
Log death rate aged 5-9 & 5.023 \\ 
  & (2.640) \\ 
  & \\ 
Log death rate aged 10+ & 7.651^{**} \\ 
  & (2.632) \\ 
  & \\ 
Log birth rate & -10.932 \\ 
  & (5.472) \\ 
  & \\ 
Gini of family income & 48.503^{**} \\ 
  & (17.461) \\ 
  & \\ 
Log median household income & 4.997 \\ 
  & (2.620) \\ 
  & \\ 
Log population & 0.117 \\ 
  & (0.228) \\ 
  & \\ 
Proportion of population aged 0-9 & 84.757^{*} \\ 
  & (32.784) \\ 
  & \\ 
Average temperature ($^{\circ}$F) & -0.093^{*} \\ 
  & (0.035) \\ 
  & \\ 
 Constant & 1,980.509^{**} \\ 
  & (24.237) \\ 
  & \\ 
\hline \\[-1.8ex] 
Observations & \multicolumn{1}{c}{46} \\ 
R$^{2}$ & \multicolumn{1}{c}{0.396} \\ 
Adjusted R$^{2}$ & \multicolumn{1}{c}{0.245} \\ 
\hline 
\hline \\[-1.8ex] 
\textit{Note:}  & \multicolumn{1}{r}{$^{*}$p$<$0.05; $^{**}$p$<$0.01} \\ 
\end{tabular} 
\end{table}

The results of a linear regression over all factors can be seen in Table \ref{tab:tycho_demographic}. Four variables were statistically significant at a p-value $<0.05$: the Gini coefficient of annual family income per state, average annual temperature, death rate of people aged 10+, and proportion of population aged 0-9. The Gini coefficient had a relatively large, positive correlation suggesting that wider family income inequality is associated with later dates of switching to the post-vaccine regime. One potential explanation of this phenomenon may be that states with higher Gini coefficients may have had large socio-economically depressed communities as well as substantial rural populations. Inoculation and vaccination education may have been more difficult in those communities and regions, thus delaying the midpoint date in those states. For example, Arkansas, Alabama, Kentucky, and Tennessee are all relatively rural states and have among the highest Gini coefficients. These states all have relatively late midpoint dates sometime in 1966. Another interesting example is the District of Columbia, which had the highest Gini coefficient. Although Washington D.C. is an urban center, it had also been an area of poverty and substandard local government, which may have contributed to its late change. Warmer temperatures are correlated with early midpoint dates perhaps due to biological mechanisms underlying the contagion of measles. Additionally, measles is spread through human contact which may also be affected by weather patterns. Death rates of people aged 10+ and relatively larger populations of children aged 0-9 were associated with later midpoint dates. Both of these factors indicate higher density of young children who may never have been affected by measles. This in turn may have increased the prevalence of the virus and delayed the midpoint date. 

In addition to the regression factors, there is a significant positive correlation between change slope and midpoint date with a p-value $< 2.2 \times 10^{-16}$, suggesting that states with later changes transition more quickly from the pre-vaccine regime to the post-vaccine regime. The steeper change slope may be due to other states already having inoculated their residents. Fewer measles cases nationwide could have enabled states with later midpoint dates to more effectively contain the disease in their borders.

While this analysis does not provide conclusive results about underlying causal mechanisms, it suggests that further scientific research is warranted to understand the political and demographic factors that contributed to differential effectiveness in the early years of the measles vaccine program. Indeed, one challenge in analyzing measles at a state-level aggregation is that measles disease dynamics may vary between cities even within states \citep{dalziel2016persistent}. Nevertheless, the results indicate that future vaccination programs should particularly consider how to quickly and effectively provide vaccinations to rural areas and provide additional resources to socioeconomically disadvantaged communities. Additionally, care should be taken when accounting for the effects of weather patterns and population dynamics.

\begin{figure}[h]
  \centering
  \subfloat[]{\includegraphics[width=0.5\textwidth]{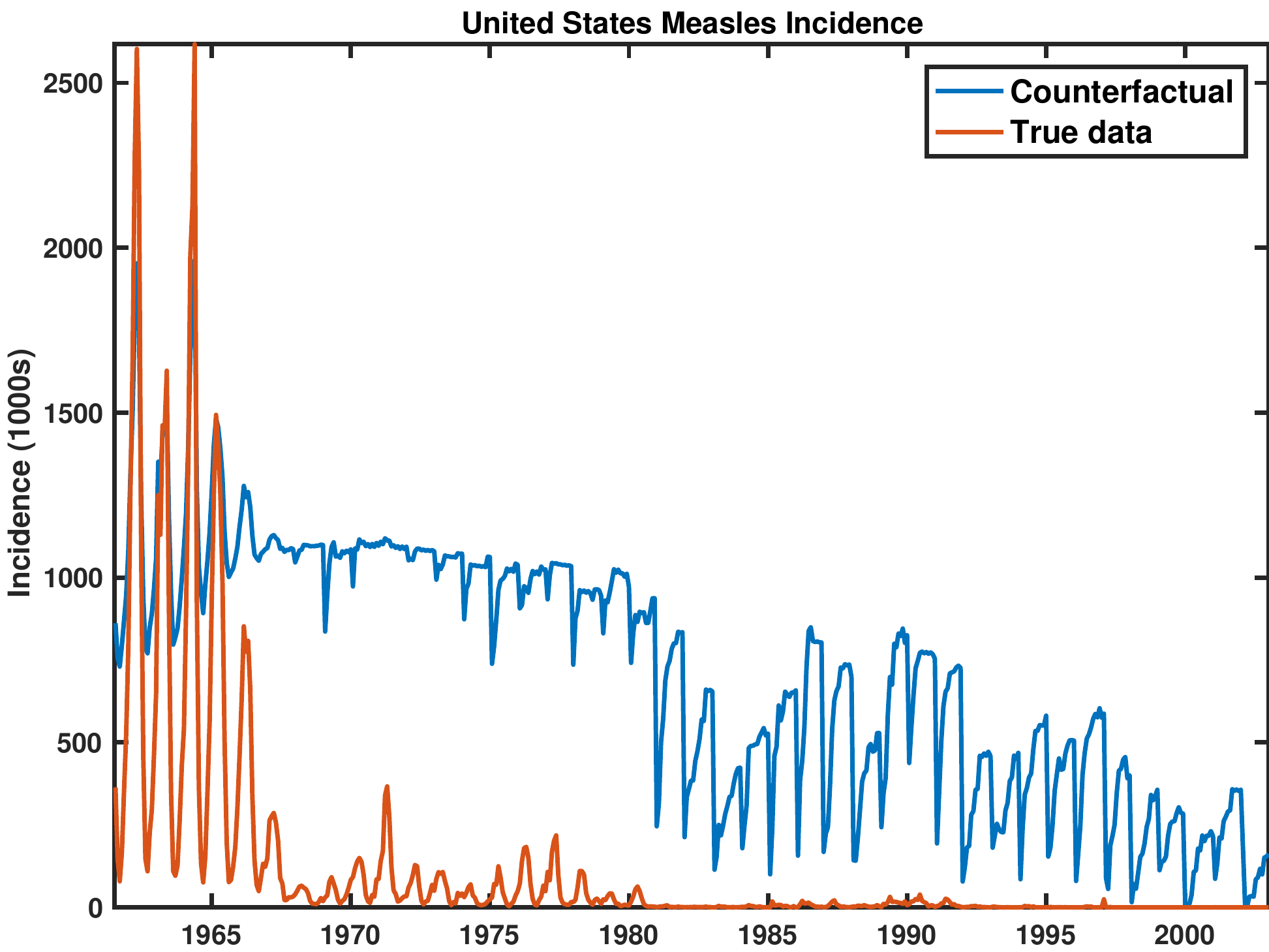}\label{fig:tycho_CF_perYear}}
  \hfill
  \subfloat[]{\includegraphics[width=0.5\textwidth]{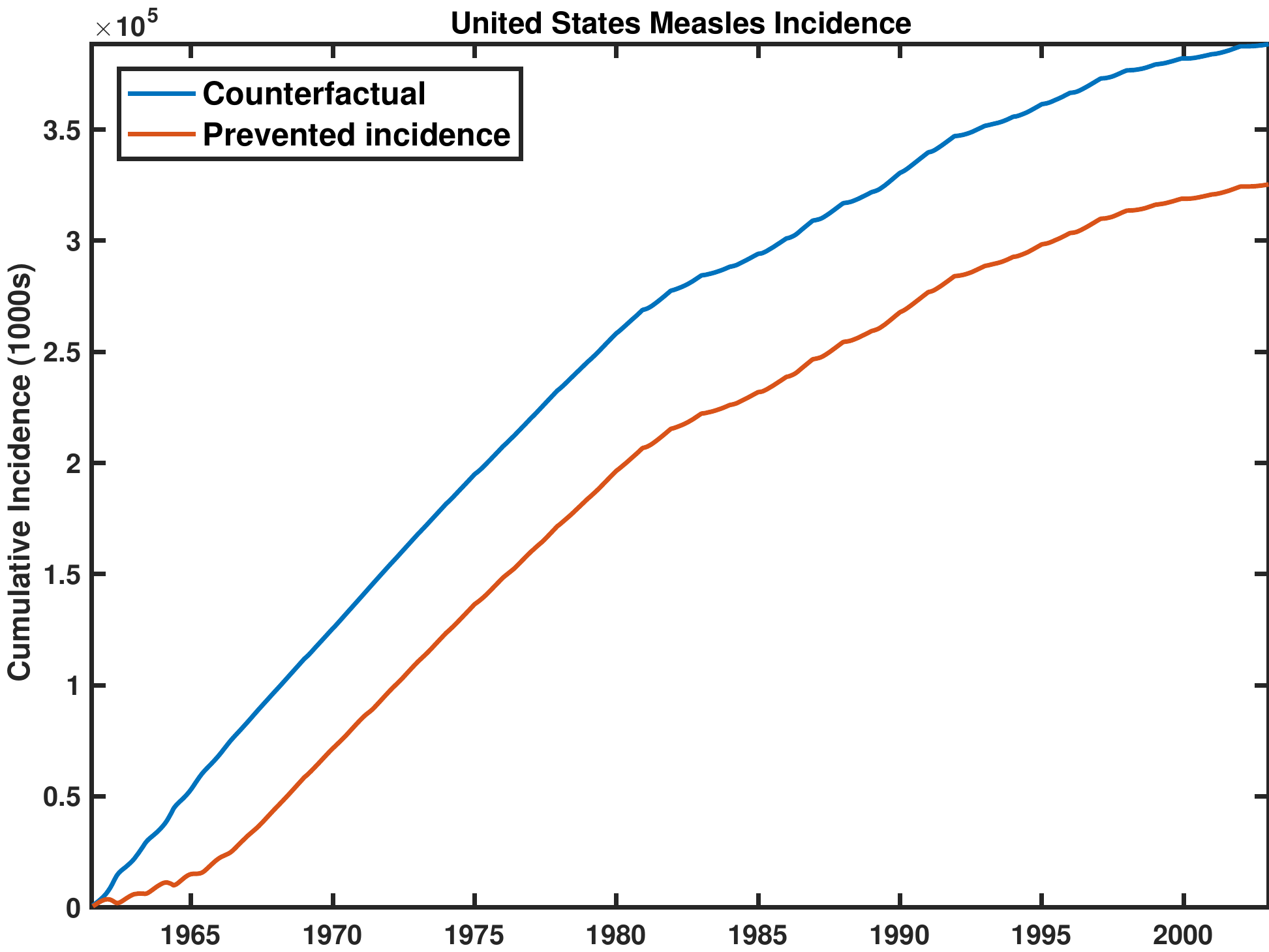}\label{fig:tycho_CF_cumulative}}
  \caption{Counterfactual posterior mean estimates for measles incidence. Plot (a) depicts the aggregated counterfactual posterior mean estimates over the entire United States. Plot (b) depicts the cumulative counterfactual incidence over the entire United States as well as estimating how many cases were ``prevented'' through the vaccination program under the assumption that the change surface corresponds to the vaccine intervention.}
  \label{fig:tycho_CF}
\end{figure}

\paragraph{Counterfactual analysis:} Using the counterfactual GPCS framework, we inferred the incidence of measles in the absence of the change surface identified by GPCS. We used the latent function that is dominant in the data before the measles vaccine to compute posterior estimates for measles incidence between the earliest detected midpoint date in 1961 and the end of the data in 2003. This estimation is inspired by the counterfactual estimation described in \citet{van2013contagious}. We argue that GPCS provides more believable counterfactual estimates than simple interpolations or regressions because GPCS is a more expressive model for measles dynamics and explicitly considers data variation both before and after the start of the measles vaccine program. Figure \ref{fig:tycho_CF} depicts the aggregated counterfactual posterior mean estimates over the entire United States. The left plot shows true and counterfactual monthly incidence, while the right plot depicts the cumulative counterfactual incidence. Under the assumption that the change surface reflects the causal effect of the vaccine program intervention, we also estimate how many cases were ``prevented'' through the vaccination program. Since disease dynamics may have many causal factors, we cannot disentangle the introduction of the measles vaccine from any contemporaneous societal or policy changes that may have impacted measles incidence. Thus these findings are a starting point for more extensive epidemiological research. Additionally, while we plot the posterior mean estimates, note that our confidence in these estimates diminishes as we consider counterfactual estimates far from the change region.

\section{Conclusion}
\label{sec:conclusions}

We presented change surfaces as an expressive generalization of changepoints that are able to model complex, multidimensional data with varying rates of change between latent functional regimes. Additionally, we showed how change surfaces can be used for counterfactual prediction. Yet we believe that change surfaces are not only a generalization of the statistical properties of change points, but truly a conceptual shift for modeling of distributional changes in data. Instead of attempting to discover discrete moments of change, change surfaces offer a more realistic framework for modeling complex data. Indeed, the change surface analyses presented in this paper demonstrate that they can provide scientific and public policy insights.  

As an instantiation of change surfaces, we presented GPCS, which uses independent Gaussian process priors and flexible RKS basis functions to enable a highly expressive change surface model. We derived counterfactual prediction methods for GPCS that naturally provide counterfactual posterior mean and variance estimates. We also demonstrate that probabilistic inference within GPCS automatically discourages extraneous complexity, naturally leading to interpretable generative hypotheses for our observations. To support GPCS we also created a novel scalable inference method for multiple additive kernels using the Weyl bound. This result extends far beyond change surfaces, enabling scalable Gaussian processes with non-separable covariance structures over multiple dimensions. Additionally, we developed an effective approach for initializing expressive spectral mixture kernels.

Using change surfaces we are able to model complex, spatio-temporal data with expressivity and clarity. We studied requests for lead testing kits in New York City between 2014-2016, a period of heightened concern regarding water quality around the United States. GPCS identified a change in the dynamics of requests, but perhaps more importantly, it illustrated how that change developed over time and varied over space. The spatio-temporal heterogeneity modeled by GPCS enabled further investigation into demographic factors that may have influenced the behavior of residents in various parts of the city. This analysis is only possible with a change surface model because standard changepoint approaches are only able to provide single, point-in-time estimates of a midpoint date. Policy makers are often interested in learning how public health risks or legal regulations affect various populations. Our results demonstrate that change surfaces can be a particularly effective method for policy makers to understand how changes develop and are distributed over a multidimensional domain.

We also used GPCS to model measles incidence in the United States over the course of the twentieth century. In addition to identifying a change in regimes around the introduction of the measles vaccine in 1963, we used the fitted change surface to illuminate heterogeneity across states. The differential change rates and midpoint dates in each state could have important scientific and policy implications for vaccination campaigns. To this end, we provide a regression analysis of  institutional and demographic factors that may have influenced the impact of the measles vaccination program.

Finally, we are excited about how the introduction of change surfaces could inspire further research into expressive modeling of complex changes. As we emphasized in Section~\ref{sec:GPCSmodel}, our use of Gaussian processes in GPCS presents but one approach to modeling change surfaces. Future work may provide alternative methods for characterizing change surfaces using statistical approaches beyond Gaussian processes. For example, other instantiations of change surfaces could utilize penalty terms to enforce the soft mutual exclusivity between the functional regimes, or else employ decision-tree like structures to divide the domain. Another fruitful methodological avenue could extend the retrospective analysis in this paper to address online or sequential change surface detection. Additionally, change surfaces may be further used for causal inference in conjuction with natural experiments, which are often used by econometricians for causal inference in observational data. For example, change surfaces may help discover regression discontinuity designs \citep{herlands2018automated} or identify heterogeneous treatment effects in real-valued data.

\acks{Thank you to Wilbert Van Panhuis and Seth Flaxman for providing much appreciated insight and suggestions. This material is based upon work supported by NSF Graduate Research Fellowship DGE-1252522 as well as NSF awards IIS-0953330 and IIS-1563887.}

\newpage

\vskip 0.2in

\nocite{herlands2016scalable}
\bibliography{GPCS}

\begin{thebibliography}{75}
\providecommand{\natexlab}[1]{#1}
\providecommand{\url}[1]{\texttt{#1}}
\expandafter\ifx\csname urlstyle\endcsname\relax
  \providecommand{\doi}[1]{doi: #1}\else
  \providecommand{\doi}{doi: \begingroup \urlstyle{rm}\Url}\fi

\bibitem[Abadie et~al.(2002)Abadie, Angrist, and
  Imbens]{abadie2002instrumental}
Alberto Abadie, Joshua Angrist, and Guido Imbens.
\newblock Instrumental variables estimates of the effect of subsidized training
  on the quantiles of trainee earnings.
\newblock \emph{Econometrica}, 70\penalty0 (1):\penalty0 91--117, 2002.

\bibitem[Adams and MacKay(2007)]{adams2007bayesian}
Ryan~Prescott Adams and David~JC MacKay.
\newblock Bayesian online changepoint detection.
\newblock \emph{stat}, 1050:\penalty0 19, 2007.

\bibitem[Aminikhanghahi and Cook(2017)]{aminikhanghahi2017survey}
Samaneh Aminikhanghahi and Diane~J Cook.
\newblock A survey of methods for time series change point detection.
\newblock \emph{Knowledge and information systems}, 51\penalty0 (2):\penalty0
  339--367, 2017.

\bibitem[Athey and Imbens(2006)]{athey2006identification}
Susan Athey and Guido~W Imbens.
\newblock Identification and inference in nonlinear difference-in-differences
  models.
\newblock \emph{Econometrica}, 74\penalty0 (2):\penalty0 431--497, 2006.

\bibitem[Aue and Horv{\'a}th(2013)]{aue2013structural}
Alexander Aue and Lajos Horv{\'a}th.
\newblock Structural breaks in time series.
\newblock \emph{Journal of Time Series Analysis}, 34\penalty0 (1):\penalty0
  1--16, 2013.

\bibitem[Bhatia(2013)]{bhatia2013matrix}
Rajendra Bhatia.
\newblock \emph{Matrix analysis}, volume 169.
\newblock Springer Science \& Business Media, 2013.

\bibitem[Brodersen et~al.(2015)Brodersen, Gallusser, Koehler, Remy, and
  Scott]{kay2015inferring}
Kay~H. Brodersen, Fabian Gallusser, Jim Koehler, Nicolas Remy, and Steven~L.
  Scott.
\newblock Inferring causal impact using bayesian structural time-series models.
\newblock \emph{Annals of Applied Statistics}, 9:\penalty0 247--274, 2015.

\bibitem[Brodsky and Darkhovsky(2013)]{brodsky2013nonparametric}
E~Brodsky and Boris~S Darkhovsky.
\newblock \emph{Nonparametric methods in change point problems}, volume 243.
\newblock Springer Science \& Business Media, 2013.

\bibitem[Carlin et~al.(1992)Carlin, Gelfand, and Smith]{carlin1992hierarchical}
Bradley~P Carlin, Alan~E Gelfand, and Adrian~FM Smith.
\newblock Hierarchical bayesian analysis of changepoint problems.
\newblock \emph{Applied statistics}, pages 389--405, 1992.

\bibitem[Census~Bureau(1999)]{censuswebsite}
U.~S. Census~Bureau.
\newblock United states historical census data.
\newblock \url{https://www.census.gov/hhes/www/income/data/historical/state/},
  1999.
\newblock Accessed: 2016-4-10.

\bibitem[Census~Bureau(2014{\natexlab{a}})]{acs2014year1}
U.~S. Census~Bureau.
\newblock American community survey 1-year estimates.
\newblock \url{http://factfinder.census.gov/}, 2014{\natexlab{a}}.
\newblock Accessed: 2016-4-10.

\bibitem[Census~Bureau(2014{\natexlab{b}})]{acs2014year5}
U.~S. Census~Bureau.
\newblock American community survey 5-year estimates.
\newblock \url{http://factfinder.census.gov/}, 2014{\natexlab{b}}.
\newblock Accessed: 2016-4-10.

\bibitem[Chen and Gupta(2011)]{chen2011parametric}
Jie Chen and Arjun~K Gupta.
\newblock \emph{Parametric statistical change point analysis: With applications
  to genetics, medicine, and finance}.
\newblock Springer Science \& Business Media, 2011.

\bibitem[Chernoff and Zacks(1964)]{chernoff1964estimating}
Herman Chernoff and Shelemyahu Zacks.
\newblock Estimating the current mean of a normal distribution which is
  subjected to changes in time.
\newblock \emph{The Annals of Mathematical Statistics}, pages 999--1018, 1964.

\bibitem[City(2016)]{leadtestingkit}
New~York City.
\newblock Water lead test kit request.
\newblock
  \url{http://www1.nyc.gov/nyc-resources/service/1266/water-lead-test-kit-request},
  2016.
\newblock Accessed: 2016-4-10.

\bibitem[Control and Prevention(2016)]{CDCmeasles}
Centers For~Disease Control and Prevention.
\newblock Epidemiology and prevention of vaccine-preventable diseases, 2016.
\newblock URL \url{https://www.cdc.gov/vaccines/pubs/pinkbook/meas.html}.

\bibitem[Dalziel et~al.(2016)Dalziel, Bj{\o}rnstad, van Panhuis, Burke,
  Metcalf, and Grenfell]{dalziel2016persistent}
Benjamin~D Dalziel, Ottar~N Bj{\o}rnstad, Willem~G van Panhuis, Donald~S Burke,
  C~Jessica~E Metcalf, and Bryan~T Grenfell.
\newblock Persistent chaos of measles epidemics in the prevaccination united
  states caused by a small change in seasonal transmission patterns.
\newblock \emph{PLoS Comput Biol}, 12\penalty0 (2):\penalty0 e1004655, 2016.

\bibitem[Earn et~al.(2000)Earn, Rohani, Bolker, and Grenfell]{Earn667}
David J.~D. Earn, Pejman Rohani, Benjamin~M. Bolker, and Bryan~T. Grenfell.
\newblock A simple model for complex dynamical transitions in epidemics.
\newblock \emph{Science}, 287\penalty0 (5453):\penalty0 667--670, 2000.
\newblock ISSN 0036-8075.
\newblock \doi{10.1126/science.287.5453.667}.
\newblock URL \url{http://science.sciencemag.org/content/287/5453/667}.

\bibitem[Editorial~Board(2016)]{newark}
The Editorial~Board.
\newblock Poisoned water in newark schools.
\newblock \emph{New York Times}, March 2016.

\bibitem[Fiedler(1971)]{fiedler1971bounds}
Miroslav Fiedler.
\newblock Bounds for the determinant of the sum of hermitian matrices.
\newblock \emph{Proceedings of the American Mathematical Society}, pages
  27--31, 1971.

\bibitem[Flaxman et~al.(2015)Flaxman, Wilson, Neill, Nickisch, and
  Smola]{flaxman2015fast}
Seth~R Flaxman, Andrew~Gordon Wilson, Daniel~B Neill, Hannes Nickisch, and
  Alexander~J Smola.
\newblock Fast kronecker inference in gaussian processes with non-gaussian
  likelihoods.
\newblock \emph{International Conference on Machine Learning 2015}, 2015.

\bibitem[Garnett et~al.(2009)Garnett, Osborne, and
  Roberts]{garnett2009sequential}
Roman Garnett, Michael~A Osborne, and Stephen~J Roberts.
\newblock Sequential bayesian prediction in the presence of changepoints.
\newblock In \emph{Proceedings of the 26th Annual International Conference on
  Machine Learning}, pages 345--352. ACM, 2009.

\bibitem[Gay(2016)]{nycleadhousing}
Mara Gay.
\newblock Elevated levels of lead found in water of some vacant public-housing
  apartments.
\newblock \emph{Wall Street Journal}, 2016.

\bibitem[Guan(2004)]{guan2004semiparametric}
Zhong Guan.
\newblock A semiparametric changepoint model.
\newblock \emph{Biometrika}, pages 849--862, 2004.

\bibitem[Hartford et~al.(2016)Hartford, Lewis, Leyton-Brown, and
  Taddy]{hartford2016counterfactual}
Jason Hartford, Greg Lewis, Kevin Leyton-Brown, and Matt Taddy.
\newblock Counterfactual prediction with deep instrumental variables networks.
\newblock \emph{arXiv preprint arXiv:1612.09596}, 2016.

\bibitem[Herlands et~al.(2016)Herlands, Wilson, Nickisch, Flaxman, Neill,
  Van~Panhuis, and Xing]{herlands2016scalable}
William Herlands, Andrew Wilson, Hannes Nickisch, Seth Flaxman, Daniel Neill,
  Wilbert Van~Panhuis, and Eric Xing.
\newblock Scalable gaussian processes for characterizing multidimensional
  change surfaces.
\newblock In \emph{Artificial Intelligence and Statistics}, pages 1013--1021,
  2016.

\bibitem[Herlands et~al.(2018)Herlands, McFowland~III, Wilson, and
  Neill]{herlands2018automated}
William Herlands, Edward McFowland~III, Andrew~Gordon Wilson, and Daniel~B
  Neill.
\newblock Automated local regression discontinuity design discovery.
\newblock In \emph{Proceedings of the 24th ACM SIGKDD International Conference
  on Knowledge Discovery \& Data Mining}, pages 1512--1520. ACM, 2018.

\bibitem[Holland(1986)]{holland1986statistics}
Paul~W Holland.
\newblock Statistics and causal inference.
\newblock \emph{Journal of the American statistical Association}, 81\penalty0
  (396):\penalty0 945--960, 1986.

\bibitem[Horv{\'a}th and Rice(2014)]{horvath2014extensions}
Lajos Horv{\'a}th and Gregory Rice.
\newblock Extensions of some classical methods in change point analysis.
\newblock \emph{Test}, 23\penalty0 (2):\penalty0 219--255, 2014.

\bibitem[Ivanoff and Merzbach(2010)]{ivanoff2010optimal}
B~Gail Ivanoff and Ely Merzbach.
\newblock Optimal detection of a change-set in a spatial poisson process.
\newblock \emph{The Annals of Applied Probability}, pages 640--659, 2010.

\bibitem[James and Matteson(2013)]{james2013ecp}
Nicholas~A James and David~S Matteson.
\newblock ecp: An r package for nonparametric multiple change point analysis of
  multivariate data.
\newblock \emph{arXiv preprint arXiv:1309.3295}, 2013.

\bibitem[Jarrett(1979)]{jarrett1979note}
RG~Jarrett.
\newblock A note on the intervals between coal-mining disasters.
\newblock \emph{Biometrika}, pages 191--193, 1979.

\bibitem[Johansson et~al.(2016)Johansson, Shalit, and
  Sontag]{johansson2016learning}
Fredrik~D Johansson, Uri Shalit, and David Sontag.
\newblock Learning representations for counterfactual inference.
\newblock \emph{arXiv preprint arXiv:1605.03661}, 2016.

\bibitem[Kapur et~al.(2011)Kapur, Pham, Gupta, and Jha]{Kapur2011}
P.~K. Kapur, H.~Pham, A.~Gupta, and P.~C. Jha.
\newblock \emph{Change-Point Models}, pages 171--213.
\newblock Springer London, London, 2011.
\newblock ISBN 978-0-85729-204-9.
\newblock \doi{10.1007/978-0-85729-204-9_5}.
\newblock URL \url{http://dx.doi.org/10.1007/978-0-85729-204-9_5}.

\bibitem[Keshavarz et~al.(2018)Keshavarz, Scott, and
  Nguyen]{keshavarz2018optimal}
Hossein Keshavarz, Clayton Scott, and XuanLong Nguyen.
\newblock Optimal change point detection in gaussian processes.
\newblock \emph{Journal of Statistical Planning and Inference}, 193:\penalty0
  151--178, 2018.

\bibitem[Killick et~al.(2012)Killick, Fearnhead, and
  Eckley]{killick2012optimal}
Rebecca Killick, Paul Fearnhead, and IA~Eckley.
\newblock Optimal detection of changepoints with a linear computational cost.
\newblock \emph{Journal of the American Statistical Association}, 107\penalty0
  (500):\penalty0 1590--1598, 2012.

\bibitem[L{\'a}zaro-Gredilla et~al.(2010)L{\'a}zaro-Gredilla,
  Qui{\~n}onero-Candela, Rasmussen, and Figueiras-Vidal]{lazaro2010sparse}
Miguel L{\'a}zaro-Gredilla, Joaquin Qui{\~n}onero-Candela, Carl~Edward
  Rasmussen, and An{\'\i}bal~R Figueiras-Vidal.
\newblock Sparse spectrum gaussian process regression.
\newblock \emph{The Journal of Machine Learning Research}, 11:\penalty0
  1865--1881, 2010.

\bibitem[Lloyd et~al.(2014)Lloyd, Duvenaud, Grosse, Tenenbaum, and
  Ghahramani]{lloyd2014automatic}
James~Robert Lloyd, David Duvenaud, Roger Grosse, Joshua Tenenbaum, and Zoubin
  Ghahramani.
\newblock Automatic construction and natural-language description of
  nonparametric regression models.
\newblock In \emph{Twenty-Eighth AAAI Conference on Artificial Intelligence},
  2014.

\bibitem[MacKay(1998)]{mackay1998introduction}
David~JC MacKay.
\newblock Introduction to gaussian processes.
\newblock \emph{NATO ASI Series F Computer and Systems Sciences}, 168:\penalty0
  133--166, 1998.

\bibitem[MacKay(2003)]{mackay2003information}
David~JC MacKay.
\newblock \emph{Information theory, inference and learning algorithms}.
\newblock Cambridge university press, 2003.

\bibitem[Majumdar et~al.(2005)Majumdar, Gelfand, and
  Banerjee]{majumdar2005spatio}
Anandamayee Majumdar, Alan~E Gelfand, and Sudipto Banerjee.
\newblock Spatio-temporal change-point modeling.
\newblock \emph{Journal of Statistical Planning and Inference}, 130\penalty0
  (1):\penalty0 149--166, 2005.

\bibitem[Martin(1990)]{martin1990use}
RJ~Martin.
\newblock The use of time-series models and methods in the analysis of
  agricultural field trials.
\newblock \emph{Communications in Statistics-Theory and Methods}, 19\penalty0
  (1):\penalty0 55--81, 1990.

\bibitem[McFowland et~al.(2016)McFowland, Somanchi, and Neill]{edTESS}
Ed~McFowland, Sriram Somanchi, and Daniel~B. Neill.
\newblock Efficient discovery of heterogeneous treatment effects in randomized
  experiments via anomalous pattern detection.
\newblock \emph{Working paper}, 2016.

\bibitem[Mining(2017)]{law1887coal}
Scottish Mining.
\newblock Coal mines regulation act, 2017.
\newblock URL \url{http://www.scottishmining.co.uk/256.html}.

\bibitem[Minka(2001)]{minka2001automatic}
Thomas~P Minka.
\newblock Automatic choice of dimensionality for pca.
\newblock In \emph{Advances in neural information processing systems}, pages
  598--604, 2001.

\bibitem[M{\o}ller et~al.(1998)M{\o}ller, Syversveen, and
  Waagepetersen]{moller1998log}
Jesper M{\o}ller, Anne~Randi Syversveen, and Rasmus~Plenge Waagepetersen.
\newblock Log gaussian cox processes.
\newblock \emph{Scandinavian journal of statistics}, 25\penalty0 (3):\penalty0
  451--482, 1998.

\bibitem[Nicholls and Nunn(2010)]{nicholls2010building}
Geoff~K Nicholls and Patrick~D Nunn.
\newblock On building and fitting a spatio-temporal change-point model for
  settlement and growth at bourewa, fiji islands.
\newblock \emph{arXiv preprint arXiv:1006.5575}, 2010.

\bibitem[Page(1954)]{page1954continuous}
ES~Page.
\newblock Continuous inspection schemes.
\newblock \emph{Biometrika}, 41\penalty0 (1/2):\penalty0 100--115, 1954.

\bibitem[Raftery and Akman(1986)]{raftery1986bayesian}
AE~Raftery and VE~Akman.
\newblock Bayesian analysis of a poisson process with a change-point.
\newblock \emph{Biometrika}, pages 85--89, 1986.

\bibitem[Rahimi and Recht(2007)]{rahimi2007random}
Ali Rahimi and Benjamin Recht.
\newblock Random features for large-scale kernel machines.
\newblock In \emph{Advances in neural information processing systems}, pages
  1177--1184, 2007.

\bibitem[Rasmussen and Williams(2006)]{rasmussen2006gaussian}
Carl Rasmussen and Chris Williams.
\newblock \emph{Gaussian Processes for Machine Learning}.
\newblock MIT Press, 2006.

\bibitem[Rasmussen and Ghahramani(2001)]{rasmussen2001occam}
Carl~Edward Rasmussen and Zoubin Ghahramani.
\newblock Occam's razor.
\newblock In \emph{Advances in neural information processing systems}, pages
  294--300, 2001.

\bibitem[Rasmussen and Nickisch(2010)]{rasmussen2010gaussian}
Carl~Edward Rasmussen and Hannes Nickisch.
\newblock Gaussian processes for machine learning (gpml) toolbox.
\newblock \emph{The Journal of Machine Learning Research}, 11:\penalty0
  3011--3015, 2010.

\bibitem[Reece et~al.(2015)Reece, Garnett, Osborne, and
  Roberts]{reece2015anomaly}
Steven Reece, Roman Garnett, Michael Osborne, and Stephen Roberts.
\newblock Anomaly detection and removal using non-stationary gaussian
  processes.
\newblock \emph{arXiv preprint arXiv:1507.00566}, 2015.

\bibitem[Ross(2013)]{ross2013parametric}
Gordon~J Ross.
\newblock Parametric and nonparametric sequential change detection in r: The
  cpm package.
\newblock \emph{Journal of Statistical Software}, page~78, 2013.

\bibitem[Rubin(1986)]{rubin1986comment}
Donald~B Rubin.
\newblock Comment: Which ifs have causal answers.
\newblock \emph{Journal of the American Statistical Association}, 81\penalty0
  (396):\penalty0 961--962, 1986.

\bibitem[Rubin(2005)]{rubin2005causal}
Donald~B Rubin.
\newblock Causal inference using potential outcomes: Design, modeling,
  decisions.
\newblock \emph{Journal of the American Statistical Association}, 100\penalty0
  (469):\penalty0 322--331, 2005.

\bibitem[Saat{\c{c}}i(2011)]{saatcci2012scalable}
Yunus Saat{\c{c}}i.
\newblock \emph{Scalable inference for structured Gaussian process models}.
\newblock PhD thesis, University of Cambridge, 2011.

\bibitem[Saat{\c{c}}i et~al.(2010)Saat{\c{c}}i, Turner, and
  Rasmussen]{saatcci2010gaussian}
Yunus Saat{\c{c}}i, Ryan~D Turner, and Carl~E Rasmussen.
\newblock Gaussian process change point models.
\newblock In \emph{Proceedings of the 27th International Conference on Machine
  Learning (ICML-10)}, pages 927--934, 2010.

\bibitem[Sch{\"o}lkopf and Smola(2002)]{scholkopf2002learning}
Bernhard Sch{\"o}lkopf and Alexander~J Smola.
\newblock \emph{Learning with kernels: support vector machines, regularization,
  optimization, and beyond}.
\newblock MIT press, 2002.

\bibitem[Schulam and Saria(2017)]{schulam2017reliable}
Peter Schulam and Suchi Saria.
\newblock Reliable decision support using counterfactual models.
\newblock In \emph{Advances in Neural Information Processing Systems}, pages
  1697--1708, 2017.

\bibitem[Sharkey and Killick(2014)]{sharkey2014nonparametric}
Paul Sharkey and Rebecca Killick.
\newblock Nonparametric methods for online changepoint detection.
\newblock Technical Report STOR601, Lancaster University, 2014.

\bibitem[Shirota and Gelfand(2016)]{shirota2016inference}
Shinichiro Shirota and Alan~E Gelfand.
\newblock Inference for log gaussian cox processes using an approximate
  marginal posterior.
\newblock \emph{arXiv preprint arXiv:1611.10359}, 2016.

\bibitem[Tartakovsky et~al.(2013)Tartakovsky, Polunchenko, and
  Sokolov]{tartakovsky2013efficient}
Alexander~G Tartakovsky, Aleksey~S Polunchenko, and Grigory Sokolov.
\newblock Efficient computer network anomaly detection by changepoint detection
  methods.
\newblock \emph{IEEE Journal of Selected Topics in Signal Processing},
  7\penalty0 (1):\penalty0 4--11, 2013.

\bibitem[van Panhuis et~al.(2013)van Panhuis, Grefenstette, Jung, Chok, Cross,
  Eng, Lee, Zadorozhny, Brown, Cummings, et~al.]{van2013contagious}
Willem~G van Panhuis, John Grefenstette, Su~Yon Jung, Nian~Shong Chok, Anne
  Cross, Heather Eng, Bruce~Y Lee, Vladimir Zadorozhny, Shawn Brown, Derek
  Cummings, et~al.
\newblock Contagious diseases in the united states from 1888 to the present.
\newblock \emph{The New England journal of medicine}, 369\penalty0
  (22):\penalty0 2152, 2013.

\bibitem[Verbitsky-Savitz and Raudenbush(2012)]{verbitsky2012epidemiologic}
Natalya Verbitsky-Savitz and Stephen~W Raudenbush.
\newblock Causal inference under interference in spatial settings: a case study
  evaluating community policing program in chicago.
\newblock \emph{Epidemiological Methods}, 1:\penalty0 106--130, 2012.

\bibitem[Water~Study(2015)]{flintwaterstudy}
Flint Water~Study.
\newblock Flint water study: Articles in the press.
\newblock \url{http://flintwaterstudy.org/articles-in-the-press/}, 2015.
\newblock Accessed: 2016-4-10.

\bibitem[Weyl(1912)]{weyl1912asymptotische}
Hermann Weyl.
\newblock {Das asymptotische Verteilungsgesetz der Eigenwerte linearer
  partieller Differentialgleichungen (mit einer Anwendung auf die Theorie der
  Hohlraumstrahlung)}.
\newblock \emph{Mathematische Annalen}, 71\penalty0 (4):\penalty0 441--479,
  1912.

\bibitem[Wilson and Adams(2013)]{wilson2013gaussian}
Andrew Wilson and Ryan Adams.
\newblock Gaussian process kernels for pattern discovery and extrapolation.
\newblock In \emph{Proceedings of The 30th International Conference on Machine
  Learning}, pages 1067--1075, 2013.

\bibitem[Wilson and Nickisch(2015)]{wilson2015kernel}
Andrew Wilson and Hannes Nickisch.
\newblock Kernel interpolation for scalable structured gaussian processes
  (kiss-gp).
\newblock In \emph{Proceedings of the 32nd International Conference on Machine
  Learning (ICML-15)}, pages 1775--1784, 2015.

\bibitem[Wilson et~al.(2012)Wilson, Ghahramani, and
  Knowles]{wilson2011gaussian}
Andrew Wilson, Zoubin Ghahramani, and David~A Knowles.
\newblock Gaussian process regression networks.
\newblock In \emph{Proceedings of the 29th International Conference on Machine
  Learning (ICML-12)}, pages 599--606, 2012.

\bibitem[Wilson et~al.(2014)Wilson, Gilboa, Cunningham, and
  Nehorai]{wilson2014fast}
Andrew Wilson, Elad Gilboa, John~P Cunningham, and Arye Nehorai.
\newblock Fast kernel learning for multidimensional pattern extrapolation.
\newblock In \emph{Advances in Neural Information Processing Systems}, pages
  3626--3634, 2014.

\bibitem[Wilson et~al.(2016{\natexlab{a}})Wilson, Hu, Salakhutdinov, and
  Xing]{wilson2016stochastic}
Andrew~G Wilson, Zhiting Hu, Ruslan~R Salakhutdinov, and Eric~P Xing.
\newblock Stochastic variational deep kernel learning.
\newblock In \emph{Advances in Neural Information Processing Systems}, pages
  2586--2594, 2016{\natexlab{a}}.

\bibitem[Wilson(2014)]{wilson2014covariance}
Andrew~Gordon Wilson.
\newblock \emph{Covariance kernels for fast automatic pattern discovery and
  extrapolation with Gaussian processes}.
\newblock PhD thesis, PhD thesis, University of Cambridge, 2014.

\bibitem[Wilson et~al.(2016{\natexlab{b}})Wilson, Hu, Salakhutdinov, and
  Xing]{wilson2016deep}
Andrew~Gordon Wilson, Zhiting Hu, Ruslan Salakhutdinov, and Eric~P Xing.
\newblock Deep kernel learning.
\newblock In \emph{Artificial Intelligence and Statistics}, pages 370--378,
  2016{\natexlab{b}}.

\end{thebibliography}

\end{document}